%% file: acl_latex.tex
\pdfoutput=1

\documentclass[11pt]{article}

\newif\ifreviewcopy
\reviewcopyfalse

\usepackage{acl}
\usepackage{times}
\usepackage{latexsym}
\usepackage[T1]{fontenc}
\usepackage[utf8]{inputenc}
\usepackage{microtype}
\usepackage[colorinlistoftodos]{todonotes}
\usepackage{marginnote}
\usepackage{amsmath}
\usepackage{enumitem}
\usepackage{placeins}
\usepackage{multicol}
\usepackage{booktabs}
\usepackage[inkscapearea=page]{svg}
\usepackage{amsmath}
\usepackage[nameinlink]{cleveref}
\usepackage{float}
\usepackage{afterpage}
\usepackage{pgfplots}
\usepackage{graphicx}
\usepackage{ulem}
\usepackage[multiple,hang,flushmargin]{footmisc}
\usepackage{multirow}
\usepackage{multicol}
\usepackage{makecell}
\usepackage{mathtools}
\usepackage{amssymb}
\usepackage{pifont}
\usepackage{xspace}
\usepackage{csquotes}
\usepackage{tabularx}
\usepackage{algorithmicx}
\usepackage{algorithm}
\usepackage{algpseudocode}

\input{sparql_style.tex}
\input{markdown_style.tex}

\input{commands.tex}

\def\DatasetNameTitle{Pap2Pat\xspace}
\def\DatasetName{{\sc{\DatasetNameTitle}}\xspace}

\def\MethodName{{\mbox{\textsc{COPGen}}}\xspace}

\title{\DatasetName: Benchmarking Outline-Guided Long-Text\\Patent Generation with Patent-Paper Pairs}  %

\author{
 \textbf{Valentin Knappich\textsuperscript{1,2}}\hfill
 \textbf{Anna Hätty\textsuperscript{1}}\hfill
 \textbf{Simon Razniewski\textsuperscript{3}}\hfill
 \textbf{Annemarie Friedrich\textsuperscript{2}}
\\
\\
 \textsuperscript{1}Bosch Center for AI, Germany\\
 \textsuperscript{2}University of Augsburg, Germany\\
 \textsuperscript{3}ScaDS.AI \& TU Dresden, Germany
\\
    \small{
        \{\href{mailto:valentin.knappich@de.bosch.com}{valentin.knappich},\href{mailto:anna.haetty@de.bosch.com}{anna.haetty}\}@de.bosch.com,
        \href{mailto:simon.razniewski@tu-dresden.de}{simon.razniewski@tu-dresden.de},
        \href{mailto:annemarie.friedrich@uni-a.de}{annemarie.friedrich@uni-a.de}
    }
}

\begin{document}
\maketitle

\input{sections/00_abstract}
\input{sections/01_introduction}

\input{sections/02_related_work}

\input{sections/03_dataset}

\input{sections/04_method}
\input{sections/05_experiments}
\input{sections/06_human_eval}
\input{sections/07_conclusion}
\input{sections/08_limitations_and_acknowledgements}

\bibliography{zotero}
\bibliographystyle{acl_natbib}

\input{sections/09_appendix}

\end{document}

%% file: sparql_style.tex
\usepackage{listings}
\usepackage{xcolor}

\definecolor{codegreen}{rgb}{0,0.6,0}
\definecolor{codegray}{rgb}{0.5,0.5,0.5}
\definecolor{codepurple}{rgb}{0.58,0,0.82}
\definecolor{backcolour}{rgb}{0.95,0.95,0.92}

\lstdefinelanguage{SPARQL}{
  morekeywords=[1]{SELECT, WHERE, FILTER, OPTIONAL, UNION, GRAPH, LIMIT, OFFSET},
  morekeywords=[2]{rdf:type},
  morekeywords=[3]{xsd:integer},
  morestring=[b][\color{blue}]",
  sensitive=false,
  morecomment=[l][\color{codegreen}]{\#},
  morecomment=[s][\color{codegreen}]{/*}{*/},
}

\newcommand{\sparqlbasicstyle}{\small\ttfamily}

\lstdefinestyle{sparqlstyle}{
  backgroundcolor=\color{backcolour},
  commentstyle=\color{codegreen},
  keywordstyle=[1]\color{blue},
  keywordstyle=[2]\color{codepurple},
  keywordstyle=[3]\color{orange},
  numberstyle=\tiny\color{codegray},
  basicstyle=\sparqlbasicstyle,
  breakatwhitespace=false,
  breaklines=true,
  captionpos=b,
  keepspaces=true,
  numbers=left,
  numbersep=5pt,
  showspaces=false,
  showstringspaces=false,
  showtabs=false,
  tabsize=2,
  extendedchars=true,
  inputencoding=utf8,
  literate={á}{{\'a}}1 {ã}{{\~a}}1 {é}{{\'e}}1 {β}{{$\beta$}}1 {–}{{-}}1 {≤}{{$\leq$}}1 {≥}{{$\geq$}}1 {’}{{'}}1 {‘}{{'}}1,
  xleftmargin=2em
}

%% file: markdown_style.tex
\lstdefinelanguage{markdown}{
  basicstyle={\small\ttfamily},
  breaklines=true,
  morekeywords={},
  keywordstyle=\bfseries,
  morestring=[b]",
  stringstyle=\itshape,
  morecomment=[l][\color{blue}]{\#},
  identifierstyle=\bfseries,
  sensitive=false,
}

\lstdefinestyle{markdownstyle}{
  backgroundcolor=\color{backcolour},
  commentstyle=\color{codegreen},
  keywordstyle=[1]\color{blue},
  keywordstyle=[2]\color{codepurple},
  keywordstyle=[3]\color{orange},
  numberstyle=\tiny\color{codegray},
  basicstyle={\small\ttfamily},
  breakatwhitespace=false,
  breaklines=true,
  captionpos=b,
  keepspaces=true,
  numbers=left,
  numbersep=5pt,
  showspaces=false,
  showstringspaces=false,
  showtabs=false,
  tabsize=2,
  xleftmargin=2em
}

%% file: commands.tex
\setlist[description]{leftmargin=\parindent,itemsep=0mm,labelindent=0pt,topsep=0pt,partopsep=0ex,parsep=0ex}

\definecolor{anne}{rgb}{0,0.5,0.9}
\definecolor{valentin}{rgb}{0.998,0.722,0.635}
\definecolor{anna}{rgb}{0,0.5,0.9}
\definecolor{simon}{rgb}{0.998,0.722,0.635}

\definecolor{darkgreen}{rgb}{0.0, 0.5, 0.0}

\ifreviewcopy
    \paperwidth=\dimexpr \paperwidth + 20cm\relax
    \oddsidemargin=\dimexpr\oddsidemargin + 10cm\relax
    \evensidemargin=\dimexpr\evensidemargin + 10cm\relax
    \marginparwidth=\dimexpr \marginparwidth + 10cm - \marginparsep\relax

    \newcommand{\valentintodo}[2][valentin]{\todo[color=#1,size=\footnotesize]{\textbf{VK:} #2}}
    \newcommand{\annetodo}[2][anne!30]{\todo[color=#1,size=\footnotesize]{\textbf{AF:} #2}}
    \newcommand{\annatodo}[2][anna]{\todo[color=#1,size=\footnotesize]{\textbf{AH:} #2}}
    \newcommand{\simontodo}[2][simon!30]{\todo[color=#1,size=\footnotesize]{\textbf{SR:} #2}}

    \newcommand{\valentintodofigure}[2][0cm]{\marginnote{\todo[color=valentin,size=\footnotesize,inline]{\textbf{VK:} #2}}[#1] }
    \newcommand{\annetodofigure}[2][0cm]{\marginnote{\todo[color=anne!30,size=\footnotesize,inline]{\textbf{AF:} #2}}[#1] }
    \newcommand{\annatodofigure}[2][0cm]{\marginnote{\todo[color=anna,size=\footnotesize,inline]{\textbf{AH:} #2}}[#1] }
    \newcommand{\simontodofigure}[2][0cm]{\marginnote{\todo[color=simon!30,size=\footnotesize,inline]{\textbf{SR:} #2}}[#1] }

\else
    \newcommand{\valentintodo}[2][valentin]{}
    \newcommand{\annetodo}[2][anne!30]{}
    \newcommand{\annatodo}[2][anna]{}
    \newcommand{\simontodo}[2][simon!30]{}

    \newcommand{\valentintodofigure}[2][0cm]{}
    \newcommand{\annetodofigure}[2][anne!30]{}
    \newcommand{\annatodofigure}[2][anna]{}
    \newcommand{\simontodofigure}[2][simon!30]{}

\fi

\pgfplotsset{compat=1.16}

%% file: sections/00_abstract.tex
\begin{abstract}
    Dealing with long and highly complex technical text is a challenge for Large Language Models (LLMs), which still have to unfold their potential in supporting expensive and time-intensive processes like patent drafting.
    Within patents, the description constitutes more than 90\% of the document on average.
    Yet, its automatic generation remains understudied.
    When drafting patent applications, patent attorneys typically receive invention reports (IRs), which are usually confidential, hindering research on LLM-supported patent drafting.
    Often, pre-publication research papers serve as IRs.
    We leverage this duality to build \DatasetName, an open and realistic benchmark for patent drafting consisting of 1.8k patent-paper pairs describing the same inventions.
    To address the complex long-document patent generation task, we propose chunk-based outline-guided generation using the research paper as technical specification of the invention.
    Our extensive evaluation using \DatasetName and a human case study show that LLMs can effectively leverage information from the paper, but still struggle to provide the necessary level of detail. 
    Fine-tuning leads to more patent-style language, but also to more hallucination. 
    We release our data and code at \url{https://github.com/boschresearch/Pap2Pat}.
\end{abstract}

%% file: sections/01_introduction.tex
\section{Introduction}

\begin{figure}[!t]
  \centering
  \includesvg[width=1.0\linewidth]{img/dataset_creation_2} %
  \caption{
    \DatasetName dataset creation (left) and experimental setup (right). 
  }
  \label{fig:dataset_creation}
\end{figure}

Securing intellectual property is a long and costly process that requires both deep technical knowledge and expertise in patent law.
This motivates the use of technology to boost patent attorney productivity.
Natural language processing (NLP) already assists prior art search \cite{shalabyPatentRetrievalLiterature2019,stamatisEndEndNeural2022,pujariMultitaskApproachNeural2021} and patent landscaping \citep{CHOI2022121413,pujariThreeRealWorldDatasets2022}.
Research on Large Language Models (LLMs) in the patent domain has recently gained momentum \cite{shomeeComprehensiveSurveyAIbased2024,jiangArtificialIntelligenceExploring2024,casolaSummarizationSimplificationGeneration2022,wangIPEvalBilingualIntellectual2024}, but patent drafting remains a largely manual task.

A patent typically consists of claims, which define the invention and
the legally relevant scope of protection, and a description, which provides technical details in sections like \textit{Field of the Invention}, \textit{Background}, \textit{Summary}, and \textit{Detailed Description}.
Prior work has primarily focused on generating abstracts and claims \cite{Hamborg2017Autom-41994,leePatentTransformer2ControllingPatent2020,christofidellisPGTPromptBased2022,leeEvaluatingGenerativePatent2023,zuoPatentEvalUnderstandingErrors2024,leeInstructPatentGPTTrainingPatent2024,baiPatentGPTLargeLanguage2024}.
The description makes up over 90\% of the document,
\footnote{Measured on our dataset using the Llama-3 tokenizer: 0.7\% abstract, 91.8\% description, 7.5\% claims.}
implying that large productivity gains are expected from writing support for this section.
Yet, generating them remains a significant challenge for LLMs due to their length, technical complexity, and specialized language
\cite{wangPatentformerNovelMethod2024,wangAutoPatentMultiAgentFramework2024}.
Existing work on automatic patent generation suffers from a number of shortcomings, such as ill-posed task setups, lack of open benchmarks, and disregard for patent descriptions \citep{jiangArtificialIntelligenceExploring2024}.
In this paper, we tackle all of these issues: we propose a novel setup, evaluation metrics, and outline-guided models for description generation %
in a realistic setting using open and human-created data.

Inventors typically submit invention reports (IRs), which patent attorneys formalize into patent applications.
In many research labs, it is common to use a pre-publication paper as invention report, which leads to so-called \textit{patent-paper pairs} (PPPs, \citet{murrayInnovationCoevolutionScientific2002}), an unrecognized treasure for AI research. %
To facilitate the study of LLMs on patent drafting, we create \DatasetName, a new benchmark of 1.8k carefully identified PPPs and patent outlines.
We develop and validate a method for reliably matching patents and papers describing the same invention (see \Cref{fig:dataset_creation}).
For LLM-supported patent drafting, we envision a practical setting in which the attorney, given a paper, %
provides an outline for the patent.
This outline acts as a flexible mechanism to control the document structure and content while keeping manual effort low.

A major challenge of the proposed task is document length: patent descriptions in \DatasetName are on average 18k tokens long, some exceeding 180k tokens.
While current LLMs increasingly support long context windows, they struggle to generate similarly long outputs \cite{liuLongGenBenchLongcontextGeneration2024,baiLongWriterUnleashing102024,wuLongGenBenchBenchmarkingLongForm2025,yeLongProcBenchmarkingLongContext2025}.
As a remedy, we propose \textit{chunk-based outline-guided patent generation} (\MethodName), which effectively generates long patent documents in chunks.

Evaluating generated patents poses significant challenges due to their length, their technical complexity, and the high cost of manual evaluation.
No standard evaluation metrics are established in the literature, and prior work \cite{wangAutoPatentMultiAgentFramework2024,wangPatentformerNovelMethod2024} commonly resorts to standard text similarity metrics which do not work well on long documents \cite{lattimerFastAccurateFactual2023,queHelloBenchEvaluatingLong2024}. 
We adapt a suite of metrics based on natural language inference (NLI) and authorship attribution to the specific case of evaluating factual correctness, coverage, and language style for long-form patent generation on \DatasetName.
Our main contributions are:
\begin{enumerate}[label={(\arabic*)},itemsep=0mm,wide,labelindent=0pt,topsep=0pt,partopsep=0ex,parsep=0ex]
  \item We create and release \DatasetName, a new benchmark for patent drafting based on open data that closely aligns with real-world settings.

  \item We derive a comprehensive suite of automatic evaluation metrics for evaluating generated patents.

  \item We propose a chunk-based outline-guided patent generation approach with effectively controllable output length. 
  Our method increases coverage considerably while keeping factuality high.

  \item We conduct extensive evaluations finding that state-of-the-art LLMs can effectively use information from the papers, but still struggle to provide the necessary level of detail,
  and that fine-tuning leads to much higher stylistic similarity with patents, but substantially decreased factuality.

  \item Our human evaluation confirms these findings and indicates promising potential for productivity gains of patent attorneys.
\end{enumerate}

%% file: sections/02_related_work.tex
\section{Related Work}

\noindent\textbf{Patent-Paper Pairs (PPPs). }
Many research labs practice concurrent patenting and academic publishing.
\citet{murrayFormalIntellectualProperty2005} find that almost 50\% of their sampled academic papers from \textit{Nature Biotechnology} have a corresponding US patent.
In economics, PPPs have been used to study innovation dynamics, like whether patenting promotes or hinders the free flow of innovations \cite{murrayFormalIntellectualProperty2005,magermanSearchAnticommonsPatentpaper2011}.
In that context, several approaches to finding PPPs have been proposed:
\citet{murrayInnovationCoevolutionScientific2002} and \citet{murrayFormalIntellectualProperty2005} identify pairs manually by analyzing their full texts and citation networks.
\citet{magermanExploringFeasibilityAccuracy2010} and \citet{vanlooyAssessmentLatentSemantic2011} explore several data mining techniques.
We refine and extend their approach, and add criteria for author overlaps, date ranges, competing candidates, and licenses.
\citet{gansPatentsPapersPairs2008} propose a taxonomy of PPPs that includes both to 1-to-1 matches and m-to-n matches.
To ensure that papers are a solid source of information about the invention, we design our matching procedure to find only 1-to-1 matches.
We publish, to the best of our knowledge, the first PPPs dataset for NLP research.

\begin{figure*}
  \small
  \setlength{\tabcolsep}{10pt}
  \begin{tabularx}{\textwidth}{p{.95\textwidth} p{0\textwidth}}
    \centering
    \begin{tabular}[t]{ccccccc}
      \toprule
      \multirow{2}{*}{\textbf{Split}} & \multirow{2}{*}{\textbf{\# pairs}} & \multirow{2}{*}{\textbf{\# patent tokens}} & \multirow{2}{*}{\textbf{\# paper tokens}} & \multicolumn{3}{c}{\textbf{\# outline bullets}}                                       \\
      \cmidrule(lr){5-7}
                                      &                                    &                                            &                                           & short                                           & medium          & long              \\
      \midrule
      train                           & 1000                               & 17.8k $\pm$ 15.1k                      & 7.9k $\pm$ 4.5k                       & 36.8 $\pm$ 29.2                                 & 73.5 $\pm$ 60.0 & 149.0 $\pm$ 122.0 \\
      val                             & 242                                & 18.2k $\pm$ 16.3k                      & 8.0k $\pm$ 4.1k                       & 37.4 $\pm$ 30.8                                 & 74.4 $\pm$ 62.9 & 150.6 $\pm$ 127.5 \\
      test                            & 500                                & 18.1k $\pm$ 13.2k                      & 8.1k $\pm$ 3.9k                       & 37.5 $\pm$ 24.7                                 & 74.9 $\pm$ 50.9 & 151.6 $\pm$ 103.7 \\
      nc-test                         & 71                                 & 18.4k $\pm$ 13.9k                      & 9.5k $\pm$ 4.6k                       & 37.8 $\pm$ 22.7                                 & 76.0 $\pm$ 46.4 & 154.4 $\pm$ \phantom{0}94.8  \\
      \midrule
      all                             & 1813                               & 17.9k $\pm$ 14.7k                      & 8.1k $\pm$ 4.3k                       & 37.1 $\pm$ 28.0                                 & 74.1 $\pm$ 57.5 & 150.1 $\pm$ 117.0 \\
      \bottomrule
    \end{tabular}
    \captionof{table}{\DatasetName statistics. Values are reported as mean $\pm$ std. Token counts correspond to the Llama-3 tokenizer.}
    \label{tab:dataset_stats} &
  \end{tabularx}
\end{figure*}

\noindent\textbf{Patent Generation. }
Most prior work on patent generation has focused on titles, abstracts, and claims.
\citet{christofidellisPGTPromptBased2022} train a multi-task GPT2 model to generate these parts.
\citet{leeEvaluatingGenerativePatent2023} pre-trains a GPT-J-6B architecture on entire patents and evaluates it on claim generation.
\citet{zuoPatentEvalUnderstandingErrors2024} use GPT-3.5-turbo and Llama-2 to generate abstracts from claims, and claims from previous claims.
\citet{wangPatentformerNovelMethod2024} experiment with the generation of individual description paragraphs based on patent claims and drawing descriptions.
In practice, it is not likely that claims are already finalized at the time of writing of the description section.
In their work concurrent to ours, \citet{wangAutoPatentMultiAgentFramework2024} leverage an agent framework to generate complete patents based on purely automatically generated invention specifications.
They share our core principle of using a divide-and-conquer strategy for long-document generation, but evaluate only using surface-level metrics.
Our more realistic setting relies on real-world invention specifications and provides deeper insights due to more sophisticated evaluation metrics.

\noindent\textbf{Outline-guided Generation}
is a paradigm in which LLMs use an outline to produce longer, more structured and coherent text.
Outlines are either generated in a planning stage 
\cite{drissiHierarchicalTextGeneration2018,yaoPlanandWriteBetterAutomatic2019,sunSummarizeOutlineElaborate2022a,yang-etal-2022-re3,leeNavigatingPathWriting2024,wangAutoSurveyLargeLanguage,shaoAssistingWritingWikipedialike2024} 
or provided as input 
\cite{fangOutlineStoryFinegrained2021a,spangher-etal-2022-sequentially,yangDOCImprovingLong2023,liAdvancingPreciseOutlineConditioned2023}.
They are created using extraction of key words 
\cite{yaoPlanandWriteBetterAutomatic2019}, 
phrases 
\cite{fangOutlineStoryFinegrained2021a}, 
or sentences 
\cite{drissiHierarchicalTextGeneration2018,sunSummarizeOutlineElaborate2022a,yang-etal-2022-re3,liAdvancingPreciseOutlineConditioned2023}, 
generated using LLMs 
\cite{yangDOCImprovingLong2023} or 
defined interactively 
\cite{goldfarb-tarrant-etal-2019-plan}.
In our work, we posit that outlines should be provided by the patent attorney to satisfy the high demand for user control. %

%% file: sections/03_dataset.tex
\section{\DatasetName{} Benchmark}

We present the \DatasetName dataset containing 1.8k PPPs, each annotated with three outlines for generation.
We describe the steps taken to construct \DatasetName, analyse the obtained corpus, and propose evaluation metrics for our new benchmark.

\subsection{Dataset Construction}
We here give an overview of the construction of \DatasetName (for details, see \Cref{sec:dataset_appendix}).

\noindent\textbf{Scraping PPPs. }
The patent and paper in a PPP typically do not cite each other, so we cannot rely on front-page or in-text citations \cite{marxRelianceScienceWorldwide2020, marxRelianceScienceInventors2022} to find PPPs.
Therefore, prior work has developed heuristics to match patents and papers based on document metadata \cite{magermanExploringFeasibilityAccuracy2010,vanlooyAssessmentLatentSemantic2011}.
We use the USPTO dataset\footnote{\url{https://bulkdata.uspto.gov}} containing 6.7M patent applications from 2005 to April 2024.
For each patent, we query SemOpenAlex \cite{farberSemOpenAlexScientificLandscape2023a} using SPARQL and retrieve papers with overlapping authors lists and publication dates.
We filter the results based on titles, abstracts, other candidate matches for the same patent, and paper licenses.
\Cref{tab:dataset_filters} shows the remaining number of candidates after each filtering step.
A major limiting factor for our benchmark size are the restrictive licenses of many scientific articles.\footnote{ACL with its CC-BY license being a positive exception.}

\noindent\textbf{Manual Validation. }
To verify the precision of our matching pipeline, the first author performs a manual validation.
We randomly sample 60 PPPs, read both abstracts, skim the documents, compare the figures and get an overview of the authors' related work. 
We spend a total of 5 hours, i.e., 5 minutes per pair on average.
In 55/60 ($91.7\%$) pairs, the paper describes the invention as a core contribution.
In three pairs, the best match for the patent would have been a prior paper by the same authors.
In two pairs, the paper would have been best matched to a related but different patent by the same inventors.
In these five imperfect matching cases, the papers still provide meaningful training and evaluation signals, as they are highly relevant to the invention and contextualized by the outline.
Overall, the precision of our matching approach is high.

\noindent\textbf{Outline Generation. }
Outlines consist of target headings and short bullet points summarizing the document structure and high-level content of every section (see \Cref{fig:generation_approach} and \Cref{sec:patent_outline_appendix}).
For \DatasetName, we generate them automatically from the original patents using Llama-3 70B \cite{dubeyLlamaHerdModels2024}.
In practice, they are provided by the user.
To ensure that the model adheres to the desired output format, we use SGLang \cite{zhengEfficientlyProgrammingLarge2023} for constrained decoding.
We enforce a fixed number of bullet points
$n_\text{bullets}$ per section, where $n_\text{bullets}$ is proportional to the length of the text in that section ($n_\text{chars}$) and defined as
\begin{align*}
  n_\text{bullets} = \begin{cases}
                       \text{max}(1, \lfloor n_\text{chars}/l \rfloor) & \text{if }n_\text{chars} > 0 \\
                       0                                               & \text{else}
                     \end{cases}
\end{align*}
\noindent where $l$ is the number of characters that each bullet point summarizes on average. We create outlines with three levels of granularity: \textit{long} ($l$=500, avg.\ 150 items), \textit{medium} ($l$=1000, avg.\ 74 items) and \textit{short} ($l$=2000, avg.\ 37 items), see \Cref{tab:dataset_stats}.
The average bullet point length is 5.4 $\pm$ 2.3 words.
For each section, we additionally provide the number of characters in the original patent to signal the desired content lengths during generation.

\begin{table}[t]
  \centering
  \setlength{\tabcolsep}{10pt}
  \begin{tabular}[t]{lc}
    \toprule
      Filter                              & \# pairs \\
      \midrule
      \makecell[tl]{Authors + Date}       & 930k       \\
      \makecell[tl]{+ Term Overlap}       & 100k       \\
      \makecell[tl]{+ Distinctiveness}    & 21k        \\
      \makecell[tl]{+ Permissive License} & 1.8k       \\
      \bottomrule
  \end{tabular}
  \captionof{table}{Filter criteria and remaining number of candidates.}
  \label{tab:dataset_filters}
\end{table}

\subsection{Task Description and Data Splits}
We propose the following generation task for \DatasetName: Given a patent outline $O$ and a research paper $C$
containing a specification of the invention, models should output a patent $P$. 
The input data provides the desired output length in characters per section.
We split our dataset randomly into \textit{train} (n=1000), \textit{test} (n=500) and \textit{validation} (n=242), see \Cref{tab:dataset_stats}.
We additionally create a non-contaminated test set (\textit{nc-test}) that contains all pairs with patents published in 2024 (n=71), i.e., after the pretraining cut-off date of all evaluated open-weight LLMs, addressing the concern that LLMs might have seen test data during pretraining \cite{ravaut2024much}.

\subsection{Corpus Statistics and Analysis}
Dataset statistics are presented in \Cref{tab:dataset_stats} (further statistics and plots in \Cref{sec:dataset_stats_appendix}).
Papers typically include details and analyses regarding experiments;
patents usually contain more information on applications and practical benefits.
We analyze the lexical overlap between the respective documents and find that only 8.3\% of the 4-grams are shared.
This highlights the complexity of the task: 
the two documents describe the same invention from different perspectives, using different linguistic styles.

\subsection{Proposed Evaluation Metrics}\label{sec:new_metrics}

We adapt a suite of metrics to analyze the performance of patent generation models from multiple perspectives.
The length and specialized domain of patent documents make automatic evaluation challenging.
We thus propose to disentangle factual content overlap from stylistic similarity using established metrics that are well-suited for long documents. 

\subsubsection{Content-level metrics}

To assess the factual consistency and coverage of the generated patent, we measure semantic overlap with the reference patent and the provided paper.
The NLI-based metric SCALE \cite{lattimerFastAccurateFactual2023} estimates factual consistency between two documents by computing pairwise entailment scores between \textit{premise chunks} and \textit{hypothesis sentences}.
While premise chunks can be large, SCALE still requires a quadratic computation.
To make this feasible on long documents, we sample 10 sentences from every hypothesis document.
For each sentence, we retrieve the 5 most relevant premise chunks according to BM25 and then compute the maximum NLI score across these chunks.
For system-level scores, we average the scores obtained for all sampled sentences.
For \DatasetName, we propose two variants of SCALE: %

\begin{description}[topsep=7pt,itemsep=7pt]
    \item[Factuality.] \textit{Ref} \textrightarrow{} \textit{Gen} and \textit{Ref+Pap} \textrightarrow{} \textit{Gen} quantify to what extent the semantic content of the generated patent (\textit{Gen}) is supported by the reference patent (\textit{Ref}), and by the reference patent and the paper (\textit{Ref+Pap}), respectively. %
    \item[Coverage.] To measure the degree to which the generated patent covers the information content of the reference patent, we use the generated patent as premise document and the reference patent as the hypothesis (\textit{Gen} \textrightarrow{} \textit{Ref}). %
\end{description}

\noindent We confirm the applicability of these scores to our benchmark by computing a few trivial baselines (see \Cref{tab:results}): taking the reference patent itself as the hypothesis document results in an upper bound of around 89\%; using the most similar patent from the train set according to BM25 similarity to the paper
results in a score of less than 30\%.
Taken together with the low standard deviations across 5 executions, this demonstrates that the score is able to differentiate and that the sample size is sufficient to make meaningful system-level comparisons.

\begin{figure*}[t]
  \centering
  \includesvg[width=\linewidth]{img/generation_method}
  \caption{
    Overview of chunk-based outline-guided patent generation (\MethodName). 
    We chunk the outline, retrieve the parts of the paper that are most relevant for that chunk, prompt the LLM, and concatenate the results.
    The desired output length and the number of allocated patent tokens per chunk determine the number of chunks.
  }
  \label{fig:generation_approach}
\end{figure*}

\subsubsection{Language-level metrics}

Patents are written in special \textbf{style}, including the use of partially legally relevant phrases and constructions. 
To estimate to what extent the generated patents adhere to this style, we compare corpus-wide n-gram profiles, an established method for authorship attribution \cite{kesVeljNGRAMBASEDAUTHORPROFILES2003,Zecevic2011NgramBT,Dam2013ABC,Mikros2013AuthorshipAI}.
Here, we use the set of all reference patents from the same split as basis to extract the profile of a typical patent attorney $P^n_\text{ref}$ and compare it to the profile extracted from the generated patents $P^n_\text{gen}$.
Each profile contains the 1000 most frequent n-grams for $n \in \{1,2,3,4\}$.
To compute the similarity between $P^n_\text{ref}$ and $P^n_\text{gen}$, we adopt the formula from \citet{kesVeljNGRAMBASEDAUTHORPROFILES2003}:
\[sim_n = \sum_{g \in P^n_\text{ref} \cup P^n_\text{gen}} 1 - \left(\frac{P^n_\text{ref}(g) - P^n_\text{gen}(g)}{P^n_\text{ref}(g) + P^n_\text{gen}(g)}\right)^2\]
We additionally integrate a profile from StyloMetrix\footnote{\href{https://github.com/ZILiAT-NASK/StyloMetrix}{https://github.com/ZILiAT-NASK/StyloMetrix}} \cite{okulskaStyloMetrixOpenSourceMultilingual2023},
which implements rule-based detection of 196 linguistic features, including e.g. verb tenses, modal verbs, POS tags, lexical items, figures of speech, and linguistic constructions such as fronting or similes.
We compute the final score as the average of the four n-gram profile similarities and the StyloMetrix profile similarity, i.e., $Style = \frac{1}{5} ((\sum_{i=1}^4 {sim_n}) + sim_{stylo})$. 

To measure the \textbf{repetitiveness}, we compute the \textit{repetition rate} RR \cite{cettoloRepetitionRateText2014}, i.e., the fraction of n-grams that appear more than once.
We compute RR over sliding windows of 256 tokens and average the scores.
To differentiate between repetitive language and unintended infinite repetitions, we additionally report the fraction of windows with an RR score greater than 80 (RR>80).

%% file: sections/04_method.tex
\section{\MethodName: Chunk-based Outline-guided Patent Generation}

Since LLMs cannot yet generate sufficiently long documents in a single call, we instead choose to generate patents in chunks. 
\Cref{fig:generation_approach} summarizes the approach.
We chunk the outline, select the most relevant parts of the paper (\textit{paper context})
depending on the chunks' outline bullet points, and prompt an LLM to generate the patent text for that chunk.
The outlines of previous chunks are included for global document context.
Finally, we concatenate the generated outputs and apply only a lightweight post-processing to remove duplicate headings at chunk boundaries.
This framework enables generating a long document in parallel, with customized context per chunk and controllable length.

\input{sections/main_table}

\noindent\textbf{Token Allocation and Chunking. } 
In the default setting, we allocate per chunk 2k tokens for the instruction, 3k tokens for the paper context, and 2k tokens for the output patent ($\mathcal{T}$\textit{$\{$inst=2k; pap=3k; pat=2k$\}$}).
We choose this setting because preliminary experiments show that on our task, LLMs only generate up to roughly 2-3k tokens regardless of the requested amount of content,
and because this allows comparing with models that support up to 8k tokens.
The chunking procedure segments the outline into chunks depending on the token allocation.
In particular, it determines the number of outline bullet points per chunk using the reserved number of patent tokens and the average number of characters per outline bullet point\footnote{We translate between number of tokens and number of characters using average corpus statistics of patents in \DatasetName.}.
It keeps sections intact whenever possible.

\noindent\textbf{Length Control Mechanism. }
\citet{baiLongWriterUnleashing102024} find that LLMs' response lengths are constrained within a certain range and only moderately adjust to length requests in the prompt, making instructions unreliable for controlling length. 
However, as the requested length decreases, the LLM is more likely to meet it.
Hence, in \MethodName, we decrease the number of allocated patent tokens per chunk, thereby increasing the number of chunks, until the desired total length is reached.
In our experimental setting, we do this on a corpus-level, i.e., we search for a setting where the average length matches that of the references, leading to the calibrated token allocation $\mathcal{T}$\textit{$\{$inst=2k; pap=3k; pat=400$\}$}.

\noindent\textbf{Paper Context Selection. }
For each chunk $i$, the retriever selects relevant paragraphs from the paper to create the paper context $\text{c}_i$.
We use BM25 \cite{robertsonProbabilisticRelevanceFramework2009} with the chunk's outline $\text{o}_i$ as query.
We always include the abstract of the paper and all headings, adding paper paragraphs successively %
in the order of their relevance ranking until the token limit is reached.

\noindent\textbf{Patent Generation. } 
The paper context $\text{c}_i$, the current outline $\text{o}_i$ and prior outlines $\text{o}_{j<i}$ are combined to a prompt (see Appendix \ref{sec:generation_prompt_appendix}). 
The LLM generates patent chunk $\text{p}_i$, using constrained decoding to adhere to the outline's headings.

%% file: sections/main_table.tex
\begin{table*}[t]
    \centering
    \small
    \resizebox{\textwidth}{!}{
      \setlength{\tabcolsep}{4pt}
      \begin{tabular}{lccccccccccc}
        \toprule
                              & \multirow{4}{*}{Tokens}
                              & \multicolumn{2}{c}{\multirow{3}{*}{Text Sim $\uparrow$}}
                              & \multicolumn{3}{c}{Content-Level Metrics (SCALE) $\uparrow$}
                              & \multicolumn{2}{c}{\multirow{3}{*}{Language $\uparrow$}}
                              & \multicolumn{2}{c}{\multirow{3}{*}{Repetitions}}
        \\
        \cmidrule(lr){5-7}
                              &
                              &
                              &
                              & Coverage
                              & \multicolumn{2}{c}{Factuality}
                              &
                              &
                              &
                              &
                              &
        \\
        \cmidrule(lr){3-4}\cmidrule(lr){6-7}\cmidrule(lr){8-9}\cmidrule(lr){10-11}
                              &
                              & BS
                              & R-L
                              & \textit{Gen}\textrightarrow{}\textit{Ref}
                              & \textit{Ref}\textrightarrow{}\textit{Gen}
                              & \textit{Ref}+\textit{Pap}\textrightarrow{}\textit{Gen}
                              & Style
                              & DS
                              & RR
                              & RR>80
        \\
        \midrule
    
        \multicolumn{12}{c}{\textbf{Heuristic Baselines / Skylines}}\vspace{.5em}                                                                                                      \\
        Reference Patent
                              & \textit{18.1k (100\%)}
                              & \textit{100}
                              & \textit{100}
                              & \textit{88.6 $\pm$ .15}
                              & \textit{88.5 $\pm$ .18}
                              & \textit{88.7 $\pm$ .18}
                              & \textit{100}
                              & \textit{100}
                              & 14.4
                              & 0.2
        \\
        Similar Patent
                              & 30.1k (166.0\%)
                              & 65.8
                              & 33.7
                              & 29.4 $\pm$ .37
                              & 27.6 $\pm$ .32
                              & 27.6 $\pm$ .36
                              & 75.3
                              & 98.3
                              & 12.3
                              & 0.1
        \\
        Outline
                              & 1.4k (7.9\%)
                              & 56.5
                              & 10.1
                              & 39.7 $\pm$ .72
                              & 61.6 $\pm$ .14
                              & 61.9 $\pm$ .15
                              & 24.4
                              & 85.6
                              & 20.1
                              & 0.2
        \\
        Paper
                              & 8.1k (44.5\%)
                              & 69.7
                              & 39.4
                              & 44.8 $\pm$ .61
                              & 46.5 $\pm$ .40
                              & \textit{89.0 $\pm$ .13}
                              & 47.2
                              & 98.4
                              & 8.4
                              & 0.0
        \\
        \cmidrule{1-11}
        \multicolumn{12}{c}{\textbf{Single LLM-call} ($\mathcal{T}$\textit{$\{$inst=2k; pap=$\infty$; pat=$\infty\}$})}\vspace{.5em}                                                                                                          \\
        Mixtral-8x7B
                              & 3.2k (17.7\%)
                              & 66.1
                              & 23.1
                              & 38.1 $\pm$ .70
                              & \textbf{66.9 }$\pm$ .87
                              & 72.0 $\pm$ .71
                              & \textbf{\itshape 42.6}
                              & 96.9
                              & 17.9
                              & 0.6
        \\
        Qwen2-72B
                              & 2.8k (15.6\%)
                              & \textbf{\itshape 66.6}
                              & 21.3
                              & \textbf{\itshape 40.3 }$\pm$ .45
                              & 65.8 $\pm$ .47
                              & 72.4 $\pm$ .43
                              & 39.6
                              & 97.3
                              & 9.5
                              & 0.5
        \\
        $\quad$w/o Paper
                              & 3.5k (19.3\%)
                              & 66.1
                              & \textbf{\itshape 23.2}
                              & 38.9 $\pm$ .29
                              & 65.9 $\pm$ .54
                              & 66.6 $\pm$ .38
                              & 37.1
                              & 96.6
                              & 9.8
                              & 0.6
        \\
        $\quad$w/o Outline
                              & 2.0k (11.1\%)
                              & 64.2
                              & 16.9
                              & 34.9 $\pm$ .40
                              & 56.7 $\pm$ .31
                              & \textbf{75.3 }$\pm$ .23
                              & 39.8
                              & \textbf{\textit{97.4}}
                              & 9.3
                              & 0.1
        \\
    
        \cmidrule{1-11}
        \multicolumn{12}{c}{\textbf{\MethodName} ($\mathcal{T}$\textit{$\{$inst=2k; pap=3k; pat=2k$\}$})}\vspace{.5em}                                                                                \\
        Llama-3 8B
                              & 9.6k (53.1\%)
                              & 68.7
                              & 41.4
                              & 40.3 $\pm$ .21
                              & 60.8 $\pm$ .49
                              & 65.7 $\pm$ .55
                              & 43.2
                              & 97.0
                              & 27.0
                              & 4.4
        \\
        Llama-3 8B SFT
                              & 27.5k (151.5\%)
                              & 70.4
                              & 42.8
                              & 42.0 $\pm$ .25
                              & 49.3 $\pm$ .39
                              & 52.1 $\pm$ .39
                              & 59.4
                              & 98.0
                              & 53.7
                              & 29.3
        \\
        $\quad$w/ rep. removal
                              & 17.3k (95.4\%)
                              & \textbf{\itshape 71.2}
                              & \textbf{\itshape 49.6}
                              & 42.1 $\pm$ .48
                              & 49.6 $\pm$ .38
                              & 53.4 $\pm$ .59
                              & \textbf{64.7}
                              & \textbf{98.5}
                              & 38.4
                              & 8.5
        \\
        Mixtral-8x7B
                              & 5.6k (31.0\%)
                              & 69.1
                              & 35.5
                              & 41.8 $\pm$ .61
                              & 62.3 $\pm$ .31
                              & 68.5 $\pm$ .30
                              & 49.3
                              & 97.5
                              & 14.5
                              & 0.4
        \\
        Llama-3 70B
                              & 6.1k (33.9\%)
                              & 70.2
                              & 39.0
                              & 42.7 $\pm$ .60
                              & \textbf{\itshape 64.5 }$\pm$ .47
                              & \textbf{\itshape 68.6 }$\pm$ .58
                              & 49.5
                              & 97.4
                              & 17.5
                              & 0.2
        \\
        Qwen2-72B
                              & 8.1k (44.8\%)
                              & 70.2
                              & 41.3
                              & \textbf{\itshape 44.1 }$\pm$ .32
                              & 62.5 $\pm$ .44
                              & 67.9 $\pm$ .35
                              & 47.5
                              & 97.3
                              & 12.8
                              & 1.2
        \\
        \cmidrule{1-11}
        \multicolumn{12}{c}{\textbf{\MethodName} ($\mathcal{T}$\textit{$\{$inst=2k; pap=3k; pat=400$\}$})}\vspace{.5em}                                                                                    \\
        Qwen2-72B
                              & 18.1k (100\%)
                              & \textbf{71.7}
                              & \textbf{50.8}
                              & \textbf{46.8 }$\pm$ .31
                              & 59.7 $\pm$ .63
                              & 65.3 $\pm$ .44
                              & 47.8
                              & 97.1
                              & 10.0
                              & 0.5
        \\
        \bottomrule
    \end{tabular}
    
    }
    \caption{
      Experimental Results. 
      \textit{Italic values} represent upper bounds that used test data in the prediction. 
      The best value per column is \textbf{bold}, the best per section \textbf{\itshape bold italics}.
      Tokens are reported as absolute and relative to the reference.
      BS = BERTScore. R-L = ROUGE-L. DS = DiscoScore. Text Sim = Standard Text Similarity Metrics.
    }
    \label{tab:results}
\end{table*}

%% file: sections/05_experiments.tex
\section{Experiments}

In this section, we describe our experiments using \DatasetName{} to assess the capabilities and limitations of current LLMs with \MethodName.

\subsection{Evaluation Metrics}
We primarily focus on the evaluation metrics proposed in \Cref{sec:new_metrics}.
For the sake of completeness, we also report ROUGE-L F1 \cite{linROUGEPackageAutomatic2004} and BERTScore F1 \cite{zhangBERTScoreEvaluatingText2020}.
For BERTScore, we circumvent the limit of 512 tokens with a sliding window, and use SciBERT \cite{beltagySciBERTPretrainedLanguage2019},
which has been shown to be effective in the patent domain \cite{pujariMultitaskApproachNeural2021}.
We also report DiscoScore \cite{zhao-etal-2023-discoscore}, a state-of-the-art \textbf{coherence} metric rooted in Centering Theory \cite{grosz-etal-1995-centering} and based on BERT \citep{devlin-etal-2019-bert}.
We use the DS-SENT-NN variant since the authors report that it performs best on long texts.
We extend the published implementation\footnote{\url{https://github.com/AIPHES/DiscoScore}} with a sliding window to obtain BERT embeddings.

\subsection{Models}

\noindent\textbf{Choice of LLMs. }
To make our work reproducible, we only leverage recently published open-weight LLMs with state-of-the-art results on other generation tasks.
We include Llama-3 8B and 70B \cite{dubeyLlamaHerdModels2024}, Mixtral-8x7B \cite{jiangMixtralExperts2024}, and Qwen2-72B \cite{yangQwen2TechnicalReport2024}.

\noindent\textbf{Baselines.}
We include four baselines, upper bounds, and sanity checks for the evaluation setup. 
In particular, we consider the paper, the outline, the reference patent and the most similar patent from the train set (highest BM25 similarity to the paper) as the generated document. %
Mixtral-8x7B and Qwen2-72B support 32k tokens, thus we use them to determine the performance without \MethodName by prompting them with the complete paper and outline (\textbf{Single LLM-call}).

\noindent\textbf{Fine-tuning. }
We fine-tune Llama-3 8B on the training split of \DatasetName using LoRA \cite{huLoRALowRankAdaptation2021}. 
We adopt the hyperparameters proposed by \citet{tribesHyperparameterOptimizationLarge2024}, see \Cref{sec:hyperparameter_appendix}. 
Since the fine-tuned model frequently generates infinite repetitions, we design a post-processing procedure that detects and removes them (see \Cref{sec:rep_removal}).

\subsection{Main Findings}

\Cref{tab:results} shows our main evaluation results.
Overall, we find that \textbf{\MethodName} strongly improves upon using a \textbf{Single LLM-call},
in which the generated patents are much too short (on average less than one fifth of the reference patents).
\MethodName provides an effective remedy, creating much longer and length-controllable outputs with higher coverage (correlating with higher text similarity scores).
In terms of language style, \MethodName also has a clear advantage.
For RR, values around 12-14\% %
seem to be ideal according to the scores of the reference and similar patents.

While the fine-tuned models produce more repetitions, both the Single LLM-call and zero-shot \MethodName have similar repetition rates and only the smaller and fine-tuned Llama models get stuck in infinite repetitions.
All models achieve very high coherence scores, indicating that all LLMs in the study do not suffer from generating incoherent text.

\noindent\textbf{Impact of Outline and Paper as Input.}
Ablating the paper or the outline from the prompt in the Single LLM-call baseline leads to lower coverage.
When ablating the outline, the output drops markedly in length and sticks closely to the paper (as shown by the high score for \textit{Ref+Pap$\rightarrow$Gen}), but without producing the content desired for the patent (\textit{Ref$\rightarrow$Gen} drops by almost 10pp.).
\MethodName is able to leverage the outline and paper context effectively, as demonstrated by its much higher coverage.

\noindent\textbf{Length Control and Coverage-Factuality Tradeoff.}
\Cref{fig:length_control} shows that
allocating fewer tokens per chunk for the patent, i.e.,
generating more chunks, leads to a near-linear increase in output length.
Our length control mechanism is able to find an optimal setting in which the average output length corresponds to that of the reference patents (last row of \Cref{tab:results}), resulting in the overall best text similarity and coverage scores, while keeping factuality high (as opposed to the fine-tuned models that produce similar length).
Coverage and factuality are intuitively antagonistic (see \Cref{fig:length_control}), similar to precision and recall:
as the generated text becomes longer, it becomes increasingly difficult to maintain factuality but easier to achieve higher coverage.
This also explains why the Single LLM-call baselines achieve higher factuality scores than \MethodName runs that generate more than five times as much text.

\subsection{Analysis of \MethodName Settings}

We now study various settings of \MethodName.

\noindent\textbf{Impact of Outline Granularity.}
In \Cref{fig:granularity}, we observe a consistent improvement across coverage, factuality and BERTScore with more detailed outlines:
users can directly improve output quality by providing more details. %
Notably, contrary to our experiments on length control, the improvements in coverage are achieved without longer outputs and without decreasing factuality.

\begin{figure}[t]
  \input{img/length_control.pgf}
  \caption{
    Controlling output length.
    Each point on the x-axis represents a run of Qwen2-72B with varying token allocation \mbox{$\mathcal{T}$\textit{$\{$inst=2k; pap=3k; pat=$l_{pat}\}$}} where $l_{pat}$ ranges from 200 to 10,000. 
    A lower $l_{pat}$ results in more chunks. 
    The dashed blue line represents the average reference patent length. 
  }
  \label{fig:length_control}
\end{figure}

\noindent\textbf{Impact of Selection of Paper Context.}
We study how providing informative context taken from the paper influences results.
In \Cref{fig:retrievers}, we show retriever ablation results including two baselines: 
\textit{NoPaper} does not add any context from the paper, and \textit{AbstractOnly} uses only the abstract.
As an upper bound, we use \textit{BM25Oracle,} where the BM25 query is the original patent text instead of the outline.
We observe a monotonically increasing performance, demonstrating that associated papers provide valuable information.
There is only a small gap between \textit{BM25} and \textit{BM25Oracle}, suggesting that the outline is an effective BM25 query and that more elaborate retrieval methods could close the gap further.

\noindent\textbf{Test Data Contamination. }
If patents are (partially) memorized during pre-training, one would expect a sudden drop in performance when evaluating on patents published after the pre-training cutoff date, i.e., on the non-contaminated test set.
We see only minor differences in text similarity and SCALE metrics, though the style scores drop significantly (see \Cref{sec:nc-test-results}). 
This is likely due to domain distribution shifts (see \Cref{sec:dataset_stats_appendix}).
Overall, the results do not indicate systematic issues with memorization.

\noindent\textbf{Fine-tuning. }
Fine-tuning results in significantly longer outputs, which is consistent with the findings of \citet{baiLongWriterUnleashing102024}.
However, this increased length is partially due to infinite repetitions, an issue also observed in concurrent work on patent generation by \citet{wangAutoPatentMultiAgentFramework2024}.
Prior work has identified repetitive training data as a key factor contributing to infinite repetitions \cite{liRepetitionRepetitionOut2023}.
We hence hypothesize that the cause is patents' inherently repetitive style.
Patents often present numerous variations of the invention, each introduced with similar phrasing and detailing different combinations of components.
This also results in a 71\% higher RR than for papers.
Despite these repetitions, fine-tuning improves standard text similarity metrics like ROUGE-L and BERTScore.
\citet{wangAutoPatentMultiAgentFramework2024} find the same, along with much lower scores in their human evaluation, concluding that the repetitions lead to over-rewarding in n-gram-based metrics.
We find contrary evidence that removing many repetitions from the generated patents in post-processing actually further improves ROUGE-L by 6.8pp.
Our evaluation instead reveals that the improvement achieved by fine-tuning is likely mainly due to stylistic similarity: style scores increase by over 20pp. while factuality metrics decrease by over 10pp.

\begin{figure}[t]
  \input{img/granularity.pgf}
  \caption{
    Outline granularity ablation of Qwen2-72B with default token allocation. 
  }
  \label{fig:granularity}
\end{figure}

\begin{figure}[t]
  \centering
  \resizebox{\linewidth}{!}{
    \input{img/retrievers.pgf}
  }
  \caption{Retriever ablation. The plot depicts the results of Qwen2-72B using the long outline and default token allocation.}
  \label{fig:retrievers}
\end{figure}

%% file: img/length_control.pgf
\begingroup%
\makeatletter%
\begin{pgfpicture}%
\pgfpathrectangle{\pgfpointorigin}{\pgfqpoint{3.086370in}{1.748672in}}%
\pgfusepath{use as bounding box, clip}%
\begin{pgfscope}%
\pgfsetbuttcap%
\pgfsetmiterjoin%
\definecolor{currentfill}{rgb}{1.000000,1.000000,1.000000}%
\pgfsetfillcolor{currentfill}%
\pgfsetlinewidth{0.000000pt}%
\definecolor{currentstroke}{rgb}{0.500000,0.500000,0.500000}%
\pgfsetstrokecolor{currentstroke}%
\pgfsetdash{}{0pt}%
\pgfpathmoveto{\pgfqpoint{0.000000in}{0.000000in}}%
\pgfpathlineto{\pgfqpoint{3.086370in}{0.000000in}}%
\pgfpathlineto{\pgfqpoint{3.086370in}{1.748672in}}%
\pgfpathlineto{\pgfqpoint{0.000000in}{1.748672in}}%
\pgfpathlineto{\pgfqpoint{0.000000in}{0.000000in}}%
\pgfpathclose%
\pgfusepath{fill}%
\end{pgfscope}%
\begin{pgfscope}%
\pgfsetbuttcap%
\pgfsetmiterjoin%
\definecolor{currentfill}{rgb}{0.898039,0.898039,0.898039}%
\pgfsetfillcolor{currentfill}%
\pgfsetlinewidth{0.000000pt}%
\definecolor{currentstroke}{rgb}{0.000000,0.000000,0.000000}%
\pgfsetstrokecolor{currentstroke}%
\pgfsetstrokeopacity{0.000000}%
\pgfsetdash{}{0pt}%
\pgfpathmoveto{\pgfqpoint{0.491005in}{0.693440in}}%
\pgfpathlineto{\pgfqpoint{2.609202in}{0.693440in}}%
\pgfpathlineto{\pgfqpoint{2.609202in}{1.648672in}}%
\pgfpathlineto{\pgfqpoint{0.491005in}{1.648672in}}%
\pgfpathlineto{\pgfqpoint{0.491005in}{0.693440in}}%
\pgfpathclose%
\pgfusepath{fill}%
\end{pgfscope}%
\begin{pgfscope}%
\pgfpathrectangle{\pgfqpoint{0.491005in}{0.693440in}}{\pgfqpoint{2.118198in}{0.955233in}}%
\pgfusepath{clip}%
\pgfsetrectcap%
\pgfsetroundjoin%
\pgfsetlinewidth{0.803000pt}%
\definecolor{currentstroke}{rgb}{1.000000,1.000000,1.000000}%
\pgfsetstrokecolor{currentstroke}%
\pgfsetdash{}{0pt}%
\pgfpathmoveto{\pgfqpoint{0.505936in}{0.693440in}}%
\pgfpathlineto{\pgfqpoint{0.505936in}{1.648672in}}%
\pgfusepath{stroke}%
\end{pgfscope}%
\begin{pgfscope}%
\pgfsetbuttcap%
\pgfsetroundjoin%
\definecolor{currentfill}{rgb}{0.333333,0.333333,0.333333}%
\pgfsetfillcolor{currentfill}%
\pgfsetlinewidth{0.803000pt}%
\definecolor{currentstroke}{rgb}{0.333333,0.333333,0.333333}%
\pgfsetstrokecolor{currentstroke}%
\pgfsetdash{}{0pt}%
\pgfsys@defobject{currentmarker}{\pgfqpoint{0.000000in}{-0.048611in}}{\pgfqpoint{0.000000in}{0.000000in}}{%
\pgfpathmoveto{\pgfqpoint{0.000000in}{0.000000in}}%
\pgfpathlineto{\pgfqpoint{0.000000in}{-0.048611in}}%
\pgfusepath{stroke,fill}%
}%
\begin{pgfscope}%
\pgfsys@transformshift{0.505936in}{0.693440in}%
\pgfsys@useobject{currentmarker}{}%
\end{pgfscope}%
\end{pgfscope}%
\begin{pgfscope}%
\definecolor{textcolor}{rgb}{0.333333,0.333333,0.333333}%
\pgfsetstrokecolor{textcolor}%
\pgfsetfillcolor{textcolor}%
\pgftext[x=0.505936in,y=0.596217in,,top]{\color{textcolor}{\rmfamily\fontsize{7.000000}{8.400000}\selectfont\catcode`\^=\active\def^{\ifmmode\sp\else\^{}\fi}\catcode`\%=\active\def
\end{pgfscope}%
\begin{pgfscope}%
\pgfpathrectangle{\pgfqpoint{0.491005in}{0.693440in}}{\pgfqpoint{2.118198in}{0.955233in}}%
\pgfusepath{clip}%
\pgfsetrectcap%
\pgfsetroundjoin%
\pgfsetlinewidth{0.803000pt}%
\definecolor{currentstroke}{rgb}{1.000000,1.000000,1.000000}%
\pgfsetstrokecolor{currentstroke}%
\pgfsetdash{}{0pt}%
\pgfpathmoveto{\pgfqpoint{1.035218in}{0.693440in}}%
\pgfpathlineto{\pgfqpoint{1.035218in}{1.648672in}}%
\pgfusepath{stroke}%
\end{pgfscope}%
\begin{pgfscope}%
\pgfsetbuttcap%
\pgfsetroundjoin%
\definecolor{currentfill}{rgb}{0.333333,0.333333,0.333333}%
\pgfsetfillcolor{currentfill}%
\pgfsetlinewidth{0.803000pt}%
\definecolor{currentstroke}{rgb}{0.333333,0.333333,0.333333}%
\pgfsetstrokecolor{currentstroke}%
\pgfsetdash{}{0pt}%
\pgfsys@defobject{currentmarker}{\pgfqpoint{0.000000in}{-0.048611in}}{\pgfqpoint{0.000000in}{0.000000in}}{%
\pgfpathmoveto{\pgfqpoint{0.000000in}{0.000000in}}%
\pgfpathlineto{\pgfqpoint{0.000000in}{-0.048611in}}%
\pgfusepath{stroke,fill}%
}%
\begin{pgfscope}%
\pgfsys@transformshift{1.035218in}{0.693440in}%
\pgfsys@useobject{currentmarker}{}%
\end{pgfscope}%
\end{pgfscope}%
\begin{pgfscope}%
\definecolor{textcolor}{rgb}{0.333333,0.333333,0.333333}%
\pgfsetstrokecolor{textcolor}%
\pgfsetfillcolor{textcolor}%
\pgftext[x=1.035218in,y=0.596217in,,top]{\color{textcolor}{\rmfamily\fontsize{7.000000}{8.400000}\selectfont\catcode`\^=\active\def^{\ifmmode\sp\else\^{}\fi}\catcode`\%=\active\def
\end{pgfscope}%
\begin{pgfscope}%
\pgfpathrectangle{\pgfqpoint{0.491005in}{0.693440in}}{\pgfqpoint{2.118198in}{0.955233in}}%
\pgfusepath{clip}%
\pgfsetrectcap%
\pgfsetroundjoin%
\pgfsetlinewidth{0.803000pt}%
\definecolor{currentstroke}{rgb}{1.000000,1.000000,1.000000}%
\pgfsetstrokecolor{currentstroke}%
\pgfsetdash{}{0pt}%
\pgfpathmoveto{\pgfqpoint{1.564500in}{0.693440in}}%
\pgfpathlineto{\pgfqpoint{1.564500in}{1.648672in}}%
\pgfusepath{stroke}%
\end{pgfscope}%
\begin{pgfscope}%
\pgfsetbuttcap%
\pgfsetroundjoin%
\definecolor{currentfill}{rgb}{0.333333,0.333333,0.333333}%
\pgfsetfillcolor{currentfill}%
\pgfsetlinewidth{0.803000pt}%
\definecolor{currentstroke}{rgb}{0.333333,0.333333,0.333333}%
\pgfsetstrokecolor{currentstroke}%
\pgfsetdash{}{0pt}%
\pgfsys@defobject{currentmarker}{\pgfqpoint{0.000000in}{-0.048611in}}{\pgfqpoint{0.000000in}{0.000000in}}{%
\pgfpathmoveto{\pgfqpoint{0.000000in}{0.000000in}}%
\pgfpathlineto{\pgfqpoint{0.000000in}{-0.048611in}}%
\pgfusepath{stroke,fill}%
}%
\begin{pgfscope}%
\pgfsys@transformshift{1.564500in}{0.693440in}%
\pgfsys@useobject{currentmarker}{}%
\end{pgfscope}%
\end{pgfscope}%
\begin{pgfscope}%
\definecolor{textcolor}{rgb}{0.333333,0.333333,0.333333}%
\pgfsetstrokecolor{textcolor}%
\pgfsetfillcolor{textcolor}%
\pgftext[x=1.564500in,y=0.596217in,,top]{\color{textcolor}{\rmfamily\fontsize{7.000000}{8.400000}\selectfont\catcode`\^=\active\def^{\ifmmode\sp\else\^{}\fi}\catcode`\%=\active\def
\end{pgfscope}%
\begin{pgfscope}%
\pgfpathrectangle{\pgfqpoint{0.491005in}{0.693440in}}{\pgfqpoint{2.118198in}{0.955233in}}%
\pgfusepath{clip}%
\pgfsetrectcap%
\pgfsetroundjoin%
\pgfsetlinewidth{0.803000pt}%
\definecolor{currentstroke}{rgb}{1.000000,1.000000,1.000000}%
\pgfsetstrokecolor{currentstroke}%
\pgfsetdash{}{0pt}%
\pgfpathmoveto{\pgfqpoint{2.093782in}{0.693440in}}%
\pgfpathlineto{\pgfqpoint{2.093782in}{1.648672in}}%
\pgfusepath{stroke}%
\end{pgfscope}%
\begin{pgfscope}%
\pgfsetbuttcap%
\pgfsetroundjoin%
\definecolor{currentfill}{rgb}{0.333333,0.333333,0.333333}%
\pgfsetfillcolor{currentfill}%
\pgfsetlinewidth{0.803000pt}%
\definecolor{currentstroke}{rgb}{0.333333,0.333333,0.333333}%
\pgfsetstrokecolor{currentstroke}%
\pgfsetdash{}{0pt}%
\pgfsys@defobject{currentmarker}{\pgfqpoint{0.000000in}{-0.048611in}}{\pgfqpoint{0.000000in}{0.000000in}}{%
\pgfpathmoveto{\pgfqpoint{0.000000in}{0.000000in}}%
\pgfpathlineto{\pgfqpoint{0.000000in}{-0.048611in}}%
\pgfusepath{stroke,fill}%
}%
\begin{pgfscope}%
\pgfsys@transformshift{2.093782in}{0.693440in}%
\pgfsys@useobject{currentmarker}{}%
\end{pgfscope}%
\end{pgfscope}%
\begin{pgfscope}%
\definecolor{textcolor}{rgb}{0.333333,0.333333,0.333333}%
\pgfsetstrokecolor{textcolor}%
\pgfsetfillcolor{textcolor}%
\pgftext[x=2.093782in,y=0.596217in,,top]{\color{textcolor}{\rmfamily\fontsize{7.000000}{8.400000}\selectfont\catcode`\^=\active\def^{\ifmmode\sp\else\^{}\fi}\catcode`\%=\active\def
\end{pgfscope}%
\begin{pgfscope}%
\definecolor{textcolor}{rgb}{0.333333,0.333333,0.333333}%
\pgfsetstrokecolor{textcolor}%
\pgfsetfillcolor{textcolor}%
\pgftext[x=1.550104in,y=0.454242in,,top]{\color{textcolor}{\rmfamily\fontsize{9.000000}{10.800000}\selectfont\catcode`\^=\active\def^{\ifmmode\sp\else\^{}\fi}\catcode`\%=\active\def
\end{pgfscope}%
\begin{pgfscope}%
\pgfpathrectangle{\pgfqpoint{0.491005in}{0.693440in}}{\pgfqpoint{2.118198in}{0.955233in}}%
\pgfusepath{clip}%
\pgfsetrectcap%
\pgfsetroundjoin%
\pgfsetlinewidth{0.803000pt}%
\definecolor{currentstroke}{rgb}{1.000000,1.000000,1.000000}%
\pgfsetstrokecolor{currentstroke}%
\pgfsetdash{}{0pt}%
\pgfpathmoveto{\pgfqpoint{0.491005in}{0.794496in}}%
\pgfpathlineto{\pgfqpoint{2.609202in}{0.794496in}}%
\pgfusepath{stroke}%
\end{pgfscope}%
\begin{pgfscope}%
\pgfsetbuttcap%
\pgfsetroundjoin%
\definecolor{currentfill}{rgb}{0.121569,0.466667,0.705882}%
\pgfsetfillcolor{currentfill}%
\pgfsetlinewidth{0.803000pt}%
\definecolor{currentstroke}{rgb}{0.121569,0.466667,0.705882}%
\pgfsetstrokecolor{currentstroke}%
\pgfsetdash{}{0pt}%
\pgfsys@defobject{currentmarker}{\pgfqpoint{-0.048611in}{0.000000in}}{\pgfqpoint{-0.000000in}{0.000000in}}{%
\pgfpathmoveto{\pgfqpoint{-0.000000in}{0.000000in}}%
\pgfpathlineto{\pgfqpoint{-0.048611in}{0.000000in}}%
\pgfusepath{stroke,fill}%
}%
\begin{pgfscope}%
\pgfsys@transformshift{0.491005in}{0.794496in}%
\pgfsys@useobject{currentmarker}{}%
\end{pgfscope}%
\end{pgfscope}%
\begin{pgfscope}%
\definecolor{textcolor}{rgb}{0.121569,0.466667,0.705882}%
\pgfsetstrokecolor{textcolor}%
\pgfsetfillcolor{textcolor}%
\pgftext[x=0.338420in, y=0.760738in, left, base]{\color{textcolor}{\rmfamily\fontsize{7.000000}{8.400000}\selectfont\catcode`\^=\active\def^{\ifmmode\sp\else\^{}\fi}\catcode`\%=\active\def
\end{pgfscope}%
\begin{pgfscope}%
\pgfpathrectangle{\pgfqpoint{0.491005in}{0.693440in}}{\pgfqpoint{2.118198in}{0.955233in}}%
\pgfusepath{clip}%
\pgfsetrectcap%
\pgfsetroundjoin%
\pgfsetlinewidth{0.803000pt}%
\definecolor{currentstroke}{rgb}{1.000000,1.000000,1.000000}%
\pgfsetstrokecolor{currentstroke}%
\pgfsetdash{}{0pt}%
\pgfpathmoveto{\pgfqpoint{0.491005in}{0.986618in}}%
\pgfpathlineto{\pgfqpoint{2.609202in}{0.986618in}}%
\pgfusepath{stroke}%
\end{pgfscope}%
\begin{pgfscope}%
\pgfsetbuttcap%
\pgfsetroundjoin%
\definecolor{currentfill}{rgb}{0.121569,0.466667,0.705882}%
\pgfsetfillcolor{currentfill}%
\pgfsetlinewidth{0.803000pt}%
\definecolor{currentstroke}{rgb}{0.121569,0.466667,0.705882}%
\pgfsetstrokecolor{currentstroke}%
\pgfsetdash{}{0pt}%
\pgfsys@defobject{currentmarker}{\pgfqpoint{-0.048611in}{0.000000in}}{\pgfqpoint{-0.000000in}{0.000000in}}{%
\pgfpathmoveto{\pgfqpoint{-0.000000in}{0.000000in}}%
\pgfpathlineto{\pgfqpoint{-0.048611in}{0.000000in}}%
\pgfusepath{stroke,fill}%
}%
\begin{pgfscope}%
\pgfsys@transformshift{0.491005in}{0.986618in}%
\pgfsys@useobject{currentmarker}{}%
\end{pgfscope}%
\end{pgfscope}%
\begin{pgfscope}%
\definecolor{textcolor}{rgb}{0.121569,0.466667,0.705882}%
\pgfsetstrokecolor{textcolor}%
\pgfsetfillcolor{textcolor}%
\pgftext[x=0.283057in, y=0.952861in, left, base]{\color{textcolor}{\rmfamily\fontsize{7.000000}{8.400000}\selectfont\catcode`\^=\active\def^{\ifmmode\sp\else\^{}\fi}\catcode`\%=\active\def
\end{pgfscope}%
\begin{pgfscope}%
\pgfpathrectangle{\pgfqpoint{0.491005in}{0.693440in}}{\pgfqpoint{2.118198in}{0.955233in}}%
\pgfusepath{clip}%
\pgfsetrectcap%
\pgfsetroundjoin%
\pgfsetlinewidth{0.803000pt}%
\definecolor{currentstroke}{rgb}{1.000000,1.000000,1.000000}%
\pgfsetstrokecolor{currentstroke}%
\pgfsetdash{}{0pt}%
\pgfpathmoveto{\pgfqpoint{0.491005in}{1.178741in}}%
\pgfpathlineto{\pgfqpoint{2.609202in}{1.178741in}}%
\pgfusepath{stroke}%
\end{pgfscope}%
\begin{pgfscope}%
\pgfsetbuttcap%
\pgfsetroundjoin%
\definecolor{currentfill}{rgb}{0.121569,0.466667,0.705882}%
\pgfsetfillcolor{currentfill}%
\pgfsetlinewidth{0.803000pt}%
\definecolor{currentstroke}{rgb}{0.121569,0.466667,0.705882}%
\pgfsetstrokecolor{currentstroke}%
\pgfsetdash{}{0pt}%
\pgfsys@defobject{currentmarker}{\pgfqpoint{-0.048611in}{0.000000in}}{\pgfqpoint{-0.000000in}{0.000000in}}{%
\pgfpathmoveto{\pgfqpoint{-0.000000in}{0.000000in}}%
\pgfpathlineto{\pgfqpoint{-0.048611in}{0.000000in}}%
\pgfusepath{stroke,fill}%
}%
\begin{pgfscope}%
\pgfsys@transformshift{0.491005in}{1.178741in}%
\pgfsys@useobject{currentmarker}{}%
\end{pgfscope}%
\end{pgfscope}%
\begin{pgfscope}%
\definecolor{textcolor}{rgb}{0.121569,0.466667,0.705882}%
\pgfsetstrokecolor{textcolor}%
\pgfsetfillcolor{textcolor}%
\pgftext[x=0.283057in, y=1.144983in, left, base]{\color{textcolor}{\rmfamily\fontsize{7.000000}{8.400000}\selectfont\catcode`\^=\active\def^{\ifmmode\sp\else\^{}\fi}\catcode`\%=\active\def
\end{pgfscope}%
\begin{pgfscope}%
\pgfpathrectangle{\pgfqpoint{0.491005in}{0.693440in}}{\pgfqpoint{2.118198in}{0.955233in}}%
\pgfusepath{clip}%
\pgfsetrectcap%
\pgfsetroundjoin%
\pgfsetlinewidth{0.803000pt}%
\definecolor{currentstroke}{rgb}{1.000000,1.000000,1.000000}%
\pgfsetstrokecolor{currentstroke}%
\pgfsetdash{}{0pt}%
\pgfpathmoveto{\pgfqpoint{0.491005in}{1.370863in}}%
\pgfpathlineto{\pgfqpoint{2.609202in}{1.370863in}}%
\pgfusepath{stroke}%
\end{pgfscope}%
\begin{pgfscope}%
\pgfsetbuttcap%
\pgfsetroundjoin%
\definecolor{currentfill}{rgb}{0.121569,0.466667,0.705882}%
\pgfsetfillcolor{currentfill}%
\pgfsetlinewidth{0.803000pt}%
\definecolor{currentstroke}{rgb}{0.121569,0.466667,0.705882}%
\pgfsetstrokecolor{currentstroke}%
\pgfsetdash{}{0pt}%
\pgfsys@defobject{currentmarker}{\pgfqpoint{-0.048611in}{0.000000in}}{\pgfqpoint{-0.000000in}{0.000000in}}{%
\pgfpathmoveto{\pgfqpoint{-0.000000in}{0.000000in}}%
\pgfpathlineto{\pgfqpoint{-0.048611in}{0.000000in}}%
\pgfusepath{stroke,fill}%
}%
\begin{pgfscope}%
\pgfsys@transformshift{0.491005in}{1.370863in}%
\pgfsys@useobject{currentmarker}{}%
\end{pgfscope}%
\end{pgfscope}%
\begin{pgfscope}%
\definecolor{textcolor}{rgb}{0.121569,0.466667,0.705882}%
\pgfsetstrokecolor{textcolor}%
\pgfsetfillcolor{textcolor}%
\pgftext[x=0.283057in, y=1.337106in, left, base]{\color{textcolor}{\rmfamily\fontsize{7.000000}{8.400000}\selectfont\catcode`\^=\active\def^{\ifmmode\sp\else\^{}\fi}\catcode`\%=\active\def
\end{pgfscope}%
\begin{pgfscope}%
\pgfpathrectangle{\pgfqpoint{0.491005in}{0.693440in}}{\pgfqpoint{2.118198in}{0.955233in}}%
\pgfusepath{clip}%
\pgfsetrectcap%
\pgfsetroundjoin%
\pgfsetlinewidth{0.803000pt}%
\definecolor{currentstroke}{rgb}{1.000000,1.000000,1.000000}%
\pgfsetstrokecolor{currentstroke}%
\pgfsetdash{}{0pt}%
\pgfpathmoveto{\pgfqpoint{0.491005in}{1.562986in}}%
\pgfpathlineto{\pgfqpoint{2.609202in}{1.562986in}}%
\pgfusepath{stroke}%
\end{pgfscope}%
\begin{pgfscope}%
\pgfsetbuttcap%
\pgfsetroundjoin%
\definecolor{currentfill}{rgb}{0.121569,0.466667,0.705882}%
\pgfsetfillcolor{currentfill}%
\pgfsetlinewidth{0.803000pt}%
\definecolor{currentstroke}{rgb}{0.121569,0.466667,0.705882}%
\pgfsetstrokecolor{currentstroke}%
\pgfsetdash{}{0pt}%
\pgfsys@defobject{currentmarker}{\pgfqpoint{-0.048611in}{0.000000in}}{\pgfqpoint{-0.000000in}{0.000000in}}{%
\pgfpathmoveto{\pgfqpoint{-0.000000in}{0.000000in}}%
\pgfpathlineto{\pgfqpoint{-0.048611in}{0.000000in}}%
\pgfusepath{stroke,fill}%
}%
\begin{pgfscope}%
\pgfsys@transformshift{0.491005in}{1.562986in}%
\pgfsys@useobject{currentmarker}{}%
\end{pgfscope}%
\end{pgfscope}%
\begin{pgfscope}%
\definecolor{textcolor}{rgb}{0.121569,0.466667,0.705882}%
\pgfsetstrokecolor{textcolor}%
\pgfsetfillcolor{textcolor}%
\pgftext[x=0.283057in, y=1.529228in, left, base]{\color{textcolor}{\rmfamily\fontsize{7.000000}{8.400000}\selectfont\catcode`\^=\active\def^{\ifmmode\sp\else\^{}\fi}\catcode`\%=\active\def
\end{pgfscope}%
\begin{pgfscope}%
\definecolor{textcolor}{rgb}{0.121569,0.466667,0.705882}%
\pgfsetstrokecolor{textcolor}%
\pgfsetfillcolor{textcolor}%
\pgftext[x=0.227501in,y=1.171056in,,bottom,rotate=90.000000]{\color{textcolor}{\rmfamily\fontsize{9.000000}{10.800000}\selectfont\catcode`\^=\active\def^{\ifmmode\sp\else\^{}\fi}\catcode`\%=\active\def
\end{pgfscope}%
\begin{pgfscope}%
\pgfpathrectangle{\pgfqpoint{0.491005in}{0.693440in}}{\pgfqpoint{2.118198in}{0.955233in}}%
\pgfusepath{clip}%
\pgfsetrectcap%
\pgfsetroundjoin%
\pgfsetlinewidth{0.702625pt}%
\definecolor{currentstroke}{rgb}{0.121569,0.466667,0.705882}%
\pgfsetstrokecolor{currentstroke}%
\pgfsetdash{}{0pt}%
\pgfpathmoveto{\pgfqpoint{0.587287in}{0.736859in}}%
\pgfpathlineto{\pgfqpoint{0.615709in}{0.752229in}}%
\pgfpathlineto{\pgfqpoint{0.631323in}{0.767599in}}%
\pgfpathlineto{\pgfqpoint{0.651541in}{0.779126in}}%
\pgfpathlineto{\pgfqpoint{0.681658in}{0.798338in}}%
\pgfpathlineto{\pgfqpoint{0.703676in}{0.806023in}}%
\pgfpathlineto{\pgfqpoint{0.727282in}{0.825236in}}%
\pgfpathlineto{\pgfqpoint{0.769465in}{0.848290in}}%
\pgfpathlineto{\pgfqpoint{0.825940in}{0.871345in}}%
\pgfpathlineto{\pgfqpoint{0.918405in}{0.921297in}}%
\pgfpathlineto{\pgfqpoint{1.085923in}{0.998146in}}%
\pgfpathlineto{\pgfqpoint{1.204059in}{1.048098in}}%
\pgfpathlineto{\pgfqpoint{1.523745in}{1.151844in}}%
\pgfpathlineto{\pgfqpoint{1.716404in}{1.297857in}}%
\pgfpathlineto{\pgfqpoint{1.998829in}{1.428500in}}%
\pgfpathlineto{\pgfqpoint{2.512921in}{1.605253in}}%
\pgfusepath{stroke}%
\end{pgfscope}%
\begin{pgfscope}%
\pgfpathrectangle{\pgfqpoint{0.491005in}{0.693440in}}{\pgfqpoint{2.118198in}{0.955233in}}%
\pgfusepath{clip}%
\pgfsetbuttcap%
\pgfsetroundjoin%
\definecolor{currentfill}{rgb}{0.121569,0.466667,0.705882}%
\pgfsetfillcolor{currentfill}%
\pgfsetlinewidth{1.003750pt}%
\definecolor{currentstroke}{rgb}{0.121569,0.466667,0.705882}%
\pgfsetstrokecolor{currentstroke}%
\pgfsetdash{}{0pt}%
\pgfsys@defobject{currentmarker}{\pgfqpoint{-0.013889in}{-0.013889in}}{\pgfqpoint{0.013889in}{0.013889in}}{%
\pgfpathmoveto{\pgfqpoint{0.000000in}{-0.013889in}}%
\pgfpathcurveto{\pgfqpoint{0.003683in}{-0.013889in}}{\pgfqpoint{0.007216in}{-0.012425in}}{\pgfqpoint{0.009821in}{-0.009821in}}%
\pgfpathcurveto{\pgfqpoint{0.012425in}{-0.007216in}}{\pgfqpoint{0.013889in}{-0.003683in}}{\pgfqpoint{0.013889in}{0.000000in}}%
\pgfpathcurveto{\pgfqpoint{0.013889in}{0.003683in}}{\pgfqpoint{0.012425in}{0.007216in}}{\pgfqpoint{0.009821in}{0.009821in}}%
\pgfpathcurveto{\pgfqpoint{0.007216in}{0.012425in}}{\pgfqpoint{0.003683in}{0.013889in}}{\pgfqpoint{0.000000in}{0.013889in}}%
\pgfpathcurveto{\pgfqpoint{-0.003683in}{0.013889in}}{\pgfqpoint{-0.007216in}{0.012425in}}{\pgfqpoint{-0.009821in}{0.009821in}}%
\pgfpathcurveto{\pgfqpoint{-0.012425in}{0.007216in}}{\pgfqpoint{-0.013889in}{0.003683in}}{\pgfqpoint{-0.013889in}{0.000000in}}%
\pgfpathcurveto{\pgfqpoint{-0.013889in}{-0.003683in}}{\pgfqpoint{-0.012425in}{-0.007216in}}{\pgfqpoint{-0.009821in}{-0.009821in}}%
\pgfpathcurveto{\pgfqpoint{-0.007216in}{-0.012425in}}{\pgfqpoint{-0.003683in}{-0.013889in}}{\pgfqpoint{0.000000in}{-0.013889in}}%
\pgfpathlineto{\pgfqpoint{0.000000in}{-0.013889in}}%
\pgfpathclose%
\pgfusepath{stroke,fill}%
}%
\begin{pgfscope}%
\pgfsys@transformshift{0.587287in}{0.736859in}%
\pgfsys@useobject{currentmarker}{}%
\end{pgfscope}%
\begin{pgfscope}%
\pgfsys@transformshift{0.615709in}{0.752229in}%
\pgfsys@useobject{currentmarker}{}%
\end{pgfscope}%
\begin{pgfscope}%
\pgfsys@transformshift{0.631323in}{0.767599in}%
\pgfsys@useobject{currentmarker}{}%
\end{pgfscope}%
\begin{pgfscope}%
\pgfsys@transformshift{0.651541in}{0.779126in}%
\pgfsys@useobject{currentmarker}{}%
\end{pgfscope}%
\begin{pgfscope}%
\pgfsys@transformshift{0.681658in}{0.798338in}%
\pgfsys@useobject{currentmarker}{}%
\end{pgfscope}%
\begin{pgfscope}%
\pgfsys@transformshift{0.703676in}{0.806023in}%
\pgfsys@useobject{currentmarker}{}%
\end{pgfscope}%
\begin{pgfscope}%
\pgfsys@transformshift{0.727282in}{0.825236in}%
\pgfsys@useobject{currentmarker}{}%
\end{pgfscope}%
\begin{pgfscope}%
\pgfsys@transformshift{0.769465in}{0.848290in}%
\pgfsys@useobject{currentmarker}{}%
\end{pgfscope}%
\begin{pgfscope}%
\pgfsys@transformshift{0.825940in}{0.871345in}%
\pgfsys@useobject{currentmarker}{}%
\end{pgfscope}%
\begin{pgfscope}%
\pgfsys@transformshift{0.918405in}{0.921297in}%
\pgfsys@useobject{currentmarker}{}%
\end{pgfscope}%
\begin{pgfscope}%
\pgfsys@transformshift{1.085923in}{0.998146in}%
\pgfsys@useobject{currentmarker}{}%
\end{pgfscope}%
\begin{pgfscope}%
\pgfsys@transformshift{1.204059in}{1.048098in}%
\pgfsys@useobject{currentmarker}{}%
\end{pgfscope}%
\begin{pgfscope}%
\pgfsys@transformshift{1.523745in}{1.151844in}%
\pgfsys@useobject{currentmarker}{}%
\end{pgfscope}%
\begin{pgfscope}%
\pgfsys@transformshift{1.716404in}{1.297857in}%
\pgfsys@useobject{currentmarker}{}%
\end{pgfscope}%
\begin{pgfscope}%
\pgfsys@transformshift{1.998829in}{1.428500in}%
\pgfsys@useobject{currentmarker}{}%
\end{pgfscope}%
\begin{pgfscope}%
\pgfsys@transformshift{2.512921in}{1.605253in}%
\pgfsys@useobject{currentmarker}{}%
\end{pgfscope}%
\end{pgfscope}%
\begin{pgfscope}%
\pgfpathrectangle{\pgfqpoint{0.491005in}{0.693440in}}{\pgfqpoint{2.118198in}{0.955233in}}%
\pgfusepath{clip}%
\pgfsetbuttcap%
\pgfsetroundjoin%
\pgfsetlinewidth{0.401500pt}%
\definecolor{currentstroke}{rgb}{0.121569,0.466667,0.705882}%
\pgfsetstrokecolor{currentstroke}%
\pgfsetdash{{1.480000pt}{0.640000pt}}{0.000000pt}%
\pgfpathmoveto{\pgfqpoint{0.491005in}{1.297857in}}%
\pgfpathlineto{\pgfqpoint{2.609202in}{1.297857in}}%
\pgfusepath{stroke}%
\end{pgfscope}%
\begin{pgfscope}%
\pgfsetrectcap%
\pgfsetmiterjoin%
\pgfsetlinewidth{1.003750pt}%
\definecolor{currentstroke}{rgb}{1.000000,1.000000,1.000000}%
\pgfsetstrokecolor{currentstroke}%
\pgfsetdash{}{0pt}%
\pgfpathmoveto{\pgfqpoint{0.491005in}{0.693440in}}%
\pgfpathlineto{\pgfqpoint{0.491005in}{1.648672in}}%
\pgfusepath{stroke}%
\end{pgfscope}%
\begin{pgfscope}%
\pgfsetrectcap%
\pgfsetmiterjoin%
\pgfsetlinewidth{1.003750pt}%
\definecolor{currentstroke}{rgb}{1.000000,1.000000,1.000000}%
\pgfsetstrokecolor{currentstroke}%
\pgfsetdash{}{0pt}%
\pgfpathmoveto{\pgfqpoint{2.609202in}{0.693440in}}%
\pgfpathlineto{\pgfqpoint{2.609202in}{1.648672in}}%
\pgfusepath{stroke}%
\end{pgfscope}%
\begin{pgfscope}%
\pgfsetrectcap%
\pgfsetmiterjoin%
\pgfsetlinewidth{1.003750pt}%
\definecolor{currentstroke}{rgb}{1.000000,1.000000,1.000000}%
\pgfsetstrokecolor{currentstroke}%
\pgfsetdash{}{0pt}%
\pgfpathmoveto{\pgfqpoint{0.491005in}{0.693440in}}%
\pgfpathlineto{\pgfqpoint{2.609202in}{0.693440in}}%
\pgfusepath{stroke}%
\end{pgfscope}%
\begin{pgfscope}%
\pgfsetrectcap%
\pgfsetmiterjoin%
\pgfsetlinewidth{1.003750pt}%
\definecolor{currentstroke}{rgb}{1.000000,1.000000,1.000000}%
\pgfsetstrokecolor{currentstroke}%
\pgfsetdash{}{0pt}%
\pgfpathmoveto{\pgfqpoint{0.491005in}{1.648672in}}%
\pgfpathlineto{\pgfqpoint{2.609202in}{1.648672in}}%
\pgfusepath{stroke}%
\end{pgfscope}%
\begin{pgfscope}%
\pgfsetbuttcap%
\pgfsetroundjoin%
\definecolor{currentfill}{rgb}{0.333333,0.333333,0.333333}%
\pgfsetfillcolor{currentfill}%
\pgfsetlinewidth{0.803000pt}%
\definecolor{currentstroke}{rgb}{0.333333,0.333333,0.333333}%
\pgfsetstrokecolor{currentstroke}%
\pgfsetdash{}{0pt}%
\pgfsys@defobject{currentmarker}{\pgfqpoint{0.000000in}{0.000000in}}{\pgfqpoint{0.048611in}{0.000000in}}{%
\pgfpathmoveto{\pgfqpoint{0.000000in}{0.000000in}}%
\pgfpathlineto{\pgfqpoint{0.048611in}{0.000000in}}%
\pgfusepath{stroke,fill}%
}%
\begin{pgfscope}%
\pgfsys@transformshift{2.609202in}{0.763891in}%
\pgfsys@useobject{currentmarker}{}%
\end{pgfscope}%
\end{pgfscope}%
\begin{pgfscope}%
\definecolor{textcolor}{rgb}{0.333333,0.333333,0.333333}%
\pgfsetstrokecolor{textcolor}%
\pgfsetfillcolor{textcolor}%
\pgftext[x=2.706425in, y=0.730133in, left, base]{\color{textcolor}{\rmfamily\fontsize{7.000000}{8.400000}\selectfont\catcode`\^=\active\def^{\ifmmode\sp\else\^{}\fi}\catcode`\%=\active\def
\end{pgfscope}%
\begin{pgfscope}%
\pgfsetbuttcap%
\pgfsetroundjoin%
\definecolor{currentfill}{rgb}{0.333333,0.333333,0.333333}%
\pgfsetfillcolor{currentfill}%
\pgfsetlinewidth{0.803000pt}%
\definecolor{currentstroke}{rgb}{0.333333,0.333333,0.333333}%
\pgfsetstrokecolor{currentstroke}%
\pgfsetdash{}{0pt}%
\pgfsys@defobject{currentmarker}{\pgfqpoint{0.000000in}{0.000000in}}{\pgfqpoint{0.048611in}{0.000000in}}{%
\pgfpathmoveto{\pgfqpoint{0.000000in}{0.000000in}}%
\pgfpathlineto{\pgfqpoint{0.048611in}{0.000000in}}%
\pgfusepath{stroke,fill}%
}%
\begin{pgfscope}%
\pgfsys@transformshift{2.609202in}{0.966629in}%
\pgfsys@useobject{currentmarker}{}%
\end{pgfscope}%
\end{pgfscope}%
\begin{pgfscope}%
\definecolor{textcolor}{rgb}{0.333333,0.333333,0.333333}%
\pgfsetstrokecolor{textcolor}%
\pgfsetfillcolor{textcolor}%
\pgftext[x=2.706425in, y=0.932871in, left, base]{\color{textcolor}{\rmfamily\fontsize{7.000000}{8.400000}\selectfont\catcode`\^=\active\def^{\ifmmode\sp\else\^{}\fi}\catcode`\%=\active\def
\end{pgfscope}%
\begin{pgfscope}%
\pgfsetbuttcap%
\pgfsetroundjoin%
\definecolor{currentfill}{rgb}{0.333333,0.333333,0.333333}%
\pgfsetfillcolor{currentfill}%
\pgfsetlinewidth{0.803000pt}%
\definecolor{currentstroke}{rgb}{0.333333,0.333333,0.333333}%
\pgfsetstrokecolor{currentstroke}%
\pgfsetdash{}{0pt}%
\pgfsys@defobject{currentmarker}{\pgfqpoint{0.000000in}{0.000000in}}{\pgfqpoint{0.048611in}{0.000000in}}{%
\pgfpathmoveto{\pgfqpoint{0.000000in}{0.000000in}}%
\pgfpathlineto{\pgfqpoint{0.048611in}{0.000000in}}%
\pgfusepath{stroke,fill}%
}%
\begin{pgfscope}%
\pgfsys@transformshift{2.609202in}{1.169366in}%
\pgfsys@useobject{currentmarker}{}%
\end{pgfscope}%
\end{pgfscope}%
\begin{pgfscope}%
\definecolor{textcolor}{rgb}{0.333333,0.333333,0.333333}%
\pgfsetstrokecolor{textcolor}%
\pgfsetfillcolor{textcolor}%
\pgftext[x=2.706425in, y=1.135609in, left, base]{\color{textcolor}{\rmfamily\fontsize{7.000000}{8.400000}\selectfont\catcode`\^=\active\def^{\ifmmode\sp\else\^{}\fi}\catcode`\%=\active\def
\end{pgfscope}%
\begin{pgfscope}%
\pgfsetbuttcap%
\pgfsetroundjoin%
\definecolor{currentfill}{rgb}{0.333333,0.333333,0.333333}%
\pgfsetfillcolor{currentfill}%
\pgfsetlinewidth{0.803000pt}%
\definecolor{currentstroke}{rgb}{0.333333,0.333333,0.333333}%
\pgfsetstrokecolor{currentstroke}%
\pgfsetdash{}{0pt}%
\pgfsys@defobject{currentmarker}{\pgfqpoint{0.000000in}{0.000000in}}{\pgfqpoint{0.048611in}{0.000000in}}{%
\pgfpathmoveto{\pgfqpoint{0.000000in}{0.000000in}}%
\pgfpathlineto{\pgfqpoint{0.048611in}{0.000000in}}%
\pgfusepath{stroke,fill}%
}%
\begin{pgfscope}%
\pgfsys@transformshift{2.609202in}{1.372104in}%
\pgfsys@useobject{currentmarker}{}%
\end{pgfscope}%
\end{pgfscope}%
\begin{pgfscope}%
\definecolor{textcolor}{rgb}{0.333333,0.333333,0.333333}%
\pgfsetstrokecolor{textcolor}%
\pgfsetfillcolor{textcolor}%
\pgftext[x=2.706425in, y=1.338347in, left, base]{\color{textcolor}{\rmfamily\fontsize{7.000000}{8.400000}\selectfont\catcode`\^=\active\def^{\ifmmode\sp\else\^{}\fi}\catcode`\%=\active\def
\end{pgfscope}%
\begin{pgfscope}%
\pgfsetbuttcap%
\pgfsetroundjoin%
\definecolor{currentfill}{rgb}{0.333333,0.333333,0.333333}%
\pgfsetfillcolor{currentfill}%
\pgfsetlinewidth{0.803000pt}%
\definecolor{currentstroke}{rgb}{0.333333,0.333333,0.333333}%
\pgfsetstrokecolor{currentstroke}%
\pgfsetdash{}{0pt}%
\pgfsys@defobject{currentmarker}{\pgfqpoint{0.000000in}{0.000000in}}{\pgfqpoint{0.048611in}{0.000000in}}{%
\pgfpathmoveto{\pgfqpoint{0.000000in}{0.000000in}}%
\pgfpathlineto{\pgfqpoint{0.048611in}{0.000000in}}%
\pgfusepath{stroke,fill}%
}%
\begin{pgfscope}%
\pgfsys@transformshift{2.609202in}{1.574842in}%
\pgfsys@useobject{currentmarker}{}%
\end{pgfscope}%
\end{pgfscope}%
\begin{pgfscope}%
\definecolor{textcolor}{rgb}{0.333333,0.333333,0.333333}%
\pgfsetstrokecolor{textcolor}%
\pgfsetfillcolor{textcolor}%
\pgftext[x=2.706425in, y=1.541084in, left, base]{\color{textcolor}{\rmfamily\fontsize{7.000000}{8.400000}\selectfont\catcode`\^=\active\def^{\ifmmode\sp\else\^{}\fi}\catcode`\%=\active\def
\end{pgfscope}%
\begin{pgfscope}%
\definecolor{textcolor}{rgb}{0.000000,0.000000,0.000000}%
\pgfsetstrokecolor{textcolor}%
\pgfsetfillcolor{textcolor}%
\pgftext[x=2.872706in,y=1.171056in,,top,rotate=90.000000]{\color{textcolor}{\rmfamily\fontsize{9.000000}{10.800000}\selectfont\catcode`\^=\active\def^{\ifmmode\sp\else\^{}\fi}\catcode`\%=\active\def
\end{pgfscope}%
\begin{pgfscope}%
\pgfpathrectangle{\pgfqpoint{0.491005in}{0.693440in}}{\pgfqpoint{2.118198in}{0.955233in}}%
\pgfusepath{clip}%
\pgfsetrectcap%
\pgfsetroundjoin%
\pgfsetlinewidth{0.702625pt}%
\definecolor{currentstroke}{rgb}{1.000000,0.498039,0.054902}%
\pgfsetstrokecolor{currentstroke}%
\pgfsetdash{}{0pt}%
\pgfpathmoveto{\pgfqpoint{0.587287in}{0.736859in}}%
\pgfpathlineto{\pgfqpoint{0.615709in}{0.763891in}}%
\pgfpathlineto{\pgfqpoint{0.631323in}{0.780786in}}%
\pgfpathlineto{\pgfqpoint{0.651541in}{0.784165in}}%
\pgfpathlineto{\pgfqpoint{0.681658in}{0.801060in}}%
\pgfpathlineto{\pgfqpoint{0.703676in}{0.811196in}}%
\pgfpathlineto{\pgfqpoint{0.727282in}{0.821333in}}%
\pgfpathlineto{\pgfqpoint{0.769465in}{0.824712in}}%
\pgfpathlineto{\pgfqpoint{0.825940in}{0.851744in}}%
\pgfpathlineto{\pgfqpoint{0.918405in}{0.858502in}}%
\pgfpathlineto{\pgfqpoint{1.085923in}{0.902428in}}%
\pgfpathlineto{\pgfqpoint{1.204059in}{0.892292in}}%
\pgfpathlineto{\pgfqpoint{1.523745in}{0.912565in}}%
\pgfpathlineto{\pgfqpoint{1.716404in}{0.926081in}}%
\pgfpathlineto{\pgfqpoint{1.998829in}{0.959871in}}%
\pgfpathlineto{\pgfqpoint{2.512921in}{0.963250in}}%
\pgfusepath{stroke}%
\end{pgfscope}%
\begin{pgfscope}%
\pgfpathrectangle{\pgfqpoint{0.491005in}{0.693440in}}{\pgfqpoint{2.118198in}{0.955233in}}%
\pgfusepath{clip}%
\pgfsetbuttcap%
\pgfsetmiterjoin%
\definecolor{currentfill}{rgb}{1.000000,0.498039,0.054902}%
\pgfsetfillcolor{currentfill}%
\pgfsetlinewidth{1.003750pt}%
\definecolor{currentstroke}{rgb}{1.000000,0.498039,0.054902}%
\pgfsetstrokecolor{currentstroke}%
\pgfsetdash{}{0pt}%
\pgfsys@defobject{currentmarker}{\pgfqpoint{-0.013889in}{-0.013889in}}{\pgfqpoint{0.013889in}{0.013889in}}{%
\pgfpathmoveto{\pgfqpoint{-0.013889in}{-0.013889in}}%
\pgfpathlineto{\pgfqpoint{0.013889in}{-0.013889in}}%
\pgfpathlineto{\pgfqpoint{0.013889in}{0.013889in}}%
\pgfpathlineto{\pgfqpoint{-0.013889in}{0.013889in}}%
\pgfpathlineto{\pgfqpoint{-0.013889in}{-0.013889in}}%
\pgfpathclose%
\pgfusepath{stroke,fill}%
}%
\begin{pgfscope}%
\pgfsys@transformshift{0.587287in}{0.736859in}%
\pgfsys@useobject{currentmarker}{}%
\end{pgfscope}%
\begin{pgfscope}%
\pgfsys@transformshift{0.615709in}{0.763891in}%
\pgfsys@useobject{currentmarker}{}%
\end{pgfscope}%
\begin{pgfscope}%
\pgfsys@transformshift{0.631323in}{0.780786in}%
\pgfsys@useobject{currentmarker}{}%
\end{pgfscope}%
\begin{pgfscope}%
\pgfsys@transformshift{0.651541in}{0.784165in}%
\pgfsys@useobject{currentmarker}{}%
\end{pgfscope}%
\begin{pgfscope}%
\pgfsys@transformshift{0.681658in}{0.801060in}%
\pgfsys@useobject{currentmarker}{}%
\end{pgfscope}%
\begin{pgfscope}%
\pgfsys@transformshift{0.703676in}{0.811196in}%
\pgfsys@useobject{currentmarker}{}%
\end{pgfscope}%
\begin{pgfscope}%
\pgfsys@transformshift{0.727282in}{0.821333in}%
\pgfsys@useobject{currentmarker}{}%
\end{pgfscope}%
\begin{pgfscope}%
\pgfsys@transformshift{0.769465in}{0.824712in}%
\pgfsys@useobject{currentmarker}{}%
\end{pgfscope}%
\begin{pgfscope}%
\pgfsys@transformshift{0.825940in}{0.851744in}%
\pgfsys@useobject{currentmarker}{}%
\end{pgfscope}%
\begin{pgfscope}%
\pgfsys@transformshift{0.918405in}{0.858502in}%
\pgfsys@useobject{currentmarker}{}%
\end{pgfscope}%
\begin{pgfscope}%
\pgfsys@transformshift{1.085923in}{0.902428in}%
\pgfsys@useobject{currentmarker}{}%
\end{pgfscope}%
\begin{pgfscope}%
\pgfsys@transformshift{1.204059in}{0.892292in}%
\pgfsys@useobject{currentmarker}{}%
\end{pgfscope}%
\begin{pgfscope}%
\pgfsys@transformshift{1.523745in}{0.912565in}%
\pgfsys@useobject{currentmarker}{}%
\end{pgfscope}%
\begin{pgfscope}%
\pgfsys@transformshift{1.716404in}{0.926081in}%
\pgfsys@useobject{currentmarker}{}%
\end{pgfscope}%
\begin{pgfscope}%
\pgfsys@transformshift{1.998829in}{0.959871in}%
\pgfsys@useobject{currentmarker}{}%
\end{pgfscope}%
\begin{pgfscope}%
\pgfsys@transformshift{2.512921in}{0.963250in}%
\pgfsys@useobject{currentmarker}{}%
\end{pgfscope}%
\end{pgfscope}%
\begin{pgfscope}%
\pgfpathrectangle{\pgfqpoint{0.491005in}{0.693440in}}{\pgfqpoint{2.118198in}{0.955233in}}%
\pgfusepath{clip}%
\pgfsetrectcap%
\pgfsetroundjoin%
\pgfsetlinewidth{0.702625pt}%
\definecolor{currentstroke}{rgb}{0.172549,0.627451,0.172549}%
\pgfsetstrokecolor{currentstroke}%
\pgfsetdash{}{0pt}%
\pgfpathmoveto{\pgfqpoint{0.587287in}{1.605253in}}%
\pgfpathlineto{\pgfqpoint{0.615709in}{1.584979in}}%
\pgfpathlineto{\pgfqpoint{0.631323in}{1.551189in}}%
\pgfpathlineto{\pgfqpoint{0.651541in}{1.554568in}}%
\pgfpathlineto{\pgfqpoint{0.681658in}{1.524158in}}%
\pgfpathlineto{\pgfqpoint{0.703676in}{1.520779in}}%
\pgfpathlineto{\pgfqpoint{0.727282in}{1.497126in}}%
\pgfpathlineto{\pgfqpoint{0.769465in}{1.473473in}}%
\pgfpathlineto{\pgfqpoint{0.825940in}{1.473473in}}%
\pgfpathlineto{\pgfqpoint{0.918405in}{1.432926in}}%
\pgfpathlineto{\pgfqpoint{1.085923in}{1.399136in}}%
\pgfpathlineto{\pgfqpoint{1.204059in}{1.405894in}}%
\pgfpathlineto{\pgfqpoint{1.523745in}{1.375483in}}%
\pgfpathlineto{\pgfqpoint{1.716404in}{1.361967in}}%
\pgfpathlineto{\pgfqpoint{1.998829in}{1.358588in}}%
\pgfpathlineto{\pgfqpoint{2.512921in}{1.355209in}}%
\pgfusepath{stroke}%
\end{pgfscope}%
\begin{pgfscope}%
\pgfpathrectangle{\pgfqpoint{0.491005in}{0.693440in}}{\pgfqpoint{2.118198in}{0.955233in}}%
\pgfusepath{clip}%
\pgfsetbuttcap%
\pgfsetmiterjoin%
\definecolor{currentfill}{rgb}{0.172549,0.627451,0.172549}%
\pgfsetfillcolor{currentfill}%
\pgfsetlinewidth{1.003750pt}%
\definecolor{currentstroke}{rgb}{0.172549,0.627451,0.172549}%
\pgfsetstrokecolor{currentstroke}%
\pgfsetdash{}{0pt}%
\pgfsys@defobject{currentmarker}{\pgfqpoint{-0.013889in}{-0.013889in}}{\pgfqpoint{0.013889in}{0.013889in}}{%
\pgfpathmoveto{\pgfqpoint{0.000000in}{0.013889in}}%
\pgfpathlineto{\pgfqpoint{-0.013889in}{-0.013889in}}%
\pgfpathlineto{\pgfqpoint{0.013889in}{-0.013889in}}%
\pgfpathlineto{\pgfqpoint{0.000000in}{0.013889in}}%
\pgfpathclose%
\pgfusepath{stroke,fill}%
}%
\begin{pgfscope}%
\pgfsys@transformshift{0.587287in}{1.605253in}%
\pgfsys@useobject{currentmarker}{}%
\end{pgfscope}%
\begin{pgfscope}%
\pgfsys@transformshift{0.615709in}{1.584979in}%
\pgfsys@useobject{currentmarker}{}%
\end{pgfscope}%
\begin{pgfscope}%
\pgfsys@transformshift{0.631323in}{1.551189in}%
\pgfsys@useobject{currentmarker}{}%
\end{pgfscope}%
\begin{pgfscope}%
\pgfsys@transformshift{0.651541in}{1.554568in}%
\pgfsys@useobject{currentmarker}{}%
\end{pgfscope}%
\begin{pgfscope}%
\pgfsys@transformshift{0.681658in}{1.524158in}%
\pgfsys@useobject{currentmarker}{}%
\end{pgfscope}%
\begin{pgfscope}%
\pgfsys@transformshift{0.703676in}{1.520779in}%
\pgfsys@useobject{currentmarker}{}%
\end{pgfscope}%
\begin{pgfscope}%
\pgfsys@transformshift{0.727282in}{1.497126in}%
\pgfsys@useobject{currentmarker}{}%
\end{pgfscope}%
\begin{pgfscope}%
\pgfsys@transformshift{0.769465in}{1.473473in}%
\pgfsys@useobject{currentmarker}{}%
\end{pgfscope}%
\begin{pgfscope}%
\pgfsys@transformshift{0.825940in}{1.473473in}%
\pgfsys@useobject{currentmarker}{}%
\end{pgfscope}%
\begin{pgfscope}%
\pgfsys@transformshift{0.918405in}{1.432926in}%
\pgfsys@useobject{currentmarker}{}%
\end{pgfscope}%
\begin{pgfscope}%
\pgfsys@transformshift{1.085923in}{1.399136in}%
\pgfsys@useobject{currentmarker}{}%
\end{pgfscope}%
\begin{pgfscope}%
\pgfsys@transformshift{1.204059in}{1.405894in}%
\pgfsys@useobject{currentmarker}{}%
\end{pgfscope}%
\begin{pgfscope}%
\pgfsys@transformshift{1.523745in}{1.375483in}%
\pgfsys@useobject{currentmarker}{}%
\end{pgfscope}%
\begin{pgfscope}%
\pgfsys@transformshift{1.716404in}{1.361967in}%
\pgfsys@useobject{currentmarker}{}%
\end{pgfscope}%
\begin{pgfscope}%
\pgfsys@transformshift{1.998829in}{1.358588in}%
\pgfsys@useobject{currentmarker}{}%
\end{pgfscope}%
\begin{pgfscope}%
\pgfsys@transformshift{2.512921in}{1.355209in}%
\pgfsys@useobject{currentmarker}{}%
\end{pgfscope}%
\end{pgfscope}%
\begin{pgfscope}%
\pgfsetrectcap%
\pgfsetmiterjoin%
\pgfsetlinewidth{1.003750pt}%
\definecolor{currentstroke}{rgb}{1.000000,1.000000,1.000000}%
\pgfsetstrokecolor{currentstroke}%
\pgfsetdash{}{0pt}%
\pgfpathmoveto{\pgfqpoint{0.491005in}{0.693440in}}%
\pgfpathlineto{\pgfqpoint{0.491005in}{1.648672in}}%
\pgfusepath{stroke}%
\end{pgfscope}%
\begin{pgfscope}%
\pgfsetrectcap%
\pgfsetmiterjoin%
\pgfsetlinewidth{1.003750pt}%
\definecolor{currentstroke}{rgb}{1.000000,1.000000,1.000000}%
\pgfsetstrokecolor{currentstroke}%
\pgfsetdash{}{0pt}%
\pgfpathmoveto{\pgfqpoint{2.609202in}{0.693440in}}%
\pgfpathlineto{\pgfqpoint{2.609202in}{1.648672in}}%
\pgfusepath{stroke}%
\end{pgfscope}%
\begin{pgfscope}%
\pgfsetrectcap%
\pgfsetmiterjoin%
\pgfsetlinewidth{1.003750pt}%
\definecolor{currentstroke}{rgb}{1.000000,1.000000,1.000000}%
\pgfsetstrokecolor{currentstroke}%
\pgfsetdash{}{0pt}%
\pgfpathmoveto{\pgfqpoint{0.491005in}{0.693440in}}%
\pgfpathlineto{\pgfqpoint{2.609202in}{0.693440in}}%
\pgfusepath{stroke}%
\end{pgfscope}%
\begin{pgfscope}%
\pgfsetrectcap%
\pgfsetmiterjoin%
\pgfsetlinewidth{1.003750pt}%
\definecolor{currentstroke}{rgb}{1.000000,1.000000,1.000000}%
\pgfsetstrokecolor{currentstroke}%
\pgfsetdash{}{0pt}%
\pgfpathmoveto{\pgfqpoint{0.491005in}{1.648672in}}%
\pgfpathlineto{\pgfqpoint{2.609202in}{1.648672in}}%
\pgfusepath{stroke}%
\end{pgfscope}%
\begin{pgfscope}%
\pgfsetbuttcap%
\pgfsetmiterjoin%
\definecolor{currentfill}{rgb}{0.898039,0.898039,0.898039}%
\pgfsetfillcolor{currentfill}%
\pgfsetfillopacity{0.800000}%
\pgfsetlinewidth{0.501875pt}%
\definecolor{currentstroke}{rgb}{0.800000,0.800000,0.800000}%
\pgfsetstrokecolor{currentstroke}%
\pgfsetstrokeopacity{0.800000}%
\pgfsetdash{}{0pt}%
\pgfpathmoveto{\pgfqpoint{0.119444in}{0.100000in}}%
\pgfpathlineto{\pgfqpoint{2.966926in}{0.100000in}}%
\pgfpathquadraticcurveto{\pgfqpoint{2.986370in}{0.100000in}}{\pgfqpoint{2.986370in}{0.119444in}}%
\pgfpathlineto{\pgfqpoint{2.986370in}{0.255556in}}%
\pgfpathquadraticcurveto{\pgfqpoint{2.986370in}{0.275000in}}{\pgfqpoint{2.966926in}{0.275000in}}%
\pgfpathlineto{\pgfqpoint{0.119444in}{0.275000in}}%
\pgfpathquadraticcurveto{\pgfqpoint{0.100000in}{0.275000in}}{\pgfqpoint{0.100000in}{0.255556in}}%
\pgfpathlineto{\pgfqpoint{0.100000in}{0.119444in}}%
\pgfpathquadraticcurveto{\pgfqpoint{0.100000in}{0.100000in}}{\pgfqpoint{0.119444in}{0.100000in}}%
\pgfpathlineto{\pgfqpoint{0.119444in}{0.100000in}}%
\pgfpathclose%
\pgfusepath{stroke,fill}%
\end{pgfscope}%
\begin{pgfscope}%
\pgfsetrectcap%
\pgfsetroundjoin%
\pgfsetlinewidth{0.702625pt}%
\definecolor{currentstroke}{rgb}{0.121569,0.466667,0.705882}%
\pgfsetstrokecolor{currentstroke}%
\pgfsetdash{}{0pt}%
\pgfpathmoveto{\pgfqpoint{0.138889in}{0.197222in}}%
\pgfpathlineto{\pgfqpoint{0.236111in}{0.197222in}}%
\pgfpathlineto{\pgfqpoint{0.333333in}{0.197222in}}%
\pgfusepath{stroke}%
\end{pgfscope}%
\begin{pgfscope}%
\pgfsetbuttcap%
\pgfsetroundjoin%
\definecolor{currentfill}{rgb}{0.121569,0.466667,0.705882}%
\pgfsetfillcolor{currentfill}%
\pgfsetlinewidth{1.003750pt}%
\definecolor{currentstroke}{rgb}{0.121569,0.466667,0.705882}%
\pgfsetstrokecolor{currentstroke}%
\pgfsetdash{}{0pt}%
\pgfsys@defobject{currentmarker}{\pgfqpoint{-0.013889in}{-0.013889in}}{\pgfqpoint{0.013889in}{0.013889in}}{%
\pgfpathmoveto{\pgfqpoint{0.000000in}{-0.013889in}}%
\pgfpathcurveto{\pgfqpoint{0.003683in}{-0.013889in}}{\pgfqpoint{0.007216in}{-0.012425in}}{\pgfqpoint{0.009821in}{-0.009821in}}%
\pgfpathcurveto{\pgfqpoint{0.012425in}{-0.007216in}}{\pgfqpoint{0.013889in}{-0.003683in}}{\pgfqpoint{0.013889in}{0.000000in}}%
\pgfpathcurveto{\pgfqpoint{0.013889in}{0.003683in}}{\pgfqpoint{0.012425in}{0.007216in}}{\pgfqpoint{0.009821in}{0.009821in}}%
\pgfpathcurveto{\pgfqpoint{0.007216in}{0.012425in}}{\pgfqpoint{0.003683in}{0.013889in}}{\pgfqpoint{0.000000in}{0.013889in}}%
\pgfpathcurveto{\pgfqpoint{-0.003683in}{0.013889in}}{\pgfqpoint{-0.007216in}{0.012425in}}{\pgfqpoint{-0.009821in}{0.009821in}}%
\pgfpathcurveto{\pgfqpoint{-0.012425in}{0.007216in}}{\pgfqpoint{-0.013889in}{0.003683in}}{\pgfqpoint{-0.013889in}{0.000000in}}%
\pgfpathcurveto{\pgfqpoint{-0.013889in}{-0.003683in}}{\pgfqpoint{-0.012425in}{-0.007216in}}{\pgfqpoint{-0.009821in}{-0.009821in}}%
\pgfpathcurveto{\pgfqpoint{-0.007216in}{-0.012425in}}{\pgfqpoint{-0.003683in}{-0.013889in}}{\pgfqpoint{0.000000in}{-0.013889in}}%
\pgfpathlineto{\pgfqpoint{0.000000in}{-0.013889in}}%
\pgfpathclose%
\pgfusepath{stroke,fill}%
}%
\begin{pgfscope}%
\pgfsys@transformshift{0.236111in}{0.197222in}%
\pgfsys@useobject{currentmarker}{}%
\end{pgfscope}%
\end{pgfscope}%
\begin{pgfscope}%
\definecolor{textcolor}{rgb}{0.000000,0.000000,0.000000}%
\pgfsetstrokecolor{textcolor}%
\pgfsetfillcolor{textcolor}%
\pgftext[x=0.411111in,y=0.163194in,left,base]{\color{textcolor}{\rmfamily\fontsize{7.000000}{8.400000}\selectfont\catcode`\^=\active\def^{\ifmmode\sp\else\^{}\fi}\catcode`\%=\active\def
\end{pgfscope}%
\begin{pgfscope}%
\pgfsetrectcap%
\pgfsetroundjoin%
\pgfsetlinewidth{0.702625pt}%
\definecolor{currentstroke}{rgb}{1.000000,0.498039,0.054902}%
\pgfsetstrokecolor{currentstroke}%
\pgfsetdash{}{0pt}%
\pgfpathmoveto{\pgfqpoint{1.376702in}{0.197222in}}%
\pgfpathlineto{\pgfqpoint{1.473924in}{0.197222in}}%
\pgfpathlineto{\pgfqpoint{1.571146in}{0.197222in}}%
\pgfusepath{stroke}%
\end{pgfscope}%
\begin{pgfscope}%
\pgfsetbuttcap%
\pgfsetmiterjoin%
\definecolor{currentfill}{rgb}{1.000000,0.498039,0.054902}%
\pgfsetfillcolor{currentfill}%
\pgfsetlinewidth{1.003750pt}%
\definecolor{currentstroke}{rgb}{1.000000,0.498039,0.054902}%
\pgfsetstrokecolor{currentstroke}%
\pgfsetdash{}{0pt}%
\pgfsys@defobject{currentmarker}{\pgfqpoint{-0.013889in}{-0.013889in}}{\pgfqpoint{0.013889in}{0.013889in}}{%
\pgfpathmoveto{\pgfqpoint{-0.013889in}{-0.013889in}}%
\pgfpathlineto{\pgfqpoint{0.013889in}{-0.013889in}}%
\pgfpathlineto{\pgfqpoint{0.013889in}{0.013889in}}%
\pgfpathlineto{\pgfqpoint{-0.013889in}{0.013889in}}%
\pgfpathlineto{\pgfqpoint{-0.013889in}{-0.013889in}}%
\pgfpathclose%
\pgfusepath{stroke,fill}%
}%
\begin{pgfscope}%
\pgfsys@transformshift{1.473924in}{0.197222in}%
\pgfsys@useobject{currentmarker}{}%
\end{pgfscope}%
\end{pgfscope}%
\begin{pgfscope}%
\definecolor{textcolor}{rgb}{0.000000,0.000000,0.000000}%
\pgfsetstrokecolor{textcolor}%
\pgfsetfillcolor{textcolor}%
\pgftext[x=1.648924in,y=0.163194in,left,base]{\color{textcolor}{\rmfamily\fontsize{7.000000}{8.400000}\selectfont\catcode`\^=\active\def^{\ifmmode\sp\else\^{}\fi}\catcode`\%=\active\def
\end{pgfscope}%
\begin{pgfscope}%
\pgfsetrectcap%
\pgfsetroundjoin%
\pgfsetlinewidth{0.702625pt}%
\definecolor{currentstroke}{rgb}{0.172549,0.627451,0.172549}%
\pgfsetstrokecolor{currentstroke}%
\pgfsetdash{}{0pt}%
\pgfpathmoveto{\pgfqpoint{2.186059in}{0.197222in}}%
\pgfpathlineto{\pgfqpoint{2.283282in}{0.197222in}}%
\pgfpathlineto{\pgfqpoint{2.380504in}{0.197222in}}%
\pgfusepath{stroke}%
\end{pgfscope}%
\begin{pgfscope}%
\pgfsetbuttcap%
\pgfsetmiterjoin%
\definecolor{currentfill}{rgb}{0.172549,0.627451,0.172549}%
\pgfsetfillcolor{currentfill}%
\pgfsetlinewidth{1.003750pt}%
\definecolor{currentstroke}{rgb}{0.172549,0.627451,0.172549}%
\pgfsetstrokecolor{currentstroke}%
\pgfsetdash{}{0pt}%
\pgfsys@defobject{currentmarker}{\pgfqpoint{-0.013889in}{-0.013889in}}{\pgfqpoint{0.013889in}{0.013889in}}{%
\pgfpathmoveto{\pgfqpoint{0.000000in}{0.013889in}}%
\pgfpathlineto{\pgfqpoint{-0.013889in}{-0.013889in}}%
\pgfpathlineto{\pgfqpoint{0.013889in}{-0.013889in}}%
\pgfpathlineto{\pgfqpoint{0.000000in}{0.013889in}}%
\pgfpathclose%
\pgfusepath{stroke,fill}%
}%
\begin{pgfscope}%
\pgfsys@transformshift{2.283282in}{0.197222in}%
\pgfsys@useobject{currentmarker}{}%
\end{pgfscope}%
\end{pgfscope}%
\begin{pgfscope}%
\definecolor{textcolor}{rgb}{0.000000,0.000000,0.000000}%
\pgfsetstrokecolor{textcolor}%
\pgfsetfillcolor{textcolor}%
\pgftext[x=2.458282in,y=0.163194in,left,base]{\color{textcolor}{\rmfamily\fontsize{7.000000}{8.400000}\selectfont\catcode`\^=\active\def^{\ifmmode\sp\else\^{}\fi}\catcode`\%=\active\def
\end{pgfscope}%
\end{pgfpicture}%
\makeatother%
\endgroup%

%% file: img/granularity.pgf
\begingroup%
\makeatletter%
\begin{pgfpicture}%
\pgfpathrectangle{\pgfpointorigin}{\pgfqpoint{2.910059in}{1.303473in}}%
\pgfusepath{use as bounding box, clip}%
\begin{pgfscope}%
\pgfsetbuttcap%
\pgfsetmiterjoin%
\definecolor{currentfill}{rgb}{1.000000,1.000000,1.000000}%
\pgfsetfillcolor{currentfill}%
\pgfsetlinewidth{0.000000pt}%
\definecolor{currentstroke}{rgb}{0.500000,0.500000,0.500000}%
\pgfsetstrokecolor{currentstroke}%
\pgfsetdash{}{0pt}%
\pgfpathmoveto{\pgfqpoint{0.000000in}{-0.000000in}}%
\pgfpathlineto{\pgfqpoint{2.910059in}{-0.000000in}}%
\pgfpathlineto{\pgfqpoint{2.910059in}{1.303473in}}%
\pgfpathlineto{\pgfqpoint{0.000000in}{1.303473in}}%
\pgfpathlineto{\pgfqpoint{0.000000in}{-0.000000in}}%
\pgfpathclose%
\pgfusepath{fill}%
\end{pgfscope}%
\begin{pgfscope}%
\pgfsetbuttcap%
\pgfsetmiterjoin%
\definecolor{currentfill}{rgb}{0.898039,0.898039,0.898039}%
\pgfsetfillcolor{currentfill}%
\pgfsetlinewidth{0.000000pt}%
\definecolor{currentstroke}{rgb}{0.000000,0.000000,0.000000}%
\pgfsetstrokecolor{currentstroke}%
\pgfsetstrokeopacity{0.000000}%
\pgfsetdash{}{0pt}%
\pgfpathmoveto{\pgfqpoint{0.307948in}{0.563553in}}%
\pgfpathlineto{\pgfqpoint{0.664936in}{0.563553in}}%
\pgfpathlineto{\pgfqpoint{0.664936in}{1.169715in}}%
\pgfpathlineto{\pgfqpoint{0.307948in}{1.169715in}}%
\pgfpathlineto{\pgfqpoint{0.307948in}{0.563553in}}%
\pgfpathclose%
\pgfusepath{fill}%
\end{pgfscope}%
\begin{pgfscope}%
\definecolor{textcolor}{rgb}{0.333333,0.333333,0.333333}%
\pgfsetstrokecolor{textcolor}%
\pgfsetfillcolor{textcolor}%
\pgftext[x=0.486442in,y=0.507998in,,top]{\color{textcolor}{\rmfamily\fontsize{9.000000}{10.800000}\selectfont\catcode`\^=\active\def^{\ifmmode\sp\else\^{}\fi}\catcode`\%=\active\def
\end{pgfscope}%
\begin{pgfscope}%
\pgfpathrectangle{\pgfqpoint{0.307948in}{0.563553in}}{\pgfqpoint{0.356988in}{0.606162in}}%
\pgfusepath{clip}%
\pgfsetrectcap%
\pgfsetroundjoin%
\pgfsetlinewidth{0.803000pt}%
\definecolor{currentstroke}{rgb}{1.000000,1.000000,1.000000}%
\pgfsetstrokecolor{currentstroke}%
\pgfsetdash{}{0pt}%
\pgfpathmoveto{\pgfqpoint{0.307948in}{0.691166in}}%
\pgfpathlineto{\pgfqpoint{0.664936in}{0.691166in}}%
\pgfusepath{stroke}%
\end{pgfscope}%
\begin{pgfscope}%
\pgfsetbuttcap%
\pgfsetroundjoin%
\definecolor{currentfill}{rgb}{0.333333,0.333333,0.333333}%
\pgfsetfillcolor{currentfill}%
\pgfsetlinewidth{0.803000pt}%
\definecolor{currentstroke}{rgb}{0.333333,0.333333,0.333333}%
\pgfsetstrokecolor{currentstroke}%
\pgfsetdash{}{0pt}%
\pgfsys@defobject{currentmarker}{\pgfqpoint{-0.048611in}{0.000000in}}{\pgfqpoint{-0.000000in}{0.000000in}}{%
\pgfpathmoveto{\pgfqpoint{-0.000000in}{0.000000in}}%
\pgfpathlineto{\pgfqpoint{-0.048611in}{0.000000in}}%
\pgfusepath{stroke,fill}%
}%
\begin{pgfscope}%
\pgfsys@transformshift{0.307948in}{0.691166in}%
\pgfsys@useobject{currentmarker}{}%
\end{pgfscope}%
\end{pgfscope}%
\begin{pgfscope}%
\definecolor{textcolor}{rgb}{0.333333,0.333333,0.333333}%
\pgfsetstrokecolor{textcolor}%
\pgfsetfillcolor{textcolor}%
\pgftext[x=0.100000in, y=0.657409in, left, base]{\color{textcolor}{\rmfamily\fontsize{7.000000}{8.400000}\selectfont\catcode`\^=\active\def^{\ifmmode\sp\else\^{}\fi}\catcode`\%=\active\def
\end{pgfscope}%
\begin{pgfscope}%
\pgfpathrectangle{\pgfqpoint{0.307948in}{0.563553in}}{\pgfqpoint{0.356988in}{0.606162in}}%
\pgfusepath{clip}%
\pgfsetrectcap%
\pgfsetroundjoin%
\pgfsetlinewidth{0.803000pt}%
\definecolor{currentstroke}{rgb}{1.000000,1.000000,1.000000}%
\pgfsetstrokecolor{currentstroke}%
\pgfsetdash{}{0pt}%
\pgfpathmoveto{\pgfqpoint{0.307948in}{1.089957in}}%
\pgfpathlineto{\pgfqpoint{0.664936in}{1.089957in}}%
\pgfusepath{stroke}%
\end{pgfscope}%
\begin{pgfscope}%
\pgfsetbuttcap%
\pgfsetroundjoin%
\definecolor{currentfill}{rgb}{0.333333,0.333333,0.333333}%
\pgfsetfillcolor{currentfill}%
\pgfsetlinewidth{0.803000pt}%
\definecolor{currentstroke}{rgb}{0.333333,0.333333,0.333333}%
\pgfsetstrokecolor{currentstroke}%
\pgfsetdash{}{0pt}%
\pgfsys@defobject{currentmarker}{\pgfqpoint{-0.048611in}{0.000000in}}{\pgfqpoint{-0.000000in}{0.000000in}}{%
\pgfpathmoveto{\pgfqpoint{-0.000000in}{0.000000in}}%
\pgfpathlineto{\pgfqpoint{-0.048611in}{0.000000in}}%
\pgfusepath{stroke,fill}%
}%
\begin{pgfscope}%
\pgfsys@transformshift{0.307948in}{1.089957in}%
\pgfsys@useobject{currentmarker}{}%
\end{pgfscope}%
\end{pgfscope}%
\begin{pgfscope}%
\definecolor{textcolor}{rgb}{0.333333,0.333333,0.333333}%
\pgfsetstrokecolor{textcolor}%
\pgfsetfillcolor{textcolor}%
\pgftext[x=0.100000in, y=1.056199in, left, base]{\color{textcolor}{\rmfamily\fontsize{7.000000}{8.400000}\selectfont\catcode`\^=\active\def^{\ifmmode\sp\else\^{}\fi}\catcode`\%=\active\def
\end{pgfscope}%
\begin{pgfscope}%
\pgfpathrectangle{\pgfqpoint{0.307948in}{0.563553in}}{\pgfqpoint{0.356988in}{0.606162in}}%
\pgfusepath{clip}%
\pgfsetbuttcap%
\pgfsetmiterjoin%
\definecolor{currentfill}{rgb}{0.121569,0.466667,0.705882}%
\pgfsetfillcolor{currentfill}%
\pgfsetlinewidth{0.000000pt}%
\definecolor{currentstroke}{rgb}{0.000000,0.000000,0.000000}%
\pgfsetstrokecolor{currentstroke}%
\pgfsetstrokeopacity{0.000000}%
\pgfsetdash{}{0pt}%
\pgfpathmoveto{\pgfqpoint{0.380196in}{-2.499159in}}%
\pgfpathlineto{\pgfqpoint{0.422694in}{-2.499159in}}%
\pgfpathlineto{\pgfqpoint{0.422694in}{0.715094in}}%
\pgfpathlineto{\pgfqpoint{0.380196in}{0.715094in}}%
\pgfpathlineto{\pgfqpoint{0.380196in}{-2.499159in}}%
\pgfpathclose%
\pgfusepath{fill}%
\end{pgfscope}%
\begin{pgfscope}%
\pgfpathrectangle{\pgfqpoint{0.307948in}{0.563553in}}{\pgfqpoint{0.356988in}{0.606162in}}%
\pgfusepath{clip}%
\pgfsetbuttcap%
\pgfsetmiterjoin%
\definecolor{currentfill}{rgb}{1.000000,0.498039,0.054902}%
\pgfsetfillcolor{currentfill}%
\pgfsetlinewidth{0.000000pt}%
\definecolor{currentstroke}{rgb}{0.000000,0.000000,0.000000}%
\pgfsetstrokecolor{currentstroke}%
\pgfsetstrokeopacity{0.000000}%
\pgfsetdash{}{0pt}%
\pgfpathmoveto{\pgfqpoint{0.465193in}{-2.499159in}}%
\pgfpathlineto{\pgfqpoint{0.507691in}{-2.499159in}}%
\pgfpathlineto{\pgfqpoint{0.507691in}{0.842707in}}%
\pgfpathlineto{\pgfqpoint{0.465193in}{0.842707in}}%
\pgfpathlineto{\pgfqpoint{0.465193in}{-2.499159in}}%
\pgfpathclose%
\pgfusepath{fill}%
\end{pgfscope}%
\begin{pgfscope}%
\pgfpathrectangle{\pgfqpoint{0.307948in}{0.563553in}}{\pgfqpoint{0.356988in}{0.606162in}}%
\pgfusepath{clip}%
\pgfsetbuttcap%
\pgfsetmiterjoin%
\definecolor{currentfill}{rgb}{0.172549,0.627451,0.172549}%
\pgfsetfillcolor{currentfill}%
\pgfsetlinewidth{0.000000pt}%
\definecolor{currentstroke}{rgb}{0.000000,0.000000,0.000000}%
\pgfsetstrokecolor{currentstroke}%
\pgfsetstrokeopacity{0.000000}%
\pgfsetdash{}{0pt}%
\pgfpathmoveto{\pgfqpoint{0.550190in}{-2.499159in}}%
\pgfpathlineto{\pgfqpoint{0.592688in}{-2.499159in}}%
\pgfpathlineto{\pgfqpoint{0.592688in}{1.018175in}}%
\pgfpathlineto{\pgfqpoint{0.550190in}{1.018175in}}%
\pgfpathlineto{\pgfqpoint{0.550190in}{-2.499159in}}%
\pgfpathclose%
\pgfusepath{fill}%
\end{pgfscope}%
\begin{pgfscope}%
\pgfsetrectcap%
\pgfsetmiterjoin%
\pgfsetlinewidth{1.003750pt}%
\definecolor{currentstroke}{rgb}{1.000000,1.000000,1.000000}%
\pgfsetstrokecolor{currentstroke}%
\pgfsetdash{}{0pt}%
\pgfpathmoveto{\pgfqpoint{0.307948in}{0.563553in}}%
\pgfpathlineto{\pgfqpoint{0.307948in}{1.169715in}}%
\pgfusepath{stroke}%
\end{pgfscope}%
\begin{pgfscope}%
\pgfsetrectcap%
\pgfsetmiterjoin%
\pgfsetlinewidth{1.003750pt}%
\definecolor{currentstroke}{rgb}{1.000000,1.000000,1.000000}%
\pgfsetstrokecolor{currentstroke}%
\pgfsetdash{}{0pt}%
\pgfpathmoveto{\pgfqpoint{0.664936in}{0.563553in}}%
\pgfpathlineto{\pgfqpoint{0.664936in}{1.169715in}}%
\pgfusepath{stroke}%
\end{pgfscope}%
\begin{pgfscope}%
\pgfsetrectcap%
\pgfsetmiterjoin%
\pgfsetlinewidth{1.003750pt}%
\definecolor{currentstroke}{rgb}{1.000000,1.000000,1.000000}%
\pgfsetstrokecolor{currentstroke}%
\pgfsetdash{}{0pt}%
\pgfpathmoveto{\pgfqpoint{0.307948in}{0.563553in}}%
\pgfpathlineto{\pgfqpoint{0.664936in}{0.563553in}}%
\pgfusepath{stroke}%
\end{pgfscope}%
\begin{pgfscope}%
\pgfsetrectcap%
\pgfsetmiterjoin%
\pgfsetlinewidth{1.003750pt}%
\definecolor{currentstroke}{rgb}{1.000000,1.000000,1.000000}%
\pgfsetstrokecolor{currentstroke}%
\pgfsetdash{}{0pt}%
\pgfpathmoveto{\pgfqpoint{0.307948in}{1.169715in}}%
\pgfpathlineto{\pgfqpoint{0.664936in}{1.169715in}}%
\pgfusepath{stroke}%
\end{pgfscope}%
\begin{pgfscope}%
\pgfsetbuttcap%
\pgfsetmiterjoin%
\definecolor{currentfill}{rgb}{0.898039,0.898039,0.898039}%
\pgfsetfillcolor{currentfill}%
\pgfsetlinewidth{0.000000pt}%
\definecolor{currentstroke}{rgb}{0.000000,0.000000,0.000000}%
\pgfsetstrokecolor{currentstroke}%
\pgfsetstrokeopacity{0.000000}%
\pgfsetdash{}{0pt}%
\pgfpathmoveto{\pgfqpoint{1.022989in}{0.563553in}}%
\pgfpathlineto{\pgfqpoint{1.379977in}{0.563553in}}%
\pgfpathlineto{\pgfqpoint{1.379977in}{1.169715in}}%
\pgfpathlineto{\pgfqpoint{1.022989in}{1.169715in}}%
\pgfpathlineto{\pgfqpoint{1.022989in}{0.563553in}}%
\pgfpathclose%
\pgfusepath{fill}%
\end{pgfscope}%
\begin{pgfscope}%
\definecolor{textcolor}{rgb}{0.333333,0.333333,0.333333}%
\pgfsetstrokecolor{textcolor}%
\pgfsetfillcolor{textcolor}%
\pgftext[x=1.201483in,y=0.507998in,,top]{\color{textcolor}{\rmfamily\fontsize{9.000000}{10.800000}\selectfont\catcode`\^=\active\def^{\ifmmode\sp\else\^{}\fi}\catcode`\%=\active\def
\end{pgfscope}%
\begin{pgfscope}%
\pgfpathrectangle{\pgfqpoint{1.022989in}{0.563553in}}{\pgfqpoint{0.356988in}{0.606162in}}%
\pgfusepath{clip}%
\pgfsetrectcap%
\pgfsetroundjoin%
\pgfsetlinewidth{0.803000pt}%
\definecolor{currentstroke}{rgb}{1.000000,1.000000,1.000000}%
\pgfsetstrokecolor{currentstroke}%
\pgfsetdash{}{0pt}%
\pgfpathmoveto{\pgfqpoint{1.022989in}{0.619384in}}%
\pgfpathlineto{\pgfqpoint{1.379977in}{0.619384in}}%
\pgfusepath{stroke}%
\end{pgfscope}%
\begin{pgfscope}%
\pgfsetbuttcap%
\pgfsetroundjoin%
\definecolor{currentfill}{rgb}{0.333333,0.333333,0.333333}%
\pgfsetfillcolor{currentfill}%
\pgfsetlinewidth{0.803000pt}%
\definecolor{currentstroke}{rgb}{0.333333,0.333333,0.333333}%
\pgfsetstrokecolor{currentstroke}%
\pgfsetdash{}{0pt}%
\pgfsys@defobject{currentmarker}{\pgfqpoint{-0.048611in}{0.000000in}}{\pgfqpoint{-0.000000in}{0.000000in}}{%
\pgfpathmoveto{\pgfqpoint{-0.000000in}{0.000000in}}%
\pgfpathlineto{\pgfqpoint{-0.048611in}{0.000000in}}%
\pgfusepath{stroke,fill}%
}%
\begin{pgfscope}%
\pgfsys@transformshift{1.022989in}{0.619384in}%
\pgfsys@useobject{currentmarker}{}%
\end{pgfscope}%
\end{pgfscope}%
\begin{pgfscope}%
\definecolor{textcolor}{rgb}{0.333333,0.333333,0.333333}%
\pgfsetstrokecolor{textcolor}%
\pgfsetfillcolor{textcolor}%
\pgftext[x=0.726692in, y=0.585626in, left, base]{\color{textcolor}{\rmfamily\fontsize{7.000000}{8.400000}\selectfont\catcode`\^=\active\def^{\ifmmode\sp\else\^{}\fi}\catcode`\%=\active\def
\end{pgfscope}%
\begin{pgfscope}%
\pgfpathrectangle{\pgfqpoint{1.022989in}{0.563553in}}{\pgfqpoint{0.356988in}{0.606162in}}%
\pgfusepath{clip}%
\pgfsetrectcap%
\pgfsetroundjoin%
\pgfsetlinewidth{0.803000pt}%
\definecolor{currentstroke}{rgb}{1.000000,1.000000,1.000000}%
\pgfsetstrokecolor{currentstroke}%
\pgfsetdash{}{0pt}%
\pgfpathmoveto{\pgfqpoint{1.022989in}{0.818779in}}%
\pgfpathlineto{\pgfqpoint{1.379977in}{0.818779in}}%
\pgfusepath{stroke}%
\end{pgfscope}%
\begin{pgfscope}%
\pgfsetbuttcap%
\pgfsetroundjoin%
\definecolor{currentfill}{rgb}{0.333333,0.333333,0.333333}%
\pgfsetfillcolor{currentfill}%
\pgfsetlinewidth{0.803000pt}%
\definecolor{currentstroke}{rgb}{0.333333,0.333333,0.333333}%
\pgfsetstrokecolor{currentstroke}%
\pgfsetdash{}{0pt}%
\pgfsys@defobject{currentmarker}{\pgfqpoint{-0.048611in}{0.000000in}}{\pgfqpoint{-0.000000in}{0.000000in}}{%
\pgfpathmoveto{\pgfqpoint{-0.000000in}{0.000000in}}%
\pgfpathlineto{\pgfqpoint{-0.048611in}{0.000000in}}%
\pgfusepath{stroke,fill}%
}%
\begin{pgfscope}%
\pgfsys@transformshift{1.022989in}{0.818779in}%
\pgfsys@useobject{currentmarker}{}%
\end{pgfscope}%
\end{pgfscope}%
\begin{pgfscope}%
\definecolor{textcolor}{rgb}{0.333333,0.333333,0.333333}%
\pgfsetstrokecolor{textcolor}%
\pgfsetfillcolor{textcolor}%
\pgftext[x=0.726692in, y=0.785022in, left, base]{\color{textcolor}{\rmfamily\fontsize{7.000000}{8.400000}\selectfont\catcode`\^=\active\def^{\ifmmode\sp\else\^{}\fi}\catcode`\%=\active\def
\end{pgfscope}%
\begin{pgfscope}%
\pgfpathrectangle{\pgfqpoint{1.022989in}{0.563553in}}{\pgfqpoint{0.356988in}{0.606162in}}%
\pgfusepath{clip}%
\pgfsetrectcap%
\pgfsetroundjoin%
\pgfsetlinewidth{0.803000pt}%
\definecolor{currentstroke}{rgb}{1.000000,1.000000,1.000000}%
\pgfsetstrokecolor{currentstroke}%
\pgfsetdash{}{0pt}%
\pgfpathmoveto{\pgfqpoint{1.022989in}{1.018175in}}%
\pgfpathlineto{\pgfqpoint{1.379977in}{1.018175in}}%
\pgfusepath{stroke}%
\end{pgfscope}%
\begin{pgfscope}%
\pgfsetbuttcap%
\pgfsetroundjoin%
\definecolor{currentfill}{rgb}{0.333333,0.333333,0.333333}%
\pgfsetfillcolor{currentfill}%
\pgfsetlinewidth{0.803000pt}%
\definecolor{currentstroke}{rgb}{0.333333,0.333333,0.333333}%
\pgfsetstrokecolor{currentstroke}%
\pgfsetdash{}{0pt}%
\pgfsys@defobject{currentmarker}{\pgfqpoint{-0.048611in}{0.000000in}}{\pgfqpoint{-0.000000in}{0.000000in}}{%
\pgfpathmoveto{\pgfqpoint{-0.000000in}{0.000000in}}%
\pgfpathlineto{\pgfqpoint{-0.048611in}{0.000000in}}%
\pgfusepath{stroke,fill}%
}%
\begin{pgfscope}%
\pgfsys@transformshift{1.022989in}{1.018175in}%
\pgfsys@useobject{currentmarker}{}%
\end{pgfscope}%
\end{pgfscope}%
\begin{pgfscope}%
\definecolor{textcolor}{rgb}{0.333333,0.333333,0.333333}%
\pgfsetstrokecolor{textcolor}%
\pgfsetfillcolor{textcolor}%
\pgftext[x=0.726692in, y=0.984417in, left, base]{\color{textcolor}{\rmfamily\fontsize{7.000000}{8.400000}\selectfont\catcode`\^=\active\def^{\ifmmode\sp\else\^{}\fi}\catcode`\%=\active\def
\end{pgfscope}%
\begin{pgfscope}%
\pgfpathrectangle{\pgfqpoint{1.022989in}{0.563553in}}{\pgfqpoint{0.356988in}{0.606162in}}%
\pgfusepath{clip}%
\pgfsetbuttcap%
\pgfsetmiterjoin%
\definecolor{currentfill}{rgb}{0.121569,0.466667,0.705882}%
\pgfsetfillcolor{currentfill}%
\pgfsetlinewidth{0.000000pt}%
\definecolor{currentstroke}{rgb}{0.000000,0.000000,0.000000}%
\pgfsetstrokecolor{currentstroke}%
\pgfsetstrokeopacity{0.000000}%
\pgfsetdash{}{0pt}%
\pgfpathmoveto{\pgfqpoint{1.095237in}{-3.966709in}}%
\pgfpathlineto{\pgfqpoint{1.137735in}{-3.966709in}}%
\pgfpathlineto{\pgfqpoint{1.137735in}{0.715094in}}%
\pgfpathlineto{\pgfqpoint{1.095237in}{0.715094in}}%
\pgfpathlineto{\pgfqpoint{1.095237in}{-3.966709in}}%
\pgfpathclose%
\pgfusepath{fill}%
\end{pgfscope}%
\begin{pgfscope}%
\pgfpathrectangle{\pgfqpoint{1.022989in}{0.563553in}}{\pgfqpoint{0.356988in}{0.606162in}}%
\pgfusepath{clip}%
\pgfsetbuttcap%
\pgfsetmiterjoin%
\definecolor{currentfill}{rgb}{1.000000,0.498039,0.054902}%
\pgfsetfillcolor{currentfill}%
\pgfsetlinewidth{0.000000pt}%
\definecolor{currentstroke}{rgb}{0.000000,0.000000,0.000000}%
\pgfsetstrokecolor{currentstroke}%
\pgfsetstrokeopacity{0.000000}%
\pgfsetdash{}{0pt}%
\pgfpathmoveto{\pgfqpoint{1.180234in}{-3.966709in}}%
\pgfpathlineto{\pgfqpoint{1.222732in}{-3.966709in}}%
\pgfpathlineto{\pgfqpoint{1.222732in}{0.874610in}}%
\pgfpathlineto{\pgfqpoint{1.180234in}{0.874610in}}%
\pgfpathlineto{\pgfqpoint{1.180234in}{-3.966709in}}%
\pgfpathclose%
\pgfusepath{fill}%
\end{pgfscope}%
\begin{pgfscope}%
\pgfpathrectangle{\pgfqpoint{1.022989in}{0.563553in}}{\pgfqpoint{0.356988in}{0.606162in}}%
\pgfusepath{clip}%
\pgfsetbuttcap%
\pgfsetmiterjoin%
\definecolor{currentfill}{rgb}{0.172549,0.627451,0.172549}%
\pgfsetfillcolor{currentfill}%
\pgfsetlinewidth{0.000000pt}%
\definecolor{currentstroke}{rgb}{0.000000,0.000000,0.000000}%
\pgfsetstrokecolor{currentstroke}%
\pgfsetstrokeopacity{0.000000}%
\pgfsetdash{}{0pt}%
\pgfpathmoveto{\pgfqpoint{1.265231in}{-3.966709in}}%
\pgfpathlineto{\pgfqpoint{1.307730in}{-3.966709in}}%
\pgfpathlineto{\pgfqpoint{1.307730in}{1.018175in}}%
\pgfpathlineto{\pgfqpoint{1.265231in}{1.018175in}}%
\pgfpathlineto{\pgfqpoint{1.265231in}{-3.966709in}}%
\pgfpathclose%
\pgfusepath{fill}%
\end{pgfscope}%
\begin{pgfscope}%
\pgfsetrectcap%
\pgfsetmiterjoin%
\pgfsetlinewidth{1.003750pt}%
\definecolor{currentstroke}{rgb}{1.000000,1.000000,1.000000}%
\pgfsetstrokecolor{currentstroke}%
\pgfsetdash{}{0pt}%
\pgfpathmoveto{\pgfqpoint{1.022989in}{0.563553in}}%
\pgfpathlineto{\pgfqpoint{1.022989in}{1.169715in}}%
\pgfusepath{stroke}%
\end{pgfscope}%
\begin{pgfscope}%
\pgfsetrectcap%
\pgfsetmiterjoin%
\pgfsetlinewidth{1.003750pt}%
\definecolor{currentstroke}{rgb}{1.000000,1.000000,1.000000}%
\pgfsetstrokecolor{currentstroke}%
\pgfsetdash{}{0pt}%
\pgfpathmoveto{\pgfqpoint{1.379977in}{0.563553in}}%
\pgfpathlineto{\pgfqpoint{1.379977in}{1.169715in}}%
\pgfusepath{stroke}%
\end{pgfscope}%
\begin{pgfscope}%
\pgfsetrectcap%
\pgfsetmiterjoin%
\pgfsetlinewidth{1.003750pt}%
\definecolor{currentstroke}{rgb}{1.000000,1.000000,1.000000}%
\pgfsetstrokecolor{currentstroke}%
\pgfsetdash{}{0pt}%
\pgfpathmoveto{\pgfqpoint{1.022989in}{0.563553in}}%
\pgfpathlineto{\pgfqpoint{1.379977in}{0.563553in}}%
\pgfusepath{stroke}%
\end{pgfscope}%
\begin{pgfscope}%
\pgfsetrectcap%
\pgfsetmiterjoin%
\pgfsetlinewidth{1.003750pt}%
\definecolor{currentstroke}{rgb}{1.000000,1.000000,1.000000}%
\pgfsetstrokecolor{currentstroke}%
\pgfsetdash{}{0pt}%
\pgfpathmoveto{\pgfqpoint{1.022989in}{1.169715in}}%
\pgfpathlineto{\pgfqpoint{1.379977in}{1.169715in}}%
\pgfusepath{stroke}%
\end{pgfscope}%
\begin{pgfscope}%
\pgfsetbuttcap%
\pgfsetmiterjoin%
\definecolor{currentfill}{rgb}{0.898039,0.898039,0.898039}%
\pgfsetfillcolor{currentfill}%
\pgfsetlinewidth{0.000000pt}%
\definecolor{currentstroke}{rgb}{0.000000,0.000000,0.000000}%
\pgfsetstrokecolor{currentstroke}%
\pgfsetstrokeopacity{0.000000}%
\pgfsetdash{}{0pt}%
\pgfpathmoveto{\pgfqpoint{1.738030in}{0.563553in}}%
\pgfpathlineto{\pgfqpoint{2.095018in}{0.563553in}}%
\pgfpathlineto{\pgfqpoint{2.095018in}{1.169715in}}%
\pgfpathlineto{\pgfqpoint{1.738030in}{1.169715in}}%
\pgfpathlineto{\pgfqpoint{1.738030in}{0.563553in}}%
\pgfpathclose%
\pgfusepath{fill}%
\end{pgfscope}%
\begin{pgfscope}%
\definecolor{textcolor}{rgb}{0.333333,0.333333,0.333333}%
\pgfsetstrokecolor{textcolor}%
\pgfsetfillcolor{textcolor}%
\pgftext[x=1.916524in,y=0.507998in,,top]{\color{textcolor}{\rmfamily\fontsize{9.000000}{10.800000}\selectfont\catcode`\^=\active\def^{\ifmmode\sp\else\^{}\fi}\catcode`\%=\active\def
\end{pgfscope}%
\begin{pgfscope}%
\pgfpathrectangle{\pgfqpoint{1.738030in}{0.563553in}}{\pgfqpoint{0.356988in}{0.606162in}}%
\pgfusepath{clip}%
\pgfsetrectcap%
\pgfsetroundjoin%
\pgfsetlinewidth{0.803000pt}%
\definecolor{currentstroke}{rgb}{1.000000,1.000000,1.000000}%
\pgfsetstrokecolor{currentstroke}%
\pgfsetdash{}{0pt}%
\pgfpathmoveto{\pgfqpoint{1.738030in}{0.703765in}}%
\pgfpathlineto{\pgfqpoint{2.095018in}{0.703765in}}%
\pgfusepath{stroke}%
\end{pgfscope}%
\begin{pgfscope}%
\pgfsetbuttcap%
\pgfsetroundjoin%
\definecolor{currentfill}{rgb}{0.333333,0.333333,0.333333}%
\pgfsetfillcolor{currentfill}%
\pgfsetlinewidth{0.803000pt}%
\definecolor{currentstroke}{rgb}{0.333333,0.333333,0.333333}%
\pgfsetstrokecolor{currentstroke}%
\pgfsetdash{}{0pt}%
\pgfsys@defobject{currentmarker}{\pgfqpoint{-0.048611in}{0.000000in}}{\pgfqpoint{-0.000000in}{0.000000in}}{%
\pgfpathmoveto{\pgfqpoint{-0.000000in}{0.000000in}}%
\pgfpathlineto{\pgfqpoint{-0.048611in}{0.000000in}}%
\pgfusepath{stroke,fill}%
}%
\begin{pgfscope}%
\pgfsys@transformshift{1.738030in}{0.703765in}%
\pgfsys@useobject{currentmarker}{}%
\end{pgfscope}%
\end{pgfscope}%
\begin{pgfscope}%
\definecolor{textcolor}{rgb}{0.333333,0.333333,0.333333}%
\pgfsetstrokecolor{textcolor}%
\pgfsetfillcolor{textcolor}%
\pgftext[x=1.530082in, y=0.670007in, left, base]{\color{textcolor}{\rmfamily\fontsize{7.000000}{8.400000}\selectfont\catcode`\^=\active\def^{\ifmmode\sp\else\^{}\fi}\catcode`\%=\active\def
\end{pgfscope}%
\begin{pgfscope}%
\pgfpathrectangle{\pgfqpoint{1.738030in}{0.563553in}}{\pgfqpoint{0.356988in}{0.606162in}}%
\pgfusepath{clip}%
\pgfsetrectcap%
\pgfsetroundjoin%
\pgfsetlinewidth{0.803000pt}%
\definecolor{currentstroke}{rgb}{1.000000,1.000000,1.000000}%
\pgfsetstrokecolor{currentstroke}%
\pgfsetdash{}{0pt}%
\pgfpathmoveto{\pgfqpoint{1.738030in}{0.976035in}}%
\pgfpathlineto{\pgfqpoint{2.095018in}{0.976035in}}%
\pgfusepath{stroke}%
\end{pgfscope}%
\begin{pgfscope}%
\pgfsetbuttcap%
\pgfsetroundjoin%
\definecolor{currentfill}{rgb}{0.333333,0.333333,0.333333}%
\pgfsetfillcolor{currentfill}%
\pgfsetlinewidth{0.803000pt}%
\definecolor{currentstroke}{rgb}{0.333333,0.333333,0.333333}%
\pgfsetstrokecolor{currentstroke}%
\pgfsetdash{}{0pt}%
\pgfsys@defobject{currentmarker}{\pgfqpoint{-0.048611in}{0.000000in}}{\pgfqpoint{-0.000000in}{0.000000in}}{%
\pgfpathmoveto{\pgfqpoint{-0.000000in}{0.000000in}}%
\pgfpathlineto{\pgfqpoint{-0.048611in}{0.000000in}}%
\pgfusepath{stroke,fill}%
}%
\begin{pgfscope}%
\pgfsys@transformshift{1.738030in}{0.976035in}%
\pgfsys@useobject{currentmarker}{}%
\end{pgfscope}%
\end{pgfscope}%
\begin{pgfscope}%
\definecolor{textcolor}{rgb}{0.333333,0.333333,0.333333}%
\pgfsetstrokecolor{textcolor}%
\pgfsetfillcolor{textcolor}%
\pgftext[x=1.530082in, y=0.942277in, left, base]{\color{textcolor}{\rmfamily\fontsize{7.000000}{8.400000}\selectfont\catcode`\^=\active\def^{\ifmmode\sp\else\^{}\fi}\catcode`\%=\active\def
\end{pgfscope}%
\begin{pgfscope}%
\pgfpathrectangle{\pgfqpoint{1.738030in}{0.563553in}}{\pgfqpoint{0.356988in}{0.606162in}}%
\pgfusepath{clip}%
\pgfsetbuttcap%
\pgfsetmiterjoin%
\definecolor{currentfill}{rgb}{0.121569,0.466667,0.705882}%
\pgfsetfillcolor{currentfill}%
\pgfsetlinewidth{0.000000pt}%
\definecolor{currentstroke}{rgb}{0.000000,0.000000,0.000000}%
\pgfsetstrokecolor{currentstroke}%
\pgfsetstrokeopacity{0.000000}%
\pgfsetdash{}{0pt}%
\pgfpathmoveto{\pgfqpoint{1.810278in}{-18.082857in}}%
\pgfpathlineto{\pgfqpoint{1.852777in}{-18.082857in}}%
\pgfpathlineto{\pgfqpoint{1.852777in}{0.715094in}}%
\pgfpathlineto{\pgfqpoint{1.810278in}{0.715094in}}%
\pgfpathlineto{\pgfqpoint{1.810278in}{-18.082857in}}%
\pgfpathclose%
\pgfusepath{fill}%
\end{pgfscope}%
\begin{pgfscope}%
\pgfpathrectangle{\pgfqpoint{1.738030in}{0.563553in}}{\pgfqpoint{0.356988in}{0.606162in}}%
\pgfusepath{clip}%
\pgfsetbuttcap%
\pgfsetmiterjoin%
\definecolor{currentfill}{rgb}{1.000000,0.498039,0.054902}%
\pgfsetfillcolor{currentfill}%
\pgfsetlinewidth{0.000000pt}%
\definecolor{currentstroke}{rgb}{0.000000,0.000000,0.000000}%
\pgfsetstrokecolor{currentstroke}%
\pgfsetstrokeopacity{0.000000}%
\pgfsetdash{}{0pt}%
\pgfpathmoveto{\pgfqpoint{1.895275in}{-18.082857in}}%
\pgfpathlineto{\pgfqpoint{1.937774in}{-18.082857in}}%
\pgfpathlineto{\pgfqpoint{1.937774in}{0.809989in}}%
\pgfpathlineto{\pgfqpoint{1.895275in}{0.809989in}}%
\pgfpathlineto{\pgfqpoint{1.895275in}{-18.082857in}}%
\pgfpathclose%
\pgfusepath{fill}%
\end{pgfscope}%
\begin{pgfscope}%
\pgfpathrectangle{\pgfqpoint{1.738030in}{0.563553in}}{\pgfqpoint{0.356988in}{0.606162in}}%
\pgfusepath{clip}%
\pgfsetbuttcap%
\pgfsetmiterjoin%
\definecolor{currentfill}{rgb}{0.172549,0.627451,0.172549}%
\pgfsetfillcolor{currentfill}%
\pgfsetlinewidth{0.000000pt}%
\definecolor{currentstroke}{rgb}{0.000000,0.000000,0.000000}%
\pgfsetstrokecolor{currentstroke}%
\pgfsetstrokeopacity{0.000000}%
\pgfsetdash{}{0pt}%
\pgfpathmoveto{\pgfqpoint{1.980272in}{-18.082857in}}%
\pgfpathlineto{\pgfqpoint{2.022771in}{-18.082857in}}%
\pgfpathlineto{\pgfqpoint{2.022771in}{1.018175in}}%
\pgfpathlineto{\pgfqpoint{1.980272in}{1.018175in}}%
\pgfpathlineto{\pgfqpoint{1.980272in}{-18.082857in}}%
\pgfpathclose%
\pgfusepath{fill}%
\end{pgfscope}%
\begin{pgfscope}%
\pgfsetrectcap%
\pgfsetmiterjoin%
\pgfsetlinewidth{1.003750pt}%
\definecolor{currentstroke}{rgb}{1.000000,1.000000,1.000000}%
\pgfsetstrokecolor{currentstroke}%
\pgfsetdash{}{0pt}%
\pgfpathmoveto{\pgfqpoint{1.738030in}{0.563553in}}%
\pgfpathlineto{\pgfqpoint{1.738030in}{1.169715in}}%
\pgfusepath{stroke}%
\end{pgfscope}%
\begin{pgfscope}%
\pgfsetrectcap%
\pgfsetmiterjoin%
\pgfsetlinewidth{1.003750pt}%
\definecolor{currentstroke}{rgb}{1.000000,1.000000,1.000000}%
\pgfsetstrokecolor{currentstroke}%
\pgfsetdash{}{0pt}%
\pgfpathmoveto{\pgfqpoint{2.095018in}{0.563553in}}%
\pgfpathlineto{\pgfqpoint{2.095018in}{1.169715in}}%
\pgfusepath{stroke}%
\end{pgfscope}%
\begin{pgfscope}%
\pgfsetrectcap%
\pgfsetmiterjoin%
\pgfsetlinewidth{1.003750pt}%
\definecolor{currentstroke}{rgb}{1.000000,1.000000,1.000000}%
\pgfsetstrokecolor{currentstroke}%
\pgfsetdash{}{0pt}%
\pgfpathmoveto{\pgfqpoint{1.738030in}{0.563553in}}%
\pgfpathlineto{\pgfqpoint{2.095018in}{0.563553in}}%
\pgfusepath{stroke}%
\end{pgfscope}%
\begin{pgfscope}%
\pgfsetrectcap%
\pgfsetmiterjoin%
\pgfsetlinewidth{1.003750pt}%
\definecolor{currentstroke}{rgb}{1.000000,1.000000,1.000000}%
\pgfsetstrokecolor{currentstroke}%
\pgfsetdash{}{0pt}%
\pgfpathmoveto{\pgfqpoint{1.738030in}{1.169715in}}%
\pgfpathlineto{\pgfqpoint{2.095018in}{1.169715in}}%
\pgfusepath{stroke}%
\end{pgfscope}%
\begin{pgfscope}%
\pgfsetbuttcap%
\pgfsetmiterjoin%
\definecolor{currentfill}{rgb}{0.898039,0.898039,0.898039}%
\pgfsetfillcolor{currentfill}%
\pgfsetlinewidth{0.000000pt}%
\definecolor{currentstroke}{rgb}{0.000000,0.000000,0.000000}%
\pgfsetstrokecolor{currentstroke}%
\pgfsetstrokeopacity{0.000000}%
\pgfsetdash{}{0pt}%
\pgfpathmoveto{\pgfqpoint{2.453072in}{0.563553in}}%
\pgfpathlineto{\pgfqpoint{2.810059in}{0.563553in}}%
\pgfpathlineto{\pgfqpoint{2.810059in}{1.169715in}}%
\pgfpathlineto{\pgfqpoint{2.453072in}{1.169715in}}%
\pgfpathlineto{\pgfqpoint{2.453072in}{0.563553in}}%
\pgfpathclose%
\pgfusepath{fill}%
\end{pgfscope}%
\begin{pgfscope}%
\definecolor{textcolor}{rgb}{0.333333,0.333333,0.333333}%
\pgfsetstrokecolor{textcolor}%
\pgfsetfillcolor{textcolor}%
\pgftext[x=2.631566in,y=0.507998in,,top]{\color{textcolor}{\rmfamily\fontsize{9.000000}{10.800000}\selectfont\catcode`\^=\active\def^{\ifmmode\sp\else\^{}\fi}\catcode`\%=\active\def
\end{pgfscope}%
\begin{pgfscope}%
\pgfpathrectangle{\pgfqpoint{2.453072in}{0.563553in}}{\pgfqpoint{0.356988in}{0.606162in}}%
\pgfusepath{clip}%
\pgfsetrectcap%
\pgfsetroundjoin%
\pgfsetlinewidth{0.803000pt}%
\definecolor{currentstroke}{rgb}{1.000000,1.000000,1.000000}%
\pgfsetstrokecolor{currentstroke}%
\pgfsetdash{}{0pt}%
\pgfpathmoveto{\pgfqpoint{2.453072in}{0.684786in}}%
\pgfpathlineto{\pgfqpoint{2.810059in}{0.684786in}}%
\pgfusepath{stroke}%
\end{pgfscope}%
\begin{pgfscope}%
\pgfsetbuttcap%
\pgfsetroundjoin%
\definecolor{currentfill}{rgb}{0.333333,0.333333,0.333333}%
\pgfsetfillcolor{currentfill}%
\pgfsetlinewidth{0.803000pt}%
\definecolor{currentstroke}{rgb}{0.333333,0.333333,0.333333}%
\pgfsetstrokecolor{currentstroke}%
\pgfsetdash{}{0pt}%
\pgfsys@defobject{currentmarker}{\pgfqpoint{-0.048611in}{0.000000in}}{\pgfqpoint{-0.000000in}{0.000000in}}{%
\pgfpathmoveto{\pgfqpoint{-0.000000in}{0.000000in}}%
\pgfpathlineto{\pgfqpoint{-0.048611in}{0.000000in}}%
\pgfusepath{stroke,fill}%
}%
\begin{pgfscope}%
\pgfsys@transformshift{2.453072in}{0.684786in}%
\pgfsys@useobject{currentmarker}{}%
\end{pgfscope}%
\end{pgfscope}%
\begin{pgfscope}%
\definecolor{textcolor}{rgb}{0.333333,0.333333,0.333333}%
\pgfsetstrokecolor{textcolor}%
\pgfsetfillcolor{textcolor}%
\pgftext[x=2.300486in, y=0.651028in, left, base]{\color{textcolor}{\rmfamily\fontsize{7.000000}{8.400000}\selectfont\catcode`\^=\active\def^{\ifmmode\sp\else\^{}\fi}\catcode`\%=\active\def
\end{pgfscope}%
\begin{pgfscope}%
\pgfpathrectangle{\pgfqpoint{2.453072in}{0.563553in}}{\pgfqpoint{0.356988in}{0.606162in}}%
\pgfusepath{clip}%
\pgfsetrectcap%
\pgfsetroundjoin%
\pgfsetlinewidth{0.803000pt}%
\definecolor{currentstroke}{rgb}{1.000000,1.000000,1.000000}%
\pgfsetstrokecolor{currentstroke}%
\pgfsetdash{}{0pt}%
\pgfpathmoveto{\pgfqpoint{2.453072in}{0.927250in}}%
\pgfpathlineto{\pgfqpoint{2.810059in}{0.927250in}}%
\pgfusepath{stroke}%
\end{pgfscope}%
\begin{pgfscope}%
\pgfsetbuttcap%
\pgfsetroundjoin%
\definecolor{currentfill}{rgb}{0.333333,0.333333,0.333333}%
\pgfsetfillcolor{currentfill}%
\pgfsetlinewidth{0.803000pt}%
\definecolor{currentstroke}{rgb}{0.333333,0.333333,0.333333}%
\pgfsetstrokecolor{currentstroke}%
\pgfsetdash{}{0pt}%
\pgfsys@defobject{currentmarker}{\pgfqpoint{-0.048611in}{0.000000in}}{\pgfqpoint{-0.000000in}{0.000000in}}{%
\pgfpathmoveto{\pgfqpoint{-0.000000in}{0.000000in}}%
\pgfpathlineto{\pgfqpoint{-0.048611in}{0.000000in}}%
\pgfusepath{stroke,fill}%
}%
\begin{pgfscope}%
\pgfsys@transformshift{2.453072in}{0.927250in}%
\pgfsys@useobject{currentmarker}{}%
\end{pgfscope}%
\end{pgfscope}%
\begin{pgfscope}%
\definecolor{textcolor}{rgb}{0.333333,0.333333,0.333333}%
\pgfsetstrokecolor{textcolor}%
\pgfsetfillcolor{textcolor}%
\pgftext[x=2.300486in, y=0.893493in, left, base]{\color{textcolor}{\rmfamily\fontsize{7.000000}{8.400000}\selectfont\catcode`\^=\active\def^{\ifmmode\sp\else\^{}\fi}\catcode`\%=\active\def
\end{pgfscope}%
\begin{pgfscope}%
\pgfpathrectangle{\pgfqpoint{2.453072in}{0.563553in}}{\pgfqpoint{0.356988in}{0.606162in}}%
\pgfusepath{clip}%
\pgfsetrectcap%
\pgfsetroundjoin%
\pgfsetlinewidth{0.803000pt}%
\definecolor{currentstroke}{rgb}{1.000000,1.000000,1.000000}%
\pgfsetstrokecolor{currentstroke}%
\pgfsetdash{}{0pt}%
\pgfpathmoveto{\pgfqpoint{2.453072in}{1.169715in}}%
\pgfpathlineto{\pgfqpoint{2.810059in}{1.169715in}}%
\pgfusepath{stroke}%
\end{pgfscope}%
\begin{pgfscope}%
\pgfsetbuttcap%
\pgfsetroundjoin%
\definecolor{currentfill}{rgb}{0.333333,0.333333,0.333333}%
\pgfsetfillcolor{currentfill}%
\pgfsetlinewidth{0.803000pt}%
\definecolor{currentstroke}{rgb}{0.333333,0.333333,0.333333}%
\pgfsetstrokecolor{currentstroke}%
\pgfsetdash{}{0pt}%
\pgfsys@defobject{currentmarker}{\pgfqpoint{-0.048611in}{0.000000in}}{\pgfqpoint{-0.000000in}{0.000000in}}{%
\pgfpathmoveto{\pgfqpoint{-0.000000in}{0.000000in}}%
\pgfpathlineto{\pgfqpoint{-0.048611in}{0.000000in}}%
\pgfusepath{stroke,fill}%
}%
\begin{pgfscope}%
\pgfsys@transformshift{2.453072in}{1.169715in}%
\pgfsys@useobject{currentmarker}{}%
\end{pgfscope}%
\end{pgfscope}%
\begin{pgfscope}%
\definecolor{textcolor}{rgb}{0.333333,0.333333,0.333333}%
\pgfsetstrokecolor{textcolor}%
\pgfsetfillcolor{textcolor}%
\pgftext[x=2.245124in, y=1.135957in, left, base]{\color{textcolor}{\rmfamily\fontsize{7.000000}{8.400000}\selectfont\catcode`\^=\active\def^{\ifmmode\sp\else\^{}\fi}\catcode`\%=\active\def
\end{pgfscope}%
\begin{pgfscope}%
\pgfpathrectangle{\pgfqpoint{2.453072in}{0.563553in}}{\pgfqpoint{0.356988in}{0.606162in}}%
\pgfusepath{clip}%
\pgfsetbuttcap%
\pgfsetmiterjoin%
\definecolor{currentfill}{rgb}{0.121569,0.466667,0.705882}%
\pgfsetfillcolor{currentfill}%
\pgfsetlinewidth{0.000000pt}%
\definecolor{currentstroke}{rgb}{0.000000,0.000000,0.000000}%
\pgfsetstrokecolor{currentstroke}%
\pgfsetstrokeopacity{0.000000}%
\pgfsetdash{}{0pt}%
\pgfpathmoveto{\pgfqpoint{2.525319in}{-0.042609in}}%
\pgfpathlineto{\pgfqpoint{2.567818in}{-0.042609in}}%
\pgfpathlineto{\pgfqpoint{2.567818in}{1.036359in}}%
\pgfpathlineto{\pgfqpoint{2.525319in}{1.036359in}}%
\pgfpathlineto{\pgfqpoint{2.525319in}{-0.042609in}}%
\pgfpathclose%
\pgfusepath{fill}%
\end{pgfscope}%
\begin{pgfscope}%
\pgfpathrectangle{\pgfqpoint{2.453072in}{0.563553in}}{\pgfqpoint{0.356988in}{0.606162in}}%
\pgfusepath{clip}%
\pgfsetbuttcap%
\pgfsetmiterjoin%
\definecolor{currentfill}{rgb}{1.000000,0.498039,0.054902}%
\pgfsetfillcolor{currentfill}%
\pgfsetlinewidth{0.000000pt}%
\definecolor{currentstroke}{rgb}{0.000000,0.000000,0.000000}%
\pgfsetstrokecolor{currentstroke}%
\pgfsetstrokeopacity{0.000000}%
\pgfsetdash{}{0pt}%
\pgfpathmoveto{\pgfqpoint{2.610316in}{-0.042609in}}%
\pgfpathlineto{\pgfqpoint{2.652815in}{-0.042609in}}%
\pgfpathlineto{\pgfqpoint{2.652815in}{0.939373in}}%
\pgfpathlineto{\pgfqpoint{2.610316in}{0.939373in}}%
\pgfpathlineto{\pgfqpoint{2.610316in}{-0.042609in}}%
\pgfpathclose%
\pgfusepath{fill}%
\end{pgfscope}%
\begin{pgfscope}%
\pgfpathrectangle{\pgfqpoint{2.453072in}{0.563553in}}{\pgfqpoint{0.356988in}{0.606162in}}%
\pgfusepath{clip}%
\pgfsetbuttcap%
\pgfsetmiterjoin%
\definecolor{currentfill}{rgb}{0.172549,0.627451,0.172549}%
\pgfsetfillcolor{currentfill}%
\pgfsetlinewidth{0.000000pt}%
\definecolor{currentstroke}{rgb}{0.000000,0.000000,0.000000}%
\pgfsetstrokecolor{currentstroke}%
\pgfsetstrokeopacity{0.000000}%
\pgfsetdash{}{0pt}%
\pgfpathmoveto{\pgfqpoint{2.695313in}{-0.042609in}}%
\pgfpathlineto{\pgfqpoint{2.737812in}{-0.042609in}}%
\pgfpathlineto{\pgfqpoint{2.737812in}{0.939373in}}%
\pgfpathlineto{\pgfqpoint{2.695313in}{0.939373in}}%
\pgfpathlineto{\pgfqpoint{2.695313in}{-0.042609in}}%
\pgfpathclose%
\pgfusepath{fill}%
\end{pgfscope}%
\begin{pgfscope}%
\pgfsetrectcap%
\pgfsetmiterjoin%
\pgfsetlinewidth{1.003750pt}%
\definecolor{currentstroke}{rgb}{1.000000,1.000000,1.000000}%
\pgfsetstrokecolor{currentstroke}%
\pgfsetdash{}{0pt}%
\pgfpathmoveto{\pgfqpoint{2.453072in}{0.563553in}}%
\pgfpathlineto{\pgfqpoint{2.453072in}{1.169715in}}%
\pgfusepath{stroke}%
\end{pgfscope}%
\begin{pgfscope}%
\pgfsetrectcap%
\pgfsetmiterjoin%
\pgfsetlinewidth{1.003750pt}%
\definecolor{currentstroke}{rgb}{1.000000,1.000000,1.000000}%
\pgfsetstrokecolor{currentstroke}%
\pgfsetdash{}{0pt}%
\pgfpathmoveto{\pgfqpoint{2.810059in}{0.563553in}}%
\pgfpathlineto{\pgfqpoint{2.810059in}{1.169715in}}%
\pgfusepath{stroke}%
\end{pgfscope}%
\begin{pgfscope}%
\pgfsetrectcap%
\pgfsetmiterjoin%
\pgfsetlinewidth{1.003750pt}%
\definecolor{currentstroke}{rgb}{1.000000,1.000000,1.000000}%
\pgfsetstrokecolor{currentstroke}%
\pgfsetdash{}{0pt}%
\pgfpathmoveto{\pgfqpoint{2.453072in}{0.563553in}}%
\pgfpathlineto{\pgfqpoint{2.810059in}{0.563553in}}%
\pgfusepath{stroke}%
\end{pgfscope}%
\begin{pgfscope}%
\pgfsetrectcap%
\pgfsetmiterjoin%
\pgfsetlinewidth{1.003750pt}%
\definecolor{currentstroke}{rgb}{1.000000,1.000000,1.000000}%
\pgfsetstrokecolor{currentstroke}%
\pgfsetdash{}{0pt}%
\pgfpathmoveto{\pgfqpoint{2.453072in}{1.169715in}}%
\pgfpathlineto{\pgfqpoint{2.810059in}{1.169715in}}%
\pgfusepath{stroke}%
\end{pgfscope}%
\begin{pgfscope}%
\pgfsetbuttcap%
\pgfsetmiterjoin%
\definecolor{currentfill}{rgb}{0.898039,0.898039,0.898039}%
\pgfsetfillcolor{currentfill}%
\pgfsetfillopacity{0.800000}%
\pgfsetlinewidth{0.501875pt}%
\definecolor{currentstroke}{rgb}{0.800000,0.800000,0.800000}%
\pgfsetstrokecolor{currentstroke}%
\pgfsetstrokeopacity{0.800000}%
\pgfsetdash{}{0pt}%
\pgfpathmoveto{\pgfqpoint{0.169375in}{0.100000in}}%
\pgfpathlineto{\pgfqpoint{2.741098in}{0.100000in}}%
\pgfpathquadraticcurveto{\pgfqpoint{2.766098in}{0.100000in}}{\pgfqpoint{2.766098in}{0.125000in}}%
\pgfpathlineto{\pgfqpoint{2.766098in}{0.286806in}}%
\pgfpathquadraticcurveto{\pgfqpoint{2.766098in}{0.311806in}}{\pgfqpoint{2.741098in}{0.311806in}}%
\pgfpathlineto{\pgfqpoint{0.169375in}{0.311806in}}%
\pgfpathquadraticcurveto{\pgfqpoint{0.144375in}{0.311806in}}{\pgfqpoint{0.144375in}{0.286806in}}%
\pgfpathlineto{\pgfqpoint{0.144375in}{0.125000in}}%
\pgfpathquadraticcurveto{\pgfqpoint{0.144375in}{0.100000in}}{\pgfqpoint{0.169375in}{0.100000in}}%
\pgfpathlineto{\pgfqpoint{0.169375in}{0.100000in}}%
\pgfpathclose%
\pgfusepath{stroke,fill}%
\end{pgfscope}%
\begin{pgfscope}%
\pgfsetbuttcap%
\pgfsetmiterjoin%
\definecolor{currentfill}{rgb}{0.121569,0.466667,0.705882}%
\pgfsetfillcolor{currentfill}%
\pgfsetlinewidth{1.003750pt}%
\definecolor{currentstroke}{rgb}{0.121569,0.466667,0.705882}%
\pgfsetstrokecolor{currentstroke}%
\pgfsetdash{}{0pt}%
\pgfsys@defobject{currentmarker}{\pgfqpoint{-0.041667in}{-0.041667in}}{\pgfqpoint{0.041667in}{0.041667in}}{%
\pgfpathmoveto{\pgfqpoint{-0.041667in}{-0.041667in}}%
\pgfpathlineto{\pgfqpoint{0.041667in}{-0.041667in}}%
\pgfpathlineto{\pgfqpoint{0.041667in}{0.041667in}}%
\pgfpathlineto{\pgfqpoint{-0.041667in}{0.041667in}}%
\pgfpathlineto{\pgfqpoint{-0.041667in}{-0.041667in}}%
\pgfpathclose%
\pgfusepath{stroke,fill}%
}%
\begin{pgfscope}%
\pgfsys@transformshift{0.319375in}{0.218056in}%
\pgfsys@useobject{currentmarker}{}%
\end{pgfscope}%
\end{pgfscope}%
\begin{pgfscope}%
\definecolor{textcolor}{rgb}{0.000000,0.000000,0.000000}%
\pgfsetstrokecolor{textcolor}%
\pgfsetfillcolor{textcolor}%
\pgftext[x=0.544375in,y=0.174306in,left,base]{\color{textcolor}{\rmfamily\fontsize{9.000000}{10.800000}\selectfont\catcode`\^=\active\def^{\ifmmode\sp\else\^{}\fi}\catcode`\%=\active\def
\end{pgfscope}%
\begin{pgfscope}%
\pgfsetbuttcap%
\pgfsetmiterjoin%
\definecolor{currentfill}{rgb}{1.000000,0.498039,0.054902}%
\pgfsetfillcolor{currentfill}%
\pgfsetlinewidth{1.003750pt}%
\definecolor{currentstroke}{rgb}{1.000000,0.498039,0.054902}%
\pgfsetstrokecolor{currentstroke}%
\pgfsetdash{}{0pt}%
\pgfsys@defobject{currentmarker}{\pgfqpoint{-0.041667in}{-0.041667in}}{\pgfqpoint{0.041667in}{0.041667in}}{%
\pgfpathmoveto{\pgfqpoint{-0.041667in}{-0.041667in}}%
\pgfpathlineto{\pgfqpoint{0.041667in}{-0.041667in}}%
\pgfpathlineto{\pgfqpoint{0.041667in}{0.041667in}}%
\pgfpathlineto{\pgfqpoint{-0.041667in}{0.041667in}}%
\pgfpathlineto{\pgfqpoint{-0.041667in}{-0.041667in}}%
\pgfpathclose%
\pgfusepath{stroke,fill}%
}%
\begin{pgfscope}%
\pgfsys@transformshift{1.205870in}{0.218056in}%
\pgfsys@useobject{currentmarker}{}%
\end{pgfscope}%
\end{pgfscope}%
\begin{pgfscope}%
\definecolor{textcolor}{rgb}{0.000000,0.000000,0.000000}%
\pgfsetstrokecolor{textcolor}%
\pgfsetfillcolor{textcolor}%
\pgftext[x=1.430870in,y=0.174306in,left,base]{\color{textcolor}{\rmfamily\fontsize{9.000000}{10.800000}\selectfont\catcode`\^=\active\def^{\ifmmode\sp\else\^{}\fi}\catcode`\%=\active\def
\end{pgfscope}%
\begin{pgfscope}%
\pgfsetbuttcap%
\pgfsetmiterjoin%
\definecolor{currentfill}{rgb}{0.172549,0.627451,0.172549}%
\pgfsetfillcolor{currentfill}%
\pgfsetlinewidth{1.003750pt}%
\definecolor{currentstroke}{rgb}{0.172549,0.627451,0.172549}%
\pgfsetstrokecolor{currentstroke}%
\pgfsetdash{}{0pt}%
\pgfsys@defobject{currentmarker}{\pgfqpoint{-0.041667in}{-0.041667in}}{\pgfqpoint{0.041667in}{0.041667in}}{%
\pgfpathmoveto{\pgfqpoint{-0.041667in}{-0.041667in}}%
\pgfpathlineto{\pgfqpoint{0.041667in}{-0.041667in}}%
\pgfpathlineto{\pgfqpoint{0.041667in}{0.041667in}}%
\pgfpathlineto{\pgfqpoint{-0.041667in}{0.041667in}}%
\pgfpathlineto{\pgfqpoint{-0.041667in}{-0.041667in}}%
\pgfpathclose%
\pgfusepath{stroke,fill}%
}%
\begin{pgfscope}%
\pgfsys@transformshift{2.255568in}{0.218056in}%
\pgfsys@useobject{currentmarker}{}%
\end{pgfscope}%
\end{pgfscope}%
\begin{pgfscope}%
\definecolor{textcolor}{rgb}{0.000000,0.000000,0.000000}%
\pgfsetstrokecolor{textcolor}%
\pgfsetfillcolor{textcolor}%
\pgftext[x=2.480568in,y=0.174306in,left,base]{\color{textcolor}{\rmfamily\fontsize{9.000000}{10.800000}\selectfont\catcode`\^=\active\def^{\ifmmode\sp\else\^{}\fi}\catcode`\%=\active\def
\end{pgfscope}%
\end{pgfpicture}%
\makeatother%
\endgroup%

%% file: img/retrievers.pgf
\begingroup%
\makeatletter%
\begin{pgfpicture}%
\pgfpathrectangle{\pgfpointorigin}{\pgfqpoint{3.313369in}{1.563678in}}%
\pgfusepath{use as bounding box, clip}%
\begin{pgfscope}%
\pgfsetbuttcap%
\pgfsetmiterjoin%
\definecolor{currentfill}{rgb}{1.000000,1.000000,1.000000}%
\pgfsetfillcolor{currentfill}%
\pgfsetlinewidth{0.000000pt}%
\definecolor{currentstroke}{rgb}{0.500000,0.500000,0.500000}%
\pgfsetstrokecolor{currentstroke}%
\pgfsetdash{}{0pt}%
\pgfpathmoveto{\pgfqpoint{0.000000in}{0.000000in}}%
\pgfpathlineto{\pgfqpoint{3.313369in}{0.000000in}}%
\pgfpathlineto{\pgfqpoint{3.313369in}{1.563678in}}%
\pgfpathlineto{\pgfqpoint{0.000000in}{1.563678in}}%
\pgfpathlineto{\pgfqpoint{0.000000in}{0.000000in}}%
\pgfpathclose%
\pgfusepath{fill}%
\end{pgfscope}%
\begin{pgfscope}%
\pgfsetbuttcap%
\pgfsetmiterjoin%
\definecolor{currentfill}{rgb}{0.898039,0.898039,0.898039}%
\pgfsetfillcolor{currentfill}%
\pgfsetlinewidth{0.000000pt}%
\definecolor{currentstroke}{rgb}{0.000000,0.000000,0.000000}%
\pgfsetstrokecolor{currentstroke}%
\pgfsetstrokeopacity{0.000000}%
\pgfsetdash{}{0pt}%
\pgfpathmoveto{\pgfqpoint{0.396297in}{0.738693in}}%
\pgfpathlineto{\pgfqpoint{0.898039in}{0.738693in}}%
\pgfpathlineto{\pgfqpoint{0.898039in}{1.463678in}}%
\pgfpathlineto{\pgfqpoint{0.396297in}{1.463678in}}%
\pgfpathlineto{\pgfqpoint{0.396297in}{0.738693in}}%
\pgfpathclose%
\pgfusepath{fill}%
\end{pgfscope}%
\begin{pgfscope}%
\definecolor{textcolor}{rgb}{0.333333,0.333333,0.333333}%
\pgfsetstrokecolor{textcolor}%
\pgfsetfillcolor{textcolor}%
\pgftext[x=0.647168in,y=0.683137in,,top]{\color{textcolor}{\rmfamily\fontsize{9.000000}{10.800000}\selectfont\catcode`\^=\active\def^{\ifmmode\sp\else\^{}\fi}\catcode`\%=\active\def
\end{pgfscope}%
\begin{pgfscope}%
\pgfpathrectangle{\pgfqpoint{0.396297in}{0.738693in}}{\pgfqpoint{0.501742in}{0.724986in}}%
\pgfusepath{clip}%
\pgfsetrectcap%
\pgfsetroundjoin%
\pgfsetlinewidth{0.803000pt}%
\definecolor{currentstroke}{rgb}{1.000000,1.000000,1.000000}%
\pgfsetstrokecolor{currentstroke}%
\pgfsetdash{}{0pt}%
\pgfpathmoveto{\pgfqpoint{0.396297in}{0.859523in}}%
\pgfpathlineto{\pgfqpoint{0.898039in}{0.859523in}}%
\pgfusepath{stroke}%
\end{pgfscope}%
\begin{pgfscope}%
\pgfsetbuttcap%
\pgfsetroundjoin%
\definecolor{currentfill}{rgb}{0.333333,0.333333,0.333333}%
\pgfsetfillcolor{currentfill}%
\pgfsetlinewidth{0.803000pt}%
\definecolor{currentstroke}{rgb}{0.333333,0.333333,0.333333}%
\pgfsetstrokecolor{currentstroke}%
\pgfsetdash{}{0pt}%
\pgfsys@defobject{currentmarker}{\pgfqpoint{-0.048611in}{0.000000in}}{\pgfqpoint{-0.000000in}{0.000000in}}{%
\pgfpathmoveto{\pgfqpoint{-0.000000in}{0.000000in}}%
\pgfpathlineto{\pgfqpoint{-0.048611in}{0.000000in}}%
\pgfusepath{stroke,fill}%
}%
\begin{pgfscope}%
\pgfsys@transformshift{0.396297in}{0.859523in}%
\pgfsys@useobject{currentmarker}{}%
\end{pgfscope}%
\end{pgfscope}%
\begin{pgfscope}%
\definecolor{textcolor}{rgb}{0.333333,0.333333,0.333333}%
\pgfsetstrokecolor{textcolor}%
\pgfsetfillcolor{textcolor}%
\pgftext[x=0.100000in, y=0.825766in, left, base]{\color{textcolor}{\rmfamily\fontsize{7.000000}{8.400000}\selectfont\catcode`\^=\active\def^{\ifmmode\sp\else\^{}\fi}\catcode`\%=\active\def
\end{pgfscope}%
\begin{pgfscope}%
\pgfpathrectangle{\pgfqpoint{0.396297in}{0.738693in}}{\pgfqpoint{0.501742in}{0.724986in}}%
\pgfusepath{clip}%
\pgfsetrectcap%
\pgfsetroundjoin%
\pgfsetlinewidth{0.803000pt}%
\definecolor{currentstroke}{rgb}{1.000000,1.000000,1.000000}%
\pgfsetstrokecolor{currentstroke}%
\pgfsetdash{}{0pt}%
\pgfpathmoveto{\pgfqpoint{0.396297in}{1.111255in}}%
\pgfpathlineto{\pgfqpoint{0.898039in}{1.111255in}}%
\pgfusepath{stroke}%
\end{pgfscope}%
\begin{pgfscope}%
\pgfsetbuttcap%
\pgfsetroundjoin%
\definecolor{currentfill}{rgb}{0.333333,0.333333,0.333333}%
\pgfsetfillcolor{currentfill}%
\pgfsetlinewidth{0.803000pt}%
\definecolor{currentstroke}{rgb}{0.333333,0.333333,0.333333}%
\pgfsetstrokecolor{currentstroke}%
\pgfsetdash{}{0pt}%
\pgfsys@defobject{currentmarker}{\pgfqpoint{-0.048611in}{0.000000in}}{\pgfqpoint{-0.000000in}{0.000000in}}{%
\pgfpathmoveto{\pgfqpoint{-0.000000in}{0.000000in}}%
\pgfpathlineto{\pgfqpoint{-0.048611in}{0.000000in}}%
\pgfusepath{stroke,fill}%
}%
\begin{pgfscope}%
\pgfsys@transformshift{0.396297in}{1.111255in}%
\pgfsys@useobject{currentmarker}{}%
\end{pgfscope}%
\end{pgfscope}%
\begin{pgfscope}%
\definecolor{textcolor}{rgb}{0.333333,0.333333,0.333333}%
\pgfsetstrokecolor{textcolor}%
\pgfsetfillcolor{textcolor}%
\pgftext[x=0.100000in, y=1.077497in, left, base]{\color{textcolor}{\rmfamily\fontsize{7.000000}{8.400000}\selectfont\catcode`\^=\active\def^{\ifmmode\sp\else\^{}\fi}\catcode`\%=\active\def
\end{pgfscope}%
\begin{pgfscope}%
\pgfpathrectangle{\pgfqpoint{0.396297in}{0.738693in}}{\pgfqpoint{0.501742in}{0.724986in}}%
\pgfusepath{clip}%
\pgfsetrectcap%
\pgfsetroundjoin%
\pgfsetlinewidth{0.803000pt}%
\definecolor{currentstroke}{rgb}{1.000000,1.000000,1.000000}%
\pgfsetstrokecolor{currentstroke}%
\pgfsetdash{}{0pt}%
\pgfpathmoveto{\pgfqpoint{0.396297in}{1.362986in}}%
\pgfpathlineto{\pgfqpoint{0.898039in}{1.362986in}}%
\pgfusepath{stroke}%
\end{pgfscope}%
\begin{pgfscope}%
\pgfsetbuttcap%
\pgfsetroundjoin%
\definecolor{currentfill}{rgb}{0.333333,0.333333,0.333333}%
\pgfsetfillcolor{currentfill}%
\pgfsetlinewidth{0.803000pt}%
\definecolor{currentstroke}{rgb}{0.333333,0.333333,0.333333}%
\pgfsetstrokecolor{currentstroke}%
\pgfsetdash{}{0pt}%
\pgfsys@defobject{currentmarker}{\pgfqpoint{-0.048611in}{0.000000in}}{\pgfqpoint{-0.000000in}{0.000000in}}{%
\pgfpathmoveto{\pgfqpoint{-0.000000in}{0.000000in}}%
\pgfpathlineto{\pgfqpoint{-0.048611in}{0.000000in}}%
\pgfusepath{stroke,fill}%
}%
\begin{pgfscope}%
\pgfsys@transformshift{0.396297in}{1.362986in}%
\pgfsys@useobject{currentmarker}{}%
\end{pgfscope}%
\end{pgfscope}%
\begin{pgfscope}%
\definecolor{textcolor}{rgb}{0.333333,0.333333,0.333333}%
\pgfsetstrokecolor{textcolor}%
\pgfsetfillcolor{textcolor}%
\pgftext[x=0.100000in, y=1.329228in, left, base]{\color{textcolor}{\rmfamily\fontsize{7.000000}{8.400000}\selectfont\catcode`\^=\active\def^{\ifmmode\sp\else\^{}\fi}\catcode`\%=\active\def
\end{pgfscope}%
\begin{pgfscope}%
\pgfpathrectangle{\pgfqpoint{0.396297in}{0.738693in}}{\pgfqpoint{0.501742in}{0.724986in}}%
\pgfusepath{clip}%
\pgfsetbuttcap%
\pgfsetmiterjoin%
\definecolor{currentfill}{rgb}{0.121569,0.466667,0.705882}%
\pgfsetfillcolor{currentfill}%
\pgfsetlinewidth{0.000000pt}%
\definecolor{currentstroke}{rgb}{0.000000,0.000000,0.000000}%
\pgfsetstrokecolor{currentstroke}%
\pgfsetstrokeopacity{0.000000}%
\pgfsetdash{}{0pt}%
\pgfpathmoveto{\pgfqpoint{0.478313in}{-3.168175in}}%
\pgfpathlineto{\pgfqpoint{0.526557in}{-3.168175in}}%
\pgfpathlineto{\pgfqpoint{0.526557in}{0.919939in}}%
\pgfpathlineto{\pgfqpoint{0.478313in}{0.919939in}}%
\pgfpathlineto{\pgfqpoint{0.478313in}{-3.168175in}}%
\pgfpathclose%
\pgfusepath{fill}%
\end{pgfscope}%
\begin{pgfscope}%
\pgfpathrectangle{\pgfqpoint{0.396297in}{0.738693in}}{\pgfqpoint{0.501742in}{0.724986in}}%
\pgfusepath{clip}%
\pgfsetbuttcap%
\pgfsetmiterjoin%
\definecolor{currentfill}{rgb}{1.000000,0.498039,0.054902}%
\pgfsetfillcolor{currentfill}%
\pgfsetlinewidth{0.000000pt}%
\definecolor{currentstroke}{rgb}{0.000000,0.000000,0.000000}%
\pgfsetstrokecolor{currentstroke}%
\pgfsetstrokeopacity{0.000000}%
\pgfsetdash{}{0pt}%
\pgfpathmoveto{\pgfqpoint{0.574802in}{-3.168175in}}%
\pgfpathlineto{\pgfqpoint{0.623046in}{-3.168175in}}%
\pgfpathlineto{\pgfqpoint{0.623046in}{0.990424in}}%
\pgfpathlineto{\pgfqpoint{0.574802in}{0.990424in}}%
\pgfpathlineto{\pgfqpoint{0.574802in}{-3.168175in}}%
\pgfpathclose%
\pgfusepath{fill}%
\end{pgfscope}%
\begin{pgfscope}%
\pgfpathrectangle{\pgfqpoint{0.396297in}{0.738693in}}{\pgfqpoint{0.501742in}{0.724986in}}%
\pgfusepath{clip}%
\pgfsetbuttcap%
\pgfsetmiterjoin%
\definecolor{currentfill}{rgb}{0.172549,0.627451,0.172549}%
\pgfsetfillcolor{currentfill}%
\pgfsetlinewidth{0.000000pt}%
\definecolor{currentstroke}{rgb}{0.000000,0.000000,0.000000}%
\pgfsetstrokecolor{currentstroke}%
\pgfsetstrokeopacity{0.000000}%
\pgfsetdash{}{0pt}%
\pgfpathmoveto{\pgfqpoint{0.671290in}{-3.168175in}}%
\pgfpathlineto{\pgfqpoint{0.719535in}{-3.168175in}}%
\pgfpathlineto{\pgfqpoint{0.719535in}{1.272363in}}%
\pgfpathlineto{\pgfqpoint{0.671290in}{1.272363in}}%
\pgfpathlineto{\pgfqpoint{0.671290in}{-3.168175in}}%
\pgfpathclose%
\pgfusepath{fill}%
\end{pgfscope}%
\begin{pgfscope}%
\pgfpathrectangle{\pgfqpoint{0.396297in}{0.738693in}}{\pgfqpoint{0.501742in}{0.724986in}}%
\pgfusepath{clip}%
\pgfsetbuttcap%
\pgfsetmiterjoin%
\definecolor{currentfill}{rgb}{0.839216,0.152941,0.156863}%
\pgfsetfillcolor{currentfill}%
\pgfsetlinewidth{0.000000pt}%
\definecolor{currentstroke}{rgb}{0.000000,0.000000,0.000000}%
\pgfsetstrokecolor{currentstroke}%
\pgfsetstrokeopacity{0.000000}%
\pgfsetdash{}{0pt}%
\pgfpathmoveto{\pgfqpoint{0.767779in}{-3.168175in}}%
\pgfpathlineto{\pgfqpoint{0.816024in}{-3.168175in}}%
\pgfpathlineto{\pgfqpoint{0.816024in}{1.282432in}}%
\pgfpathlineto{\pgfqpoint{0.767779in}{1.282432in}}%
\pgfpathlineto{\pgfqpoint{0.767779in}{-3.168175in}}%
\pgfpathclose%
\pgfusepath{fill}%
\end{pgfscope}%
\begin{pgfscope}%
\pgfsetrectcap%
\pgfsetmiterjoin%
\pgfsetlinewidth{1.003750pt}%
\definecolor{currentstroke}{rgb}{1.000000,1.000000,1.000000}%
\pgfsetstrokecolor{currentstroke}%
\pgfsetdash{}{0pt}%
\pgfpathmoveto{\pgfqpoint{0.396297in}{0.738693in}}%
\pgfpathlineto{\pgfqpoint{0.396297in}{1.463678in}}%
\pgfusepath{stroke}%
\end{pgfscope}%
\begin{pgfscope}%
\pgfsetrectcap%
\pgfsetmiterjoin%
\pgfsetlinewidth{1.003750pt}%
\definecolor{currentstroke}{rgb}{1.000000,1.000000,1.000000}%
\pgfsetstrokecolor{currentstroke}%
\pgfsetdash{}{0pt}%
\pgfpathmoveto{\pgfqpoint{0.898039in}{0.738693in}}%
\pgfpathlineto{\pgfqpoint{0.898039in}{1.463678in}}%
\pgfusepath{stroke}%
\end{pgfscope}%
\begin{pgfscope}%
\pgfsetrectcap%
\pgfsetmiterjoin%
\pgfsetlinewidth{1.003750pt}%
\definecolor{currentstroke}{rgb}{1.000000,1.000000,1.000000}%
\pgfsetstrokecolor{currentstroke}%
\pgfsetdash{}{0pt}%
\pgfpathmoveto{\pgfqpoint{0.396297in}{0.738693in}}%
\pgfpathlineto{\pgfqpoint{0.898039in}{0.738693in}}%
\pgfusepath{stroke}%
\end{pgfscope}%
\begin{pgfscope}%
\pgfsetrectcap%
\pgfsetmiterjoin%
\pgfsetlinewidth{1.003750pt}%
\definecolor{currentstroke}{rgb}{1.000000,1.000000,1.000000}%
\pgfsetstrokecolor{currentstroke}%
\pgfsetdash{}{0pt}%
\pgfpathmoveto{\pgfqpoint{0.396297in}{1.463678in}}%
\pgfpathlineto{\pgfqpoint{0.898039in}{1.463678in}}%
\pgfusepath{stroke}%
\end{pgfscope}%
\begin{pgfscope}%
\pgfsetbuttcap%
\pgfsetmiterjoin%
\definecolor{currentfill}{rgb}{0.898039,0.898039,0.898039}%
\pgfsetfillcolor{currentfill}%
\pgfsetlinewidth{0.000000pt}%
\definecolor{currentstroke}{rgb}{0.000000,0.000000,0.000000}%
\pgfsetstrokecolor{currentstroke}%
\pgfsetstrokeopacity{0.000000}%
\pgfsetdash{}{0pt}%
\pgfpathmoveto{\pgfqpoint{1.168074in}{0.738693in}}%
\pgfpathlineto{\pgfqpoint{1.669816in}{0.738693in}}%
\pgfpathlineto{\pgfqpoint{1.669816in}{1.463678in}}%
\pgfpathlineto{\pgfqpoint{1.168074in}{1.463678in}}%
\pgfpathlineto{\pgfqpoint{1.168074in}{0.738693in}}%
\pgfpathclose%
\pgfusepath{fill}%
\end{pgfscope}%
\begin{pgfscope}%
\definecolor{textcolor}{rgb}{0.333333,0.333333,0.333333}%
\pgfsetstrokecolor{textcolor}%
\pgfsetfillcolor{textcolor}%
\pgftext[x=1.418945in,y=0.683137in,,top]{\color{textcolor}{\rmfamily\fontsize{9.000000}{10.800000}\selectfont\catcode`\^=\active\def^{\ifmmode\sp\else\^{}\fi}\catcode`\%=\active\def
\end{pgfscope}%
\begin{pgfscope}%
\pgfpathrectangle{\pgfqpoint{1.168074in}{0.738693in}}{\pgfqpoint{0.501742in}{0.724986in}}%
\pgfusepath{clip}%
\pgfsetrectcap%
\pgfsetroundjoin%
\pgfsetlinewidth{0.803000pt}%
\definecolor{currentstroke}{rgb}{1.000000,1.000000,1.000000}%
\pgfsetstrokecolor{currentstroke}%
\pgfsetdash{}{0pt}%
\pgfpathmoveto{\pgfqpoint{1.168074in}{0.919939in}}%
\pgfpathlineto{\pgfqpoint{1.669816in}{0.919939in}}%
\pgfusepath{stroke}%
\end{pgfscope}%
\begin{pgfscope}%
\pgfsetbuttcap%
\pgfsetroundjoin%
\definecolor{currentfill}{rgb}{0.333333,0.333333,0.333333}%
\pgfsetfillcolor{currentfill}%
\pgfsetlinewidth{0.803000pt}%
\definecolor{currentstroke}{rgb}{0.333333,0.333333,0.333333}%
\pgfsetstrokecolor{currentstroke}%
\pgfsetdash{}{0pt}%
\pgfsys@defobject{currentmarker}{\pgfqpoint{-0.048611in}{0.000000in}}{\pgfqpoint{-0.000000in}{0.000000in}}{%
\pgfpathmoveto{\pgfqpoint{-0.000000in}{0.000000in}}%
\pgfpathlineto{\pgfqpoint{-0.048611in}{0.000000in}}%
\pgfusepath{stroke,fill}%
}%
\begin{pgfscope}%
\pgfsys@transformshift{1.168074in}{0.919939in}%
\pgfsys@useobject{currentmarker}{}%
\end{pgfscope}%
\end{pgfscope}%
\begin{pgfscope}%
\definecolor{textcolor}{rgb}{0.333333,0.333333,0.333333}%
\pgfsetstrokecolor{textcolor}%
\pgfsetfillcolor{textcolor}%
\pgftext[x=0.960126in, y=0.886181in, left, base]{\color{textcolor}{\rmfamily\fontsize{7.000000}{8.400000}\selectfont\catcode`\^=\active\def^{\ifmmode\sp\else\^{}\fi}\catcode`\%=\active\def
\end{pgfscope}%
\begin{pgfscope}%
\pgfpathrectangle{\pgfqpoint{1.168074in}{0.738693in}}{\pgfqpoint{0.501742in}{0.724986in}}%
\pgfusepath{clip}%
\pgfsetrectcap%
\pgfsetroundjoin%
\pgfsetlinewidth{0.803000pt}%
\definecolor{currentstroke}{rgb}{1.000000,1.000000,1.000000}%
\pgfsetstrokecolor{currentstroke}%
\pgfsetdash{}{0pt}%
\pgfpathmoveto{\pgfqpoint{1.168074in}{1.121324in}}%
\pgfpathlineto{\pgfqpoint{1.669816in}{1.121324in}}%
\pgfusepath{stroke}%
\end{pgfscope}%
\begin{pgfscope}%
\pgfsetbuttcap%
\pgfsetroundjoin%
\definecolor{currentfill}{rgb}{0.333333,0.333333,0.333333}%
\pgfsetfillcolor{currentfill}%
\pgfsetlinewidth{0.803000pt}%
\definecolor{currentstroke}{rgb}{0.333333,0.333333,0.333333}%
\pgfsetstrokecolor{currentstroke}%
\pgfsetdash{}{0pt}%
\pgfsys@defobject{currentmarker}{\pgfqpoint{-0.048611in}{0.000000in}}{\pgfqpoint{-0.000000in}{0.000000in}}{%
\pgfpathmoveto{\pgfqpoint{-0.000000in}{0.000000in}}%
\pgfpathlineto{\pgfqpoint{-0.048611in}{0.000000in}}%
\pgfusepath{stroke,fill}%
}%
\begin{pgfscope}%
\pgfsys@transformshift{1.168074in}{1.121324in}%
\pgfsys@useobject{currentmarker}{}%
\end{pgfscope}%
\end{pgfscope}%
\begin{pgfscope}%
\definecolor{textcolor}{rgb}{0.333333,0.333333,0.333333}%
\pgfsetstrokecolor{textcolor}%
\pgfsetfillcolor{textcolor}%
\pgftext[x=0.960126in, y=1.087566in, left, base]{\color{textcolor}{\rmfamily\fontsize{7.000000}{8.400000}\selectfont\catcode`\^=\active\def^{\ifmmode\sp\else\^{}\fi}\catcode`\%=\active\def
\end{pgfscope}%
\begin{pgfscope}%
\pgfpathrectangle{\pgfqpoint{1.168074in}{0.738693in}}{\pgfqpoint{0.501742in}{0.724986in}}%
\pgfusepath{clip}%
\pgfsetrectcap%
\pgfsetroundjoin%
\pgfsetlinewidth{0.803000pt}%
\definecolor{currentstroke}{rgb}{1.000000,1.000000,1.000000}%
\pgfsetstrokecolor{currentstroke}%
\pgfsetdash{}{0pt}%
\pgfpathmoveto{\pgfqpoint{1.168074in}{1.322709in}}%
\pgfpathlineto{\pgfqpoint{1.669816in}{1.322709in}}%
\pgfusepath{stroke}%
\end{pgfscope}%
\begin{pgfscope}%
\pgfsetbuttcap%
\pgfsetroundjoin%
\definecolor{currentfill}{rgb}{0.333333,0.333333,0.333333}%
\pgfsetfillcolor{currentfill}%
\pgfsetlinewidth{0.803000pt}%
\definecolor{currentstroke}{rgb}{0.333333,0.333333,0.333333}%
\pgfsetstrokecolor{currentstroke}%
\pgfsetdash{}{0pt}%
\pgfsys@defobject{currentmarker}{\pgfqpoint{-0.048611in}{0.000000in}}{\pgfqpoint{-0.000000in}{0.000000in}}{%
\pgfpathmoveto{\pgfqpoint{-0.000000in}{0.000000in}}%
\pgfpathlineto{\pgfqpoint{-0.048611in}{0.000000in}}%
\pgfusepath{stroke,fill}%
}%
\begin{pgfscope}%
\pgfsys@transformshift{1.168074in}{1.322709in}%
\pgfsys@useobject{currentmarker}{}%
\end{pgfscope}%
\end{pgfscope}%
\begin{pgfscope}%
\definecolor{textcolor}{rgb}{0.333333,0.333333,0.333333}%
\pgfsetstrokecolor{textcolor}%
\pgfsetfillcolor{textcolor}%
\pgftext[x=0.960126in, y=1.288951in, left, base]{\color{textcolor}{\rmfamily\fontsize{7.000000}{8.400000}\selectfont\catcode`\^=\active\def^{\ifmmode\sp\else\^{}\fi}\catcode`\%=\active\def
\end{pgfscope}%
\begin{pgfscope}%
\pgfpathrectangle{\pgfqpoint{1.168074in}{0.738693in}}{\pgfqpoint{0.501742in}{0.724986in}}%
\pgfusepath{clip}%
\pgfsetbuttcap%
\pgfsetmiterjoin%
\definecolor{currentfill}{rgb}{0.121569,0.466667,0.705882}%
\pgfsetfillcolor{currentfill}%
\pgfsetlinewidth{0.000000pt}%
\definecolor{currentstroke}{rgb}{0.000000,0.000000,0.000000}%
\pgfsetstrokecolor{currentstroke}%
\pgfsetstrokeopacity{0.000000}%
\pgfsetdash{}{0pt}%
\pgfpathmoveto{\pgfqpoint{1.250089in}{-11.364541in}}%
\pgfpathlineto{\pgfqpoint{1.298334in}{-11.364541in}}%
\pgfpathlineto{\pgfqpoint{1.298334in}{0.919939in}}%
\pgfpathlineto{\pgfqpoint{1.250089in}{0.919939in}}%
\pgfpathlineto{\pgfqpoint{1.250089in}{-11.364541in}}%
\pgfpathclose%
\pgfusepath{fill}%
\end{pgfscope}%
\begin{pgfscope}%
\pgfpathrectangle{\pgfqpoint{1.168074in}{0.738693in}}{\pgfqpoint{0.501742in}{0.724986in}}%
\pgfusepath{clip}%
\pgfsetbuttcap%
\pgfsetmiterjoin%
\definecolor{currentfill}{rgb}{1.000000,0.498039,0.054902}%
\pgfsetfillcolor{currentfill}%
\pgfsetlinewidth{0.000000pt}%
\definecolor{currentstroke}{rgb}{0.000000,0.000000,0.000000}%
\pgfsetstrokecolor{currentstroke}%
\pgfsetstrokeopacity{0.000000}%
\pgfsetdash{}{0pt}%
\pgfpathmoveto{\pgfqpoint{1.346578in}{-11.364541in}}%
\pgfpathlineto{\pgfqpoint{1.394823in}{-11.364541in}}%
\pgfpathlineto{\pgfqpoint{1.394823in}{1.121324in}}%
\pgfpathlineto{\pgfqpoint{1.346578in}{1.121324in}}%
\pgfpathlineto{\pgfqpoint{1.346578in}{-11.364541in}}%
\pgfpathclose%
\pgfusepath{fill}%
\end{pgfscope}%
\begin{pgfscope}%
\pgfpathrectangle{\pgfqpoint{1.168074in}{0.738693in}}{\pgfqpoint{0.501742in}{0.724986in}}%
\pgfusepath{clip}%
\pgfsetbuttcap%
\pgfsetmiterjoin%
\definecolor{currentfill}{rgb}{0.172549,0.627451,0.172549}%
\pgfsetfillcolor{currentfill}%
\pgfsetlinewidth{0.000000pt}%
\definecolor{currentstroke}{rgb}{0.000000,0.000000,0.000000}%
\pgfsetstrokecolor{currentstroke}%
\pgfsetstrokeopacity{0.000000}%
\pgfsetdash{}{0pt}%
\pgfpathmoveto{\pgfqpoint{1.443067in}{-11.364541in}}%
\pgfpathlineto{\pgfqpoint{1.491311in}{-11.364541in}}%
\pgfpathlineto{\pgfqpoint{1.491311in}{1.222016in}}%
\pgfpathlineto{\pgfqpoint{1.443067in}{1.222016in}}%
\pgfpathlineto{\pgfqpoint{1.443067in}{-11.364541in}}%
\pgfpathclose%
\pgfusepath{fill}%
\end{pgfscope}%
\begin{pgfscope}%
\pgfpathrectangle{\pgfqpoint{1.168074in}{0.738693in}}{\pgfqpoint{0.501742in}{0.724986in}}%
\pgfusepath{clip}%
\pgfsetbuttcap%
\pgfsetmiterjoin%
\definecolor{currentfill}{rgb}{0.839216,0.152941,0.156863}%
\pgfsetfillcolor{currentfill}%
\pgfsetlinewidth{0.000000pt}%
\definecolor{currentstroke}{rgb}{0.000000,0.000000,0.000000}%
\pgfsetstrokecolor{currentstroke}%
\pgfsetstrokeopacity{0.000000}%
\pgfsetdash{}{0pt}%
\pgfpathmoveto{\pgfqpoint{1.539556in}{-11.364541in}}%
\pgfpathlineto{\pgfqpoint{1.587800in}{-11.364541in}}%
\pgfpathlineto{\pgfqpoint{1.587800in}{1.282432in}}%
\pgfpathlineto{\pgfqpoint{1.539556in}{1.282432in}}%
\pgfpathlineto{\pgfqpoint{1.539556in}{-11.364541in}}%
\pgfpathclose%
\pgfusepath{fill}%
\end{pgfscope}%
\begin{pgfscope}%
\pgfsetrectcap%
\pgfsetmiterjoin%
\pgfsetlinewidth{1.003750pt}%
\definecolor{currentstroke}{rgb}{1.000000,1.000000,1.000000}%
\pgfsetstrokecolor{currentstroke}%
\pgfsetdash{}{0pt}%
\pgfpathmoveto{\pgfqpoint{1.168074in}{0.738693in}}%
\pgfpathlineto{\pgfqpoint{1.168074in}{1.463678in}}%
\pgfusepath{stroke}%
\end{pgfscope}%
\begin{pgfscope}%
\pgfsetrectcap%
\pgfsetmiterjoin%
\pgfsetlinewidth{1.003750pt}%
\definecolor{currentstroke}{rgb}{1.000000,1.000000,1.000000}%
\pgfsetstrokecolor{currentstroke}%
\pgfsetdash{}{0pt}%
\pgfpathmoveto{\pgfqpoint{1.669816in}{0.738693in}}%
\pgfpathlineto{\pgfqpoint{1.669816in}{1.463678in}}%
\pgfusepath{stroke}%
\end{pgfscope}%
\begin{pgfscope}%
\pgfsetrectcap%
\pgfsetmiterjoin%
\pgfsetlinewidth{1.003750pt}%
\definecolor{currentstroke}{rgb}{1.000000,1.000000,1.000000}%
\pgfsetstrokecolor{currentstroke}%
\pgfsetdash{}{0pt}%
\pgfpathmoveto{\pgfqpoint{1.168074in}{0.738693in}}%
\pgfpathlineto{\pgfqpoint{1.669816in}{0.738693in}}%
\pgfusepath{stroke}%
\end{pgfscope}%
\begin{pgfscope}%
\pgfsetrectcap%
\pgfsetmiterjoin%
\pgfsetlinewidth{1.003750pt}%
\definecolor{currentstroke}{rgb}{1.000000,1.000000,1.000000}%
\pgfsetstrokecolor{currentstroke}%
\pgfsetdash{}{0pt}%
\pgfpathmoveto{\pgfqpoint{1.168074in}{1.463678in}}%
\pgfpathlineto{\pgfqpoint{1.669816in}{1.463678in}}%
\pgfusepath{stroke}%
\end{pgfscope}%
\begin{pgfscope}%
\pgfsetbuttcap%
\pgfsetmiterjoin%
\definecolor{currentfill}{rgb}{0.898039,0.898039,0.898039}%
\pgfsetfillcolor{currentfill}%
\pgfsetlinewidth{0.000000pt}%
\definecolor{currentstroke}{rgb}{0.000000,0.000000,0.000000}%
\pgfsetstrokecolor{currentstroke}%
\pgfsetstrokeopacity{0.000000}%
\pgfsetdash{}{0pt}%
\pgfpathmoveto{\pgfqpoint{1.939850in}{0.738693in}}%
\pgfpathlineto{\pgfqpoint{2.441593in}{0.738693in}}%
\pgfpathlineto{\pgfqpoint{2.441593in}{1.463678in}}%
\pgfpathlineto{\pgfqpoint{1.939850in}{1.463678in}}%
\pgfpathlineto{\pgfqpoint{1.939850in}{0.738693in}}%
\pgfpathclose%
\pgfusepath{fill}%
\end{pgfscope}%
\begin{pgfscope}%
\definecolor{textcolor}{rgb}{0.333333,0.333333,0.333333}%
\pgfsetstrokecolor{textcolor}%
\pgfsetfillcolor{textcolor}%
\pgftext[x=2.190721in,y=0.683137in,,top]{\color{textcolor}{\rmfamily\fontsize{9.000000}{10.800000}\selectfont\catcode`\^=\active\def^{\ifmmode\sp\else\^{}\fi}\catcode`\%=\active\def
\end{pgfscope}%
\begin{pgfscope}%
\pgfpathrectangle{\pgfqpoint{1.939850in}{0.738693in}}{\pgfqpoint{0.501742in}{0.724986in}}%
\pgfusepath{clip}%
\pgfsetrectcap%
\pgfsetroundjoin%
\pgfsetlinewidth{0.803000pt}%
\definecolor{currentstroke}{rgb}{1.000000,1.000000,1.000000}%
\pgfsetstrokecolor{currentstroke}%
\pgfsetdash{}{0pt}%
\pgfpathmoveto{\pgfqpoint{1.939850in}{0.912281in}}%
\pgfpathlineto{\pgfqpoint{2.441593in}{0.912281in}}%
\pgfusepath{stroke}%
\end{pgfscope}%
\begin{pgfscope}%
\pgfsetbuttcap%
\pgfsetroundjoin%
\definecolor{currentfill}{rgb}{0.333333,0.333333,0.333333}%
\pgfsetfillcolor{currentfill}%
\pgfsetlinewidth{0.803000pt}%
\definecolor{currentstroke}{rgb}{0.333333,0.333333,0.333333}%
\pgfsetstrokecolor{currentstroke}%
\pgfsetdash{}{0pt}%
\pgfsys@defobject{currentmarker}{\pgfqpoint{-0.048611in}{0.000000in}}{\pgfqpoint{-0.000000in}{0.000000in}}{%
\pgfpathmoveto{\pgfqpoint{-0.000000in}{0.000000in}}%
\pgfpathlineto{\pgfqpoint{-0.048611in}{0.000000in}}%
\pgfusepath{stroke,fill}%
}%
\begin{pgfscope}%
\pgfsys@transformshift{1.939850in}{0.912281in}%
\pgfsys@useobject{currentmarker}{}%
\end{pgfscope}%
\end{pgfscope}%
\begin{pgfscope}%
\definecolor{textcolor}{rgb}{0.333333,0.333333,0.333333}%
\pgfsetstrokecolor{textcolor}%
\pgfsetfillcolor{textcolor}%
\pgftext[x=1.731902in, y=0.878523in, left, base]{\color{textcolor}{\rmfamily\fontsize{7.000000}{8.400000}\selectfont\catcode`\^=\active\def^{\ifmmode\sp\else\^{}\fi}\catcode`\%=\active\def
\end{pgfscope}%
\begin{pgfscope}%
\pgfpathrectangle{\pgfqpoint{1.939850in}{0.738693in}}{\pgfqpoint{0.501742in}{0.724986in}}%
\pgfusepath{clip}%
\pgfsetrectcap%
\pgfsetroundjoin%
\pgfsetlinewidth{0.803000pt}%
\definecolor{currentstroke}{rgb}{1.000000,1.000000,1.000000}%
\pgfsetstrokecolor{currentstroke}%
\pgfsetdash{}{0pt}%
\pgfpathmoveto{\pgfqpoint{1.939850in}{1.249550in}}%
\pgfpathlineto{\pgfqpoint{2.441593in}{1.249550in}}%
\pgfusepath{stroke}%
\end{pgfscope}%
\begin{pgfscope}%
\pgfsetbuttcap%
\pgfsetroundjoin%
\definecolor{currentfill}{rgb}{0.333333,0.333333,0.333333}%
\pgfsetfillcolor{currentfill}%
\pgfsetlinewidth{0.803000pt}%
\definecolor{currentstroke}{rgb}{0.333333,0.333333,0.333333}%
\pgfsetstrokecolor{currentstroke}%
\pgfsetdash{}{0pt}%
\pgfsys@defobject{currentmarker}{\pgfqpoint{-0.048611in}{0.000000in}}{\pgfqpoint{-0.000000in}{0.000000in}}{%
\pgfpathmoveto{\pgfqpoint{-0.000000in}{0.000000in}}%
\pgfpathlineto{\pgfqpoint{-0.048611in}{0.000000in}}%
\pgfusepath{stroke,fill}%
}%
\begin{pgfscope}%
\pgfsys@transformshift{1.939850in}{1.249550in}%
\pgfsys@useobject{currentmarker}{}%
\end{pgfscope}%
\end{pgfscope}%
\begin{pgfscope}%
\definecolor{textcolor}{rgb}{0.333333,0.333333,0.333333}%
\pgfsetstrokecolor{textcolor}%
\pgfsetfillcolor{textcolor}%
\pgftext[x=1.731902in, y=1.215792in, left, base]{\color{textcolor}{\rmfamily\fontsize{7.000000}{8.400000}\selectfont\catcode`\^=\active\def^{\ifmmode\sp\else\^{}\fi}\catcode`\%=\active\def
\end{pgfscope}%
\begin{pgfscope}%
\pgfpathrectangle{\pgfqpoint{1.939850in}{0.738693in}}{\pgfqpoint{0.501742in}{0.724986in}}%
\pgfusepath{clip}%
\pgfsetbuttcap%
\pgfsetmiterjoin%
\definecolor{currentfill}{rgb}{0.121569,0.466667,0.705882}%
\pgfsetfillcolor{currentfill}%
\pgfsetlinewidth{0.000000pt}%
\definecolor{currentstroke}{rgb}{0.000000,0.000000,0.000000}%
\pgfsetstrokecolor{currentstroke}%
\pgfsetstrokeopacity{0.000000}%
\pgfsetdash{}{0pt}%
\pgfpathmoveto{\pgfqpoint{2.021866in}{-10.554876in}}%
\pgfpathlineto{\pgfqpoint{2.070110in}{-10.554876in}}%
\pgfpathlineto{\pgfqpoint{2.070110in}{0.919939in}}%
\pgfpathlineto{\pgfqpoint{2.021866in}{0.919939in}}%
\pgfpathlineto{\pgfqpoint{2.021866in}{-10.554876in}}%
\pgfpathclose%
\pgfusepath{fill}%
\end{pgfscope}%
\begin{pgfscope}%
\pgfpathrectangle{\pgfqpoint{1.939850in}{0.738693in}}{\pgfqpoint{0.501742in}{0.724986in}}%
\pgfusepath{clip}%
\pgfsetbuttcap%
\pgfsetmiterjoin%
\definecolor{currentfill}{rgb}{1.000000,0.498039,0.054902}%
\pgfsetfillcolor{currentfill}%
\pgfsetlinewidth{0.000000pt}%
\definecolor{currentstroke}{rgb}{0.000000,0.000000,0.000000}%
\pgfsetstrokecolor{currentstroke}%
\pgfsetstrokeopacity{0.000000}%
\pgfsetdash{}{0pt}%
\pgfpathmoveto{\pgfqpoint{2.118355in}{-10.554876in}}%
\pgfpathlineto{\pgfqpoint{2.166599in}{-10.554876in}}%
\pgfpathlineto{\pgfqpoint{2.166599in}{1.008001in}}%
\pgfpathlineto{\pgfqpoint{2.118355in}{1.008001in}}%
\pgfpathlineto{\pgfqpoint{2.118355in}{-10.554876in}}%
\pgfpathclose%
\pgfusepath{fill}%
\end{pgfscope}%
\begin{pgfscope}%
\pgfpathrectangle{\pgfqpoint{1.939850in}{0.738693in}}{\pgfqpoint{0.501742in}{0.724986in}}%
\pgfusepath{clip}%
\pgfsetbuttcap%
\pgfsetmiterjoin%
\definecolor{currentfill}{rgb}{0.172549,0.627451,0.172549}%
\pgfsetfillcolor{currentfill}%
\pgfsetlinewidth{0.000000pt}%
\definecolor{currentstroke}{rgb}{0.000000,0.000000,0.000000}%
\pgfsetstrokecolor{currentstroke}%
\pgfsetstrokeopacity{0.000000}%
\pgfsetdash{}{0pt}%
\pgfpathmoveto{\pgfqpoint{2.214844in}{-10.554876in}}%
\pgfpathlineto{\pgfqpoint{2.263088in}{-10.554876in}}%
\pgfpathlineto{\pgfqpoint{2.263088in}{1.275650in}}%
\pgfpathlineto{\pgfqpoint{2.214844in}{1.275650in}}%
\pgfpathlineto{\pgfqpoint{2.214844in}{-10.554876in}}%
\pgfpathclose%
\pgfusepath{fill}%
\end{pgfscope}%
\begin{pgfscope}%
\pgfpathrectangle{\pgfqpoint{1.939850in}{0.738693in}}{\pgfqpoint{0.501742in}{0.724986in}}%
\pgfusepath{clip}%
\pgfsetbuttcap%
\pgfsetmiterjoin%
\definecolor{currentfill}{rgb}{0.839216,0.152941,0.156863}%
\pgfsetfillcolor{currentfill}%
\pgfsetlinewidth{0.000000pt}%
\definecolor{currentstroke}{rgb}{0.000000,0.000000,0.000000}%
\pgfsetstrokecolor{currentstroke}%
\pgfsetstrokeopacity{0.000000}%
\pgfsetdash{}{0pt}%
\pgfpathmoveto{\pgfqpoint{2.311333in}{-10.554876in}}%
\pgfpathlineto{\pgfqpoint{2.359577in}{-10.554876in}}%
\pgfpathlineto{\pgfqpoint{2.359577in}{1.282432in}}%
\pgfpathlineto{\pgfqpoint{2.311333in}{1.282432in}}%
\pgfpathlineto{\pgfqpoint{2.311333in}{-10.554876in}}%
\pgfpathclose%
\pgfusepath{fill}%
\end{pgfscope}%
\begin{pgfscope}%
\pgfsetrectcap%
\pgfsetmiterjoin%
\pgfsetlinewidth{1.003750pt}%
\definecolor{currentstroke}{rgb}{1.000000,1.000000,1.000000}%
\pgfsetstrokecolor{currentstroke}%
\pgfsetdash{}{0pt}%
\pgfpathmoveto{\pgfqpoint{1.939850in}{0.738693in}}%
\pgfpathlineto{\pgfqpoint{1.939850in}{1.463678in}}%
\pgfusepath{stroke}%
\end{pgfscope}%
\begin{pgfscope}%
\pgfsetrectcap%
\pgfsetmiterjoin%
\pgfsetlinewidth{1.003750pt}%
\definecolor{currentstroke}{rgb}{1.000000,1.000000,1.000000}%
\pgfsetstrokecolor{currentstroke}%
\pgfsetdash{}{0pt}%
\pgfpathmoveto{\pgfqpoint{2.441593in}{0.738693in}}%
\pgfpathlineto{\pgfqpoint{2.441593in}{1.463678in}}%
\pgfusepath{stroke}%
\end{pgfscope}%
\begin{pgfscope}%
\pgfsetrectcap%
\pgfsetmiterjoin%
\pgfsetlinewidth{1.003750pt}%
\definecolor{currentstroke}{rgb}{1.000000,1.000000,1.000000}%
\pgfsetstrokecolor{currentstroke}%
\pgfsetdash{}{0pt}%
\pgfpathmoveto{\pgfqpoint{1.939850in}{0.738693in}}%
\pgfpathlineto{\pgfqpoint{2.441593in}{0.738693in}}%
\pgfusepath{stroke}%
\end{pgfscope}%
\begin{pgfscope}%
\pgfsetrectcap%
\pgfsetmiterjoin%
\pgfsetlinewidth{1.003750pt}%
\definecolor{currentstroke}{rgb}{1.000000,1.000000,1.000000}%
\pgfsetstrokecolor{currentstroke}%
\pgfsetdash{}{0pt}%
\pgfpathmoveto{\pgfqpoint{1.939850in}{1.463678in}}%
\pgfpathlineto{\pgfqpoint{2.441593in}{1.463678in}}%
\pgfusepath{stroke}%
\end{pgfscope}%
\begin{pgfscope}%
\pgfsetbuttcap%
\pgfsetmiterjoin%
\definecolor{currentfill}{rgb}{0.898039,0.898039,0.898039}%
\pgfsetfillcolor{currentfill}%
\pgfsetlinewidth{0.000000pt}%
\definecolor{currentstroke}{rgb}{0.000000,0.000000,0.000000}%
\pgfsetstrokecolor{currentstroke}%
\pgfsetstrokeopacity{0.000000}%
\pgfsetdash{}{0pt}%
\pgfpathmoveto{\pgfqpoint{2.711627in}{0.738693in}}%
\pgfpathlineto{\pgfqpoint{3.213369in}{0.738693in}}%
\pgfpathlineto{\pgfqpoint{3.213369in}{1.463678in}}%
\pgfpathlineto{\pgfqpoint{2.711627in}{1.463678in}}%
\pgfpathlineto{\pgfqpoint{2.711627in}{0.738693in}}%
\pgfpathclose%
\pgfusepath{fill}%
\end{pgfscope}%
\begin{pgfscope}%
\definecolor{textcolor}{rgb}{0.333333,0.333333,0.333333}%
\pgfsetstrokecolor{textcolor}%
\pgfsetfillcolor{textcolor}%
\pgftext[x=2.962498in,y=0.683137in,,top]{\color{textcolor}{\rmfamily\fontsize{9.000000}{10.800000}\selectfont\catcode`\^=\active\def^{\ifmmode\sp\else\^{}\fi}\catcode`\%=\active\def
\end{pgfscope}%
\begin{pgfscope}%
\pgfpathrectangle{\pgfqpoint{2.711627in}{0.738693in}}{\pgfqpoint{0.501742in}{0.724986in}}%
\pgfusepath{clip}%
\pgfsetrectcap%
\pgfsetroundjoin%
\pgfsetlinewidth{0.803000pt}%
\definecolor{currentstroke}{rgb}{1.000000,1.000000,1.000000}%
\pgfsetstrokecolor{currentstroke}%
\pgfsetdash{}{0pt}%
\pgfpathmoveto{\pgfqpoint{2.711627in}{0.841086in}}%
\pgfpathlineto{\pgfqpoint{3.213369in}{0.841086in}}%
\pgfusepath{stroke}%
\end{pgfscope}%
\begin{pgfscope}%
\pgfsetbuttcap%
\pgfsetroundjoin%
\definecolor{currentfill}{rgb}{0.333333,0.333333,0.333333}%
\pgfsetfillcolor{currentfill}%
\pgfsetlinewidth{0.803000pt}%
\definecolor{currentstroke}{rgb}{0.333333,0.333333,0.333333}%
\pgfsetstrokecolor{currentstroke}%
\pgfsetdash{}{0pt}%
\pgfsys@defobject{currentmarker}{\pgfqpoint{-0.048611in}{0.000000in}}{\pgfqpoint{-0.000000in}{0.000000in}}{%
\pgfpathmoveto{\pgfqpoint{-0.000000in}{0.000000in}}%
\pgfpathlineto{\pgfqpoint{-0.048611in}{0.000000in}}%
\pgfusepath{stroke,fill}%
}%
\begin{pgfscope}%
\pgfsys@transformshift{2.711627in}{0.841086in}%
\pgfsys@useobject{currentmarker}{}%
\end{pgfscope}%
\end{pgfscope}%
\begin{pgfscope}%
\definecolor{textcolor}{rgb}{0.333333,0.333333,0.333333}%
\pgfsetstrokecolor{textcolor}%
\pgfsetfillcolor{textcolor}%
\pgftext[x=2.503679in, y=0.807329in, left, base]{\color{textcolor}{\rmfamily\fontsize{7.000000}{8.400000}\selectfont\catcode`\^=\active\def^{\ifmmode\sp\else\^{}\fi}\catcode`\%=\active\def
\end{pgfscope}%
\begin{pgfscope}%
\pgfpathrectangle{\pgfqpoint{2.711627in}{0.738693in}}{\pgfqpoint{0.501742in}{0.724986in}}%
\pgfusepath{clip}%
\pgfsetrectcap%
\pgfsetroundjoin%
\pgfsetlinewidth{0.803000pt}%
\definecolor{currentstroke}{rgb}{1.000000,1.000000,1.000000}%
\pgfsetstrokecolor{currentstroke}%
\pgfsetdash{}{0pt}%
\pgfpathmoveto{\pgfqpoint{2.711627in}{1.189643in}}%
\pgfpathlineto{\pgfqpoint{3.213369in}{1.189643in}}%
\pgfusepath{stroke}%
\end{pgfscope}%
\begin{pgfscope}%
\pgfsetbuttcap%
\pgfsetroundjoin%
\definecolor{currentfill}{rgb}{0.333333,0.333333,0.333333}%
\pgfsetfillcolor{currentfill}%
\pgfsetlinewidth{0.803000pt}%
\definecolor{currentstroke}{rgb}{0.333333,0.333333,0.333333}%
\pgfsetstrokecolor{currentstroke}%
\pgfsetdash{}{0pt}%
\pgfsys@defobject{currentmarker}{\pgfqpoint{-0.048611in}{0.000000in}}{\pgfqpoint{-0.000000in}{0.000000in}}{%
\pgfpathmoveto{\pgfqpoint{-0.000000in}{0.000000in}}%
\pgfpathlineto{\pgfqpoint{-0.048611in}{0.000000in}}%
\pgfusepath{stroke,fill}%
}%
\begin{pgfscope}%
\pgfsys@transformshift{2.711627in}{1.189643in}%
\pgfsys@useobject{currentmarker}{}%
\end{pgfscope}%
\end{pgfscope}%
\begin{pgfscope}%
\definecolor{textcolor}{rgb}{0.333333,0.333333,0.333333}%
\pgfsetstrokecolor{textcolor}%
\pgfsetfillcolor{textcolor}%
\pgftext[x=2.503679in, y=1.155886in, left, base]{\color{textcolor}{\rmfamily\fontsize{7.000000}{8.400000}\selectfont\catcode`\^=\active\def^{\ifmmode\sp\else\^{}\fi}\catcode`\%=\active\def
\end{pgfscope}%
\begin{pgfscope}%
\pgfpathrectangle{\pgfqpoint{2.711627in}{0.738693in}}{\pgfqpoint{0.501742in}{0.724986in}}%
\pgfusepath{clip}%
\pgfsetbuttcap%
\pgfsetmiterjoin%
\definecolor{currentfill}{rgb}{0.121569,0.466667,0.705882}%
\pgfsetfillcolor{currentfill}%
\pgfsetlinewidth{0.000000pt}%
\definecolor{currentstroke}{rgb}{0.000000,0.000000,0.000000}%
\pgfsetstrokecolor{currentstroke}%
\pgfsetstrokeopacity{0.000000}%
\pgfsetdash{}{0pt}%
\pgfpathmoveto{\pgfqpoint{2.793642in}{-1.598814in}}%
\pgfpathlineto{\pgfqpoint{2.841887in}{-1.598814in}}%
\pgfpathlineto{\pgfqpoint{2.841887in}{0.919939in}}%
\pgfpathlineto{\pgfqpoint{2.793642in}{0.919939in}}%
\pgfpathlineto{\pgfqpoint{2.793642in}{-1.598814in}}%
\pgfpathclose%
\pgfusepath{fill}%
\end{pgfscope}%
\begin{pgfscope}%
\pgfpathrectangle{\pgfqpoint{2.711627in}{0.738693in}}{\pgfqpoint{0.501742in}{0.724986in}}%
\pgfusepath{clip}%
\pgfsetbuttcap%
\pgfsetmiterjoin%
\definecolor{currentfill}{rgb}{1.000000,0.498039,0.054902}%
\pgfsetfillcolor{currentfill}%
\pgfsetlinewidth{0.000000pt}%
\definecolor{currentstroke}{rgb}{0.000000,0.000000,0.000000}%
\pgfsetstrokecolor{currentstroke}%
\pgfsetstrokeopacity{0.000000}%
\pgfsetdash{}{0pt}%
\pgfpathmoveto{\pgfqpoint{2.890131in}{-1.598814in}}%
\pgfpathlineto{\pgfqpoint{2.938376in}{-1.598814in}}%
\pgfpathlineto{\pgfqpoint{2.938376in}{0.982836in}}%
\pgfpathlineto{\pgfqpoint{2.890131in}{0.982836in}}%
\pgfpathlineto{\pgfqpoint{2.890131in}{-1.598814in}}%
\pgfpathclose%
\pgfusepath{fill}%
\end{pgfscope}%
\begin{pgfscope}%
\pgfpathrectangle{\pgfqpoint{2.711627in}{0.738693in}}{\pgfqpoint{0.501742in}{0.724986in}}%
\pgfusepath{clip}%
\pgfsetbuttcap%
\pgfsetmiterjoin%
\definecolor{currentfill}{rgb}{0.172549,0.627451,0.172549}%
\pgfsetfillcolor{currentfill}%
\pgfsetlinewidth{0.000000pt}%
\definecolor{currentstroke}{rgb}{0.000000,0.000000,0.000000}%
\pgfsetstrokecolor{currentstroke}%
\pgfsetstrokeopacity{0.000000}%
\pgfsetdash{}{0pt}%
\pgfpathmoveto{\pgfqpoint{2.986620in}{-1.598814in}}%
\pgfpathlineto{\pgfqpoint{3.034865in}{-1.598814in}}%
\pgfpathlineto{\pgfqpoint{3.034865in}{1.282432in}}%
\pgfpathlineto{\pgfqpoint{2.986620in}{1.282432in}}%
\pgfpathlineto{\pgfqpoint{2.986620in}{-1.598814in}}%
\pgfpathclose%
\pgfusepath{fill}%
\end{pgfscope}%
\begin{pgfscope}%
\pgfpathrectangle{\pgfqpoint{2.711627in}{0.738693in}}{\pgfqpoint{0.501742in}{0.724986in}}%
\pgfusepath{clip}%
\pgfsetbuttcap%
\pgfsetmiterjoin%
\definecolor{currentfill}{rgb}{0.839216,0.152941,0.156863}%
\pgfsetfillcolor{currentfill}%
\pgfsetlinewidth{0.000000pt}%
\definecolor{currentstroke}{rgb}{0.000000,0.000000,0.000000}%
\pgfsetstrokecolor{currentstroke}%
\pgfsetstrokeopacity{0.000000}%
\pgfsetdash{}{0pt}%
\pgfpathmoveto{\pgfqpoint{3.083109in}{-1.598814in}}%
\pgfpathlineto{\pgfqpoint{3.131354in}{-1.598814in}}%
\pgfpathlineto{\pgfqpoint{3.131354in}{1.282074in}}%
\pgfpathlineto{\pgfqpoint{3.083109in}{1.282074in}}%
\pgfpathlineto{\pgfqpoint{3.083109in}{-1.598814in}}%
\pgfpathclose%
\pgfusepath{fill}%
\end{pgfscope}%
\begin{pgfscope}%
\pgfsetrectcap%
\pgfsetmiterjoin%
\pgfsetlinewidth{1.003750pt}%
\definecolor{currentstroke}{rgb}{1.000000,1.000000,1.000000}%
\pgfsetstrokecolor{currentstroke}%
\pgfsetdash{}{0pt}%
\pgfpathmoveto{\pgfqpoint{2.711627in}{0.738693in}}%
\pgfpathlineto{\pgfqpoint{2.711627in}{1.463678in}}%
\pgfusepath{stroke}%
\end{pgfscope}%
\begin{pgfscope}%
\pgfsetrectcap%
\pgfsetmiterjoin%
\pgfsetlinewidth{1.003750pt}%
\definecolor{currentstroke}{rgb}{1.000000,1.000000,1.000000}%
\pgfsetstrokecolor{currentstroke}%
\pgfsetdash{}{0pt}%
\pgfpathmoveto{\pgfqpoint{3.213369in}{0.738693in}}%
\pgfpathlineto{\pgfqpoint{3.213369in}{1.463678in}}%
\pgfusepath{stroke}%
\end{pgfscope}%
\begin{pgfscope}%
\pgfsetrectcap%
\pgfsetmiterjoin%
\pgfsetlinewidth{1.003750pt}%
\definecolor{currentstroke}{rgb}{1.000000,1.000000,1.000000}%
\pgfsetstrokecolor{currentstroke}%
\pgfsetdash{}{0pt}%
\pgfpathmoveto{\pgfqpoint{2.711627in}{0.738693in}}%
\pgfpathlineto{\pgfqpoint{3.213369in}{0.738693in}}%
\pgfusepath{stroke}%
\end{pgfscope}%
\begin{pgfscope}%
\pgfsetrectcap%
\pgfsetmiterjoin%
\pgfsetlinewidth{1.003750pt}%
\definecolor{currentstroke}{rgb}{1.000000,1.000000,1.000000}%
\pgfsetstrokecolor{currentstroke}%
\pgfsetdash{}{0pt}%
\pgfpathmoveto{\pgfqpoint{2.711627in}{1.463678in}}%
\pgfpathlineto{\pgfqpoint{3.213369in}{1.463678in}}%
\pgfusepath{stroke}%
\end{pgfscope}%
\begin{pgfscope}%
\pgfsetbuttcap%
\pgfsetmiterjoin%
\definecolor{currentfill}{rgb}{0.898039,0.898039,0.898039}%
\pgfsetfillcolor{currentfill}%
\pgfsetfillopacity{0.800000}%
\pgfsetlinewidth{0.501875pt}%
\definecolor{currentstroke}{rgb}{0.800000,0.800000,0.800000}%
\pgfsetstrokecolor{currentstroke}%
\pgfsetstrokeopacity{0.800000}%
\pgfsetdash{}{0pt}%
\pgfpathmoveto{\pgfqpoint{0.467955in}{0.100000in}}%
\pgfpathlineto{\pgfqpoint{2.934177in}{0.100000in}}%
\pgfpathquadraticcurveto{\pgfqpoint{2.959177in}{0.100000in}}{\pgfqpoint{2.959177in}{0.125000in}}%
\pgfpathlineto{\pgfqpoint{2.959177in}{0.461111in}}%
\pgfpathquadraticcurveto{\pgfqpoint{2.959177in}{0.486111in}}{\pgfqpoint{2.934177in}{0.486111in}}%
\pgfpathlineto{\pgfqpoint{0.467955in}{0.486111in}}%
\pgfpathquadraticcurveto{\pgfqpoint{0.442955in}{0.486111in}}{\pgfqpoint{0.442955in}{0.461111in}}%
\pgfpathlineto{\pgfqpoint{0.442955in}{0.125000in}}%
\pgfpathquadraticcurveto{\pgfqpoint{0.442955in}{0.100000in}}{\pgfqpoint{0.467955in}{0.100000in}}%
\pgfpathlineto{\pgfqpoint{0.467955in}{0.100000in}}%
\pgfpathclose%
\pgfusepath{stroke,fill}%
\end{pgfscope}%
\begin{pgfscope}%
\pgfsetbuttcap%
\pgfsetmiterjoin%
\definecolor{currentfill}{rgb}{0.121569,0.466667,0.705882}%
\pgfsetfillcolor{currentfill}%
\pgfsetlinewidth{1.003750pt}%
\definecolor{currentstroke}{rgb}{0.121569,0.466667,0.705882}%
\pgfsetstrokecolor{currentstroke}%
\pgfsetdash{}{0pt}%
\pgfsys@defobject{currentmarker}{\pgfqpoint{-0.041667in}{-0.041667in}}{\pgfqpoint{0.041667in}{0.041667in}}{%
\pgfpathmoveto{\pgfqpoint{-0.041667in}{-0.041667in}}%
\pgfpathlineto{\pgfqpoint{0.041667in}{-0.041667in}}%
\pgfpathlineto{\pgfqpoint{0.041667in}{0.041667in}}%
\pgfpathlineto{\pgfqpoint{-0.041667in}{0.041667in}}%
\pgfpathlineto{\pgfqpoint{-0.041667in}{-0.041667in}}%
\pgfpathclose%
\pgfusepath{stroke,fill}%
}%
\begin{pgfscope}%
\pgfsys@transformshift{0.617955in}{0.392361in}%
\pgfsys@useobject{currentmarker}{}%
\end{pgfscope}%
\end{pgfscope}%
\begin{pgfscope}%
\definecolor{textcolor}{rgb}{0.000000,0.000000,0.000000}%
\pgfsetstrokecolor{textcolor}%
\pgfsetfillcolor{textcolor}%
\pgftext[x=0.842955in,y=0.348611in,left,base]{\color{textcolor}{\rmfamily\fontsize{9.000000}{10.800000}\selectfont\catcode`\^=\active\def^{\ifmmode\sp\else\^{}\fi}\catcode`\%=\active\def
\end{pgfscope}%
\begin{pgfscope}%
\pgfsetbuttcap%
\pgfsetmiterjoin%
\definecolor{currentfill}{rgb}{1.000000,0.498039,0.054902}%
\pgfsetfillcolor{currentfill}%
\pgfsetlinewidth{1.003750pt}%
\definecolor{currentstroke}{rgb}{1.000000,0.498039,0.054902}%
\pgfsetstrokecolor{currentstroke}%
\pgfsetdash{}{0pt}%
\pgfsys@defobject{currentmarker}{\pgfqpoint{-0.041667in}{-0.041667in}}{\pgfqpoint{0.041667in}{0.041667in}}{%
\pgfpathmoveto{\pgfqpoint{-0.041667in}{-0.041667in}}%
\pgfpathlineto{\pgfqpoint{0.041667in}{-0.041667in}}%
\pgfpathlineto{\pgfqpoint{0.041667in}{0.041667in}}%
\pgfpathlineto{\pgfqpoint{-0.041667in}{0.041667in}}%
\pgfpathlineto{\pgfqpoint{-0.041667in}{-0.041667in}}%
\pgfpathclose%
\pgfusepath{stroke,fill}%
}%
\begin{pgfscope}%
\pgfsys@transformshift{0.617955in}{0.218056in}%
\pgfsys@useobject{currentmarker}{}%
\end{pgfscope}%
\end{pgfscope}%
\begin{pgfscope}%
\definecolor{textcolor}{rgb}{0.000000,0.000000,0.000000}%
\pgfsetstrokecolor{textcolor}%
\pgfsetfillcolor{textcolor}%
\pgftext[x=0.842955in,y=0.174306in,left,base]{\color{textcolor}{\rmfamily\fontsize{9.000000}{10.800000}\selectfont\catcode`\^=\active\def^{\ifmmode\sp\else\^{}\fi}\catcode`\%=\active\def
\end{pgfscope}%
\begin{pgfscope}%
\pgfsetbuttcap%
\pgfsetmiterjoin%
\definecolor{currentfill}{rgb}{0.172549,0.627451,0.172549}%
\pgfsetfillcolor{currentfill}%
\pgfsetlinewidth{1.003750pt}%
\definecolor{currentstroke}{rgb}{0.172549,0.627451,0.172549}%
\pgfsetstrokecolor{currentstroke}%
\pgfsetdash{}{0pt}%
\pgfsys@defobject{currentmarker}{\pgfqpoint{-0.041667in}{-0.041667in}}{\pgfqpoint{0.041667in}{0.041667in}}{%
\pgfpathmoveto{\pgfqpoint{-0.041667in}{-0.041667in}}%
\pgfpathlineto{\pgfqpoint{0.041667in}{-0.041667in}}%
\pgfpathlineto{\pgfqpoint{0.041667in}{0.041667in}}%
\pgfpathlineto{\pgfqpoint{-0.041667in}{0.041667in}}%
\pgfpathlineto{\pgfqpoint{-0.041667in}{-0.041667in}}%
\pgfpathclose%
\pgfusepath{stroke,fill}%
}%
\begin{pgfscope}%
\pgfsys@transformshift{1.982629in}{0.392361in}%
\pgfsys@useobject{currentmarker}{}%
\end{pgfscope}%
\end{pgfscope}%
\begin{pgfscope}%
\definecolor{textcolor}{rgb}{0.000000,0.000000,0.000000}%
\pgfsetstrokecolor{textcolor}%
\pgfsetfillcolor{textcolor}%
\pgftext[x=2.207629in,y=0.348611in,left,base]{\color{textcolor}{\rmfamily\fontsize{9.000000}{10.800000}\selectfont\catcode`\^=\active\def^{\ifmmode\sp\else\^{}\fi}\catcode`\%=\active\def
\end{pgfscope}%
\begin{pgfscope}%
\pgfsetbuttcap%
\pgfsetmiterjoin%
\definecolor{currentfill}{rgb}{0.839216,0.152941,0.156863}%
\pgfsetfillcolor{currentfill}%
\pgfsetlinewidth{1.003750pt}%
\definecolor{currentstroke}{rgb}{0.839216,0.152941,0.156863}%
\pgfsetstrokecolor{currentstroke}%
\pgfsetdash{}{0pt}%
\pgfsys@defobject{currentmarker}{\pgfqpoint{-0.041667in}{-0.041667in}}{\pgfqpoint{0.041667in}{0.041667in}}{%
\pgfpathmoveto{\pgfqpoint{-0.041667in}{-0.041667in}}%
\pgfpathlineto{\pgfqpoint{0.041667in}{-0.041667in}}%
\pgfpathlineto{\pgfqpoint{0.041667in}{0.041667in}}%
\pgfpathlineto{\pgfqpoint{-0.041667in}{0.041667in}}%
\pgfpathlineto{\pgfqpoint{-0.041667in}{-0.041667in}}%
\pgfpathclose%
\pgfusepath{stroke,fill}%
}%
\begin{pgfscope}%
\pgfsys@transformshift{1.982629in}{0.218056in}%
\pgfsys@useobject{currentmarker}{}%
\end{pgfscope}%
\end{pgfscope}%
\begin{pgfscope}%
\definecolor{textcolor}{rgb}{0.000000,0.000000,0.000000}%
\pgfsetstrokecolor{textcolor}%
\pgfsetfillcolor{textcolor}%
\pgftext[x=2.207629in,y=0.174306in,left,base]{\color{textcolor}{\rmfamily\fontsize{9.000000}{10.800000}\selectfont\catcode`\^=\active\def^{\ifmmode\sp\else\^{}\fi}\catcode`\%=\active\def
\end{pgfscope}%
\end{pgfpicture}%
\makeatother%
\endgroup%

%% file: sections/06_human_eval.tex
\section{Human Evaluation}\label{sec:human_eval}

To gain further insight into the practical applicability beyond the automatic metrics, we conduct a human case study with two patent attorneys for the field of AI.
We randomly select 10 samples from the NLP
field out of the \DatasetName test set and use the outputs of Qwen2-72B with \MethodName and default token allocation.
For each of the samples, the attorneys are provided with the paper, the outline, the generated patent and the reference patent.
They are then tasked to evaluate the quality of the generated patent based on the hypothetical scenario that they got the paper, wrote the outline and sent both to the LLM.
We ask about strengths, weaknesses and potential for time savings.
The attorneys evaluated 10 and 5 samples, respectively, yielding 15 total evaluations and spending about 30 minutes per sample on average.

Among the 15 evaluations, the attorneys saw substantial time savings in 8 cases, showing promising potential to increase their productivity. 
However, they also identify two main limitations that could hinder such time savings, which are consistent with observations from our automatic evaluation, further validating our chosen metrics. 
These limitations are:

\begin{enumerate}[label=(\arabic*),itemsep=5pt,topsep=5pt,leftmargin=*,wide,labelindent=0pt]
     
\item
\textbf{Style}: The model often fails to use non-limiting language, which is essential in the patent description to avoid unnecessary limitations.
For instance, stating that \textit{\enquote{the system includes a crucial component}} may be too limiting, whereas rephrasing it as \textit{\enquote{the system may include a preferred component}} provides more flexibility to adapt claims in the grant procedure and does not narrow the claims' interpreted scope of protection. 
Furthermore, the model occasionally uses promotional phrases like \textit{\enquote{have revolutionized}} that are common in NLP papers but inappropriate in patents.

\item
\textbf{Level of detail}: The patent description should describe the invention precisely enough to enable an ordinary person skilled in the art to reproduce it.
This level of detail is often still not generated, which is also in line with the lower absolute scores for coverage compared to those for factuality.

\end{enumerate}

%% file: sections/07_conclusion.tex
\section{Conclusion}

In this work, we have presented the first study on the generation of complete patent descriptions in a realistic experimental setting, using papers as real-world invention specifications.
We create the \DatasetName{} benchmark, propose targeted evaluation metrics, develop \MethodName to enable the generation of long patent documents, and conduct a human evaluation.
We find that LLMs can effectively generate patent drafts from research papers, but that there is still headroom for improving the level of detail and linguistic style.
Promising directions for future research include developing fine-tuning methods that maintain factuality and avoid repetitions, as well as conducting user studies to identify efficient interaction patterns.

%% file: sections/08_limitations_and_acknowledgements.tex
\section*{Limitations}

This study focuses on open-weight models, leaving the exploration of closed-source commercial models as a potential avenue for future research. 
Expanding our experimental setting to include these models could provide additional insights.

In this work, we do not study the automatic generation of outlines, but consider them user input.
Generating them in a planning stage from the paper or other input could help decrease manual effort further.

Due to the immense cost of patent attorneys, our expert case study analyses only a small sample of the dataset.
Conducting large-scale human evaluations could generate further insights.

\section*{Acknowledgments}

We would like to thank the patent attorneys Philipp Mangold and Charlotte Hellmann for evaluating the quality of generated patents, and for providing valuable insights into the peculiarities of patents.

\section*{Ethics Statement}

The development of AI-powered patent generation tools may raise ethical concerns. 
We emphasize that our research is intended to support and augment the work of patent professionals, rather than to fully automate the patenting process or facilitate malicious activities such as patent trolling.

%% file: sections/09_appendix.tex
\appendix

\input{sections/091_dataset_appendix.tex}

\FloatBarrier
\section{Document Parsing Pipeline}\label{sec:parsing_appendix}
\FloatBarrier

In total, we use three different data sources for the full texts.
For the patents, we use the USPTO bulk downloads\footnote{\url{https://bulkdata.uspto.gov/}}.
For the papers, we use PubMed if available and PDF otherwise.
The goal is to extract a clean representation of the full text into a common JSON format.
In this format, every section has a title, a list of paragraphs and a list of subsections.
We write parsers for the XML formats from USPTO, PubMed and the PDF parser Grobid\footnote{\url{https://github.com/kermitt2/grobid}}.
In addition, we perform several cleaning steps:\begin{enumerate}
  \item \textbf{PDF Hierarchy Reconstruction. } The JSON format is hierarchical by nature,  for instance to enable better chunking of patents and section-based retrieval from papers. However, Grobid does not detect section levels if the sections are not numbered. To reconstruct the levels in these cases, we implement a solution that searches for the headings in the PDF file, extracts their font properties, orders them by size, boldness and capitalization, and infers the levels from that.
  \item \textbf{Patent Hierarchy Reconstruction. } In USPTO's patent XML files, there is a level attribute associated with every heading, but we find that it is rarely correct; most headings are placed on the same level. To reconstruct the levels, we pass the list of headings to an LLM and instruct it to infer the levels based on their names (e.g., "Example 1" and "Example 2" and likely children of "EXAMPLES").
  \item \textbf{Formula Conversion. } Many patents and papers include formulas that can be an important part of the document. However, in USPTO's and PubMed's XML formats, formulas are represented in MathML syntax, which is extremely hard to read and arguably hard to generate. To that end, we convert all MathML formulas to latex using pandoc\footnote{\url{https://github.com/jgm/pandoc}}.
  \item \textbf{Metadata Section Filtering. } Patents usually contain a number of metadata sections in the full text, such as information regarding funding or cross-references to related patents. To filter out these sections, we collect a list of such heading names and remove a section if its heading has a Levenshtein distance less than 3 to any one of the blacklisted headings.
\end{enumerate}

\FloatBarrier
\clearpage
\section{Repetition Removal}\label{sec:rep_removal}
\FloatBarrier

\begin{algorithm*}[ht!]
  \small
  \caption{Repetition Detection}
  \label{alg:rep_removal}
  \begin{algorithmic}[1]
    \Function{matches}{l1, l2}
      \State $n\_match \gets 0$
      \For{$i \gets 0$ \textbf{to} $\text{len}(l1) - 1$}
        \If{$l1[i] = l2[i$]}
          \State $n\_match \gets n\_match + 1$
        \EndIf
      \EndFor
      \State \Return n\_match / len(l1) $>$ 0.9
    \EndFunction
    \\
    \Function{detect\_repetitions}{words, min\_length, max\_cycle\_length}
        \State $n \gets len(words)$
        \For{$k = 1$ to max\_cycle\_length}
            \If{matches(words[-k:], words[-2 * k : -k])} \Comment{Check for repetition of length k}
                \State $i \gets 2$ \Comment{Number of times the pattern is repeated}
                \While{matches(words[-k:], words[-(i + 1) * k : -i * k])} \Comment{Search for additional occurrences}
                    \State $i \gets i + 1$
                \EndWhile
                
                \State remove\_indices $\gets$ [n - (i - 1) * k, ..., n]
                \State total\_length $\gets i * k$

                \If{total\_length $\ge$ min\_length}
                    \State words $\gets$ words[: n - (i - 1) * k]  \Comment{Keep only first occurrence of pattern}
                    \State remove\_rest $\gets$ detect\_repetitions(words, min\_length, max\_cycle\_length) \Comment{Recursive call for remaining text}
                    \State \Return remove\_indices + remove\_rest \Comment{Return indices to remove}
                \EndIf
            \EndIf
        \EndFor
        \State \Return []
    \EndFunction
  \end{algorithmic}
\end{algorithm*}
\FloatBarrier
We design a procedure to remove infinite repetitions from generated outputs to study their effect on evaluation metrics.  
In that context, we characterize an infinite repetition as a sequence of tokens that appears multiple times until the end of the generation.  
To find such repetitions, we apply the following recursive algorithm in \Cref{alg:rep_removal} to each chunk and remove the returned indices.
To account for reptitions where the model alters the patterns slightly (e.g., incrementing a number) in each iteration, we consider two word sequences equal if 90\% of their positions are equal.
By default, we use min\_length = 50 and max\_cycle\_length = 300. 

\FloatBarrier
\section{Contamination Results}\label{sec:nc-test-results}
\FloatBarrier

\input{sections/main_table_nc-test}

\FloatBarrier
\renewcommand{\sparqlbasicstyle}{\tiny\tt}
\begin{figure*}
  \section{Patent Outline Example}\label{sec:patent_outline_appendix}
  \begin{minipage}{\textwidth}
  \centering
\begin{lstlisting}[language=markdown, style=markdownstyle, caption={Example patent outline (short variant) for the pair W6364285-US20140112582. The outline corresponds to more than 5 pages. This example was randomly selected.}]
# DESCRIPTION

## CROSS-REFERENCE TO RELATED APPLICATIONS

- reference prior applications

## BACKGROUND

- limitations of current text recognition methods

## SUMMARY

- outline method and system for character recognition

## DETAILED DESCRIPTION

- introduce character recognition difficulties
- describe lateral approach to character recognition
- define views and bounding box
- explain binarization and noise removal
- describe oblique/skew detection and removal
- outline segmentation process
- explain lateral-view-based analysis and characteristic points selection
- describe generation of feature vector
- outline classification and recognition with Artificial Neural Network
- describe training and knowledge base of Artificial Neural Network
- summarize system and method block diagram

### Handling Compound Characters

- introduce compound characters
- motivate lateral view based approach
- discuss limitations of conventional character recognition algorithms
- clarify scope and interpretation of patent claims
\end{lstlisting}
\end{minipage}
\end{figure*}
\renewcommand{\sparqlbasicstyle}{\small\tt}

\FloatBarrier

\begin{figure*}
  \section{Patent Structure}\label{sec:patent_structure_appendix}
  \begin{minipage}{\textwidth}
  \includegraphics[width=\linewidth]{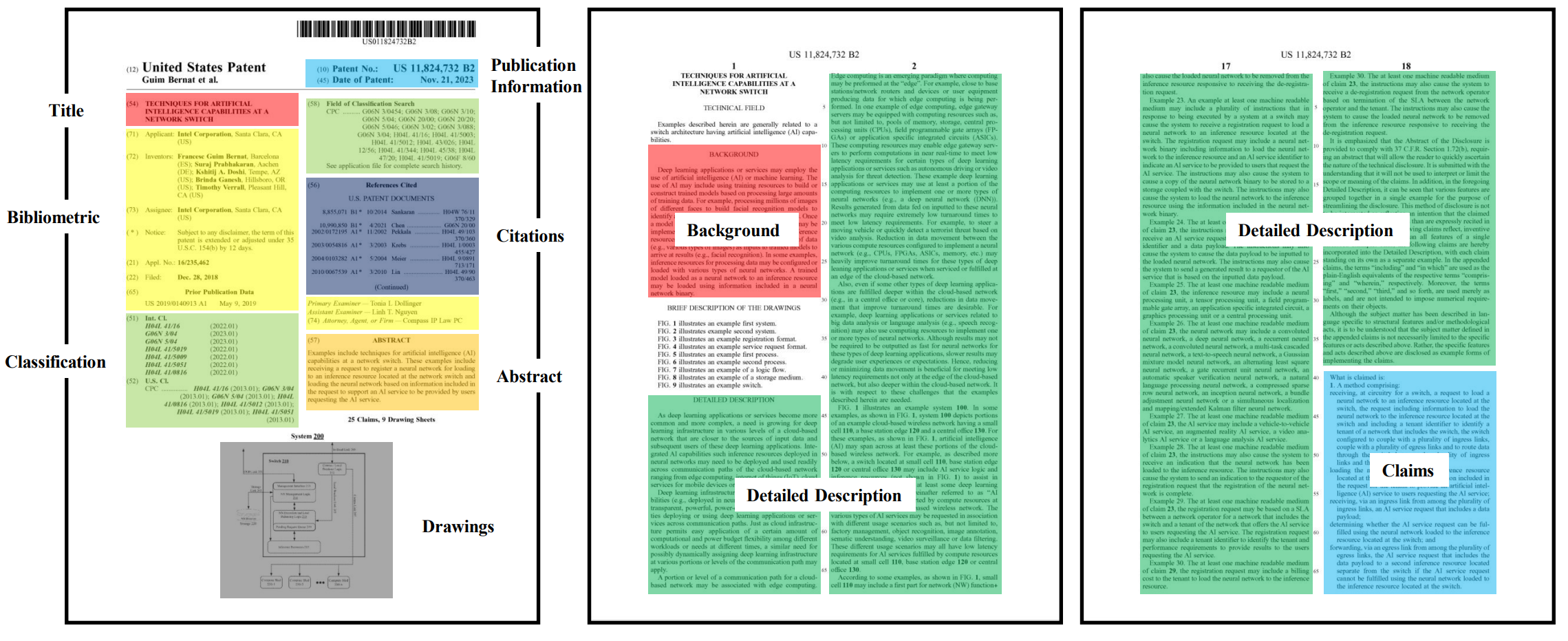}
  \caption{Illustration of the structure of a patent from \citet{jiangArtificialIntelligenceExploring2024}. Note that multiple pages from the detailed description are omitted. The description includes all sections except the front matter and claims. In our experiments, we exclude sections containing only metadata, such as statements regarding funding.}
\end{minipage}
\end{figure*}

\FloatBarrier

\begin{table*}
\begin{minipage}{\textwidth}
  \section{Hyperparameters}\label{sec:hyperparameter_appendix}
  \centering
  \begin{tabular}{llcll}
    \toprule
    \multicolumn{2}{c}{Generation} &       & \multicolumn{2}{c}{Training}                                    \\
    \cmidrule{1-2}\cmidrule{4-5}
    Parameter                      & Value &                              & Parameter           & Value      \\
    \midrule
    max sequence length            & 8192  &                              & max sequence length & 8192       \\
    temperature                    & 0.6   &                              & learning rate       & 0.00031622 \\
                                   &       &                              & scheduler           & cosine     \\
                                   &       &                              & warmup ratio        & 0.1        \\
                                   &       &                              & epochs              & 3          \\
                                   &       &                              & batch size          & 32         \\
                                   &       &                              & lora alpha          & 60         \\
                                   &       &                              & lora dropout        & 0.05       \\
                                   &       &                              & lora r              & 128        \\
    \bottomrule
  \end{tabular}
  \caption{Hyperparameters during generation and training. Training parameters are adopted from \citet{tribesHyperparameterOptimizationLarge2024}. We run the experiments on Nvidia H100 80GB GPUs. We use a single H100 for inference and training of the 8B model and 4xH100 for the inference of the larger models using tensor parallel. We estimate the total number of GPU-hours to be 720.}
\end{minipage}
\end{table*}

\FloatBarrier

\renewcommand{\sparqlbasicstyle}{\tiny\tt}
\begin{figure*}
  \section{SPARQL Query}\label{sec:sparql_appendix}
  \begin{minipage}{\textwidth}
  \centering
\begin{lstlisting}[language=SPARQL, style=sparqlstyle, caption={SPARQL query template used to retrieve papers for a given patent from SemOpenAlex. Template variables \texttt{<var>} are filled based on the query patent.}]
PREFIX fabio: <http://purl.org/spar/fabio/>
PREFIX dct: <http://purl.org/dc/terms/>
PREFIX soa: <https://semopenalex.org/ontology/>
PREFIX foaf: <http://xmlns.com/foaf/0.1/>
PREFIX datacite: <http://purl.org/spar/datacite/>

SELECT DISTINCT
	?paper ?authors ?author_names ?author_overlap ?title ?abstract ?date ?doi
	(GROUP_CONCAT(?location; SEPARATOR="\n") as ?locations)
	(GROUP_CONCAT(?url; SEPARATOR="\n") as ?urls)
	(GROUP_CONCAT(?pdf_url; SEPARATOR="\n") as ?pdf_urls)
	(GROUP_CONCAT(?license; SEPARATOR="\n") as ?licenses)

WHERE {
    SELECT DISTINCT
        ?paper
        (GROUP_CONCAT(DISTINCT ?author; SEPARATOR="\n") as ?authors)
        (GROUP_CONCAT(DISTINCT ?author_name; SEPARATOR="\n") as ?author_names)
        (SAMPLE(?author_overlap) as ?author_overlap)
        (SAMPLE(?title) as ?title)
        (SAMPLE(?abstract) as ?abstract)
        (SAMPLE(?date) as ?date)
        (SAMPLE(?doi) as ?doi)
        ?location
        ?url
        ?pdf_url
        ?license

    WHERE {
        # Information required for Matching
        ?paper a soa:Work ;
            dct:creator ?author ;
            dct:title ?title ;
            dct:abstract ?abstract ;
            dct:date ?date ;
            datacite:doi ?doi .

        ?author foaf:name ?author_name .

        # Optional Information for Downloading
        OPTIONAL {
            ?paper soa:hasLocation ?location .
            OPTIONAL { ?location fabio:hasURL ?url_ . }
            OPTIONAL { ?location soa:pdfUrl ?pdf_url_ . }
            OPTIONAL { ?location dct:license ?license_ . }
            BIND(COALESCE(?url_, "<EMPTY_PLACEHOLDER>") as ?url)
            BIND(COALESCE(?pdf_url_, "<EMPTY_PLACEHOLDER>") as ?pdf_url)
            BIND(COALESCE(?license_, "<EMPTY_PLACEHOLDER>") as ?license)
        }

        # Count number of matching authors
        {
            SELECT ?paper ?author_overlap
            WHERE {
                {
                    SELECT ?paper (COUNT(DISTINCT ?author) AS ?matching_authors)
                    WHERE {
                        ?paper dct:creator ?author .
                        ?author ?p ?name .
                        FILTER (?p IN (foaf:name, soa:alternativeName))
                        FILTER (?name IN (<AUTHOR_LIST>))
                    }
                    GROUP BY ?paper
                    ORDER BY DESC(?matching_authors)
                    LIMIT 500 # having too many results in the subquery will make query time out
                }

                ?paper dct:date ?date .

                BIND (xsd:float(?matching_authors) / xsd:float(<NUM_AUTHORS>) as ?author_overlap)
                FILTER (?author_overlap >= <AUTHOR_OVERLAP_THRESHOLD>)
                FILTER (?date > "<DATE_EARLIEST>"^^xsd:dateTime)
                FILTER (?date < "<DATE_LATEST>"^^xsd:dateTime)
            }
        }
    }
    GROUP BY ?paper ?location ?url ?pdf_url ?license
    HAVING (COUNT(?author) > 1)
}
GROUP BY ?paper ?authors ?author_names ?author_overlap ?title ?abstract ?date ?doi
\end{lstlisting}
\end{minipage}
\end{figure*}
\renewcommand{\sparqlbasicstyle}{\small\tt}

\FloatBarrier

\renewcommand{\sparqlbasicstyle}{\tiny\tt}
\begin{figure*}
  \section{Summary Generation Prompt}\label{sec:summary_prompt_appendix}
  \begin{minipage}{\textwidth}
  \centering
\begin{lstlisting}[language=SPARQL, style=sparqlstyle, caption={Prompt used to generate bullet point summaries with Llama-3 70B}]
<|begin_of_text|><|start_header_id|>system<|end_header_id|>

You are a highly skilled patent attorney with decades of experience in drafting high-quality patent applications. You answer every question in the most concise way possible, without adding unnecessary explanations.<|eot_id|><|start_header_id|>user<|end_header_id|>

### INSTRUCTION

For the sections of a patent application shown below, write a bullet list that summarizes the discourse structure of the document.

### OUTPUT FORMAT

The output needs to be in markdown syntax. Keep the headings as they are and add bullet lists summarizing the structure of each section. Do NOT write nested lists.

### GUIDELINES

Here are important guidelines you need to follow:

- **{n_words} words per bullet**: Every bullet point should summarize roughly {n_words} words in just a couple of words on a very high level.
- **Structure, not content**: The bullet points should not contain all the content. You should not write a summary of the content, but a summary of the structure! For instance, you should write 'motivate neural networks' rather than writing what the motivation for a neural network is. 
- **Start with verbal phrases**: If applicable, start the bullet points with phrases like 'define', 'motivate', 'summarize', 'limitations of', 'application of' or 'embodiment'. You are not restricted to this set of phrases, just use them as inspiration. Avoid overusing the phrase 'describe'. 
- **Specificity**: Avoid overly generic bullet points like 'define method' at all cost!
- **Conciseness**: Keep every bullet point as concise as possible! Do NOT write more than 5 words per bullet!
- **{n_bullets} bullet points in total**: You have a fixed budget of bullet points for the whole text. Make sure to write exactly {n_bullets} bullet points in total. This is with respect to all the text you are shown.
- **Coverage**: Since you cannot write more than {n_bullets} bullet points in total, make sure you don't make the list too fine-grained in the beginning. All text must be covered! Use numbers '(i/n)' after the dash as progress indicators with respect to the current section.

### EXAMPLE

Here is an example of the output format:

```md
# HEADING 1 (0 bullet points)

## HEADING 1.1 (2 bullet points)

- (1/2) introduce neural networks
- (2/2) advantage over svm

## HEADING 1.2 (3 bullet points)

- (1/3) derivation of backpropagation
- (2/3) software design of automatic differentiation
- (3/3) example applications
```

### Inputs

Here is the patent application you need to summarize:

```md
{context}
```<|eot_id|>
\end{lstlisting}
\end{minipage}
\end{figure*}
\renewcommand{\sparqlbasicstyle}{\small\tt}

\FloatBarrier

\renewcommand{\sparqlbasicstyle}{\tiny\tt}
\begin{figure*}
  \section{Patent Generation Prompt}\label{sec:generation_prompt_appendix}
  \begin{minipage}{\textwidth}
  \centering
\begin{lstlisting}[language=SPARQL, style=sparqlstyle, caption={System prompt used for outline-guided paper-to-patent generation}]
### ROLE

You are a highly skilled patent attorney with decades of experience in drafting high-quality patent applications.
You assist scientists in transforming their scientific discoveries into lucrative patents.

### TASK DESCRIPTION

Your task is to draft a patent application.

### INPUTS

As input, you will be provided a research paper and a patent outline, each serving a distinct purpose.

1. Research Paper:

The research paper describes a novel invention to be patented.
The scientist has selected the most relevant excerpts from the paper.
Your task is to extract the invention from the paper and write a patent application for it.

2. Patent Outline:

The patent outline summarizes the desired discourse structure of the patent document.
It is in markdown format and contains a number of bullet points per section.
Use this outline as a rough guidance during drafting.
Note that the number of bullet points is also a strong indicator of the desired length! If bullet points are provided, each one corresponds to about 71 words or 1 paragraphs on average.
You should cover all content mentioned in the outline but you are not restricted to it! Feel free to add any further information that you feel would improve the patent application.

3. Prior Patent Outline:

Unless you are asked to generate the beginning of a patent, the user will also provide you with the outline of all prior content.
Use it as global context where you currently stand in the process and do not repeat yourself.

### GUIDELINES

There are a couple of guidelines you need to follow strictly:

- You might be asked to draft only parts of a patent document. Do not draft the whole patent but only those sections requested by the user.
- Copy the headings from the outline exactly. You must include only the headings provided in the outline!
- You must always write complete sentences and avoid keywords, bullet lists and enumerations!
- You must use proper language and maintain a very high level of detail, as you would expect to find in a good patent!
- The patent must act as a standalone document, therefore do not refer to the research paper in the patent!
\end{lstlisting}
\end{minipage}
\end{figure*}

\FloatBarrier

\begin{figure*}
\begin{minipage}{\textwidth}
  \centering
\begin{lstlisting}[language=SPARQL, style=sparqlstyle, caption={User prompt used for outline-guided paper-to-patent generation (Part 1)}]
Here are the most relevant parts of the research paper describing the invention:

```md
# Abstract

Background Heart failure patients with reduced ejection fraction (HFREF) are heterogenous, and our ability to identify patients likely to respond to therapy is limited. We present a method of identifying disease subtypes using high-dimensional clinical phenotyping and latent class analysis that may be useful in personalizing prognosis and treatment in HFREF. Methods A total of 1121 patients with nonischemic HFREF from the β-blocker Evaluation of Survival Trial were categorized according to 27 clinical features. Latent class analysis was used to generate two latent class models, LCM A and B, to identify HFREF subtypes. LCM A consisted of features associated with HF pathogenesis, whereas LCM B consisted of markers of HF progression and severity. The Seattle Heart Failure Model (SHFM) Score was also calculated for all patients. Mortality, improvement in left ventricular ejection fraction (LVEF) defined as an increase in LVEF ≥5% and a final LVEF of 35% after 12 months, and effect of bucindolol on both outcomes were compared across HFREF subtypes. Performance of models that included a combination of LCM subtypes and SHFM scores towards predicting mortality and LVEF response was estimated and subsequently validated using leave-one-out cross-validation and data from the Multicenter Oral Carvedilol Heart Failure Assessment Trial. Results A total of 6 subtypes were identified using LCM A and 5 subtypes using LCM B. Several subtypes resembled familiar clinical phenotypes. Prognosis, improvement in LVEF, and the effect of bucindolol treatment differed significantly between subtypes. Prediction improved with addition of both latent class models to SHFM for both 1-year mortality and LVEF response outcomes. Conclusions The combination of high-dimensional phenotyping and latent class analysis identifies subtypes of HFREF with implications for prognosis and response to specific therapies that may provide insight into mechanisms of disease. These subtypes may facilitate development of personalized treatment plans.

# Introduction

...

We hypothesize that subtypes of nonischemic HFREF exist that may be differentiated by constellations of clinical features that reflect underlying pathophysiology. These subtypes may have variable clinical courses and responses to treatment, and identification of these subtypes may provide insight into mechanisms of HFREF and facilitate personalized prediction of outcomes and treatment response. Traditional outcomes-driven analyses are limited in the number of clinical features that can be evaluated due to the number of potential interactions between features contributing to the development and progression of HFREF. Latent class analysis is one statistical method of identifying groups of individuals within a population that share similar patterns of categorical variables such as symptoms or comorbid conditions, and it has been used in a number of medical disciplines including heart failure for exploration, characterization, and validation of diseases subtypes as well as for risk stratification and prediction of treatment response. [3]–[9] Latent class analysis has also been used to establish diagnostic standards for complex disease syndromes, and use of latent class analysis has been proposed as a method of dealing with large numbers of complex interactions and multiple comparisons in determining likelihood of response to interventions. [10]–[12] Briefly, latent class analysis hypothesizes the existence of unobserved classes within a population that explain patterns of association between variables and uses maximum-likelihood estimation to divide the population into subgroups by calculating a probability of subgroup membership for each symptom or comorbidity. An individual’s subgroup membership may therefore depend on the presence or absence of many different characteristics in a given model. When the population in question has a shared disease, the results are data-driven definitions of disease subtypes where each subtype is characterized by a distinct combination of clinical features. Many clinical variables can thereby be incorporated into an analytic model while preserving statistical power for outcomes analysis by identifying the most prevalent combinations of variables upon which to focus. We propose using complex phenotype descriptions of patients in combination with latent class analysis to identify subtypes of nonischemic HFREF that may have different prognoses and likelihoods of treatment response.

...

# Methods

## Trial Design

The design of BEST has been described previously. [14], [15] A list of all recruitment sites is found in the Appendix S1. All patients had New York Heart Association (NYHA) class III or IV HFREF (LVEF ≤35%) and were randomized in a double-blind fashion to either bucindolol or placebo. Patients were considered ischemic if they had ≥70% obstruction in a major epicardial coronary artery by angiography or evidence of prior myocardial infarction and excluded from this analysis. [16] The primary endpoint was cumulative all-cause mortality. Secondary endpoints were all-cause mortality at one year and LVEF response defined as improvement in LVEF ≥5% with a final LVEF of ≥35% as measured using multi-gated acquisition scan (MUGA). The design of MOCHA has also been described previously. [13] All patients had an LVEF ≤35%, were mostly NYHA class II or III and had stable HF symptoms for 1 month prior to enrollment. They were randomized to placebo, low (6.25 mg bid), medium (12.5 mg bid), or high-dose (25 mg bid) carvedilol. Death and LVEF improvement as measured by MUGA were secondary endpoints in the original MOCHA analysis. Mortality data was only available up to one year of follow-up in MOCHA.

\end{lstlisting}
\end{minipage}
\end{figure*}

\begin{figure*}
\begin{minipage}{\textwidth}
  \centering
\begin{lstlisting}[language=SPARQL, style=sparqlstyle, caption={User prompt used for outline-guided paper-to-patent generation (Part 2)}]
## Identification And Definition Of Latent Classes

Patients were scored according to 27 clinical features (Tables 1 and 2). Criteria were encoded and applied in a MySQL server environment (Oracle Corporation, Redwood Shores, CA). [17] Patient clinical profiles were analyzed collectively using latent class analysis [18] applied to two sets of clinical variables we designated as Latent Class Models (LCM) A and B (Tables 1 and 2). LCM A and B differed only in the clinical variables included in each model. LCM A included variables that describe a patient’s non-cardiac characteristics that can contribute to the pathogenesis of HFREF including age, gender, race, body mass index, and presence of comorbidities such as diabetes, atrial fibrillation, or valvular disease. [19]–[23] LCM B included variables that describe cardiac function, progression, and severity of HFREF including right- and left-ventricular function, hemodynamic parameters such as heart rate and blood pressure, end-organ function such as estimated creatinine clearance, and signs of venous congestion such as jugular venous distension and alanine aminotransferase levels. [24]–[33] In total, 3 variables were included in both models: body mass index, creatinine clearance, and hematocrit. All 3 variables have been implicated in the pathogenesis of HFREF and can also be markers of severity of HFREF. [34], [35] They were included in both models to illustrate that the variable implications of clinical features in different contexts may be represented using this approach. [34], [36]–[40] Two sets of related variables were also included: age of HF onset (LCM A) vs. chronologic age (LCM B) and presence of hypertension (LCM A) vs. presence of hypotension (LCM B). Age of HF onset, a static value, may be relevant to the HFREF etiology, while chronologic age may be related to HF progression. Similarly, presence of hypertension (LCM A) may be related to HF etiology while hypotension (LCM B) may be a marker of advanced HF.

...

## Association Between Latent Class Models And Outcomes

...

## Validation Of Multivariate Models

...

# Results

## Patient Characteristics

...

## Latent Class Model A (Table 1)

LCM A subtypes were characterized by distinct collections of clinical features that frequently resembled known HFREF syndromes. Subtype A1 was characterized by advanced age of onset, non-Caucasian race, male gender, HTN, mild-moderate renal insufficiency, and elevated rates of atrial fibrillation (24.5%). Subtype A2 was characterized by middle age of onset, female gender, moderate renal insufficiency, anemia, high body mass index, and very high rates of diabetes mellitus (74.6%), hypertension (95.0%), hyperlipidemia (93.8%), and hypertriglyceridemia (91.1%). Subtype A3 was characterized by middle age of onset, female gender, Caucasian race, hyperlipidemia, hypertriglyceridemia, anemia, and the presence of left bundle branch block (LBBB). Subtype A4 was characterized by young age of onset, non-Caucasian race, obesity, anemia, and lower rates of traditional cardiac risk factors such as hyperlipidemia, hypertriglyceridemia, and diabetes mellitus. Subtype A5 was characterized by advanced age of onset, Caucasian race, atrial fibrillation (86.2%), mitral valve disease (48.3%), aortic valve disease (21.8%), history of pacemaker placement (42.5%), and a significantly higher rate of prior sudden cardiac death (16.1%). This subtype had the smallest number of subjects (7.8%), whereas subtype A6 was the largest with 28.3% of subjects. Subtype A6 was characterized by middle age of onset, Caucasian race, male gender (100%), high body mass index, hypertension, hyperlipidemia, and hypertriglyceridemia with less associated diabetes mellitus (32.8%) than was seen in Subtype A2.

## Latent Class Model B (Table 2)

...

## Association With Outcomes

...

## Differences In Treatment Effects Between Latent Classes

...

## Combined Models

...

## Model Comparisons

...

## Validation

...

\end{lstlisting}
\end{minipage}
\end{figure*}

\begin{figure*}
\begin{minipage}{\textwidth}
  \centering
\begin{lstlisting}[language=SPARQL, style=sparqlstyle, caption={User prompt used for outline-guided paper-to-patent generation (Part 3)}]

# Discussion

Using the combination of high-dimensional clinical phenotyping and latent class analysis, we have identified a number of HFREF subtypes with distinct clinical profiles that demonstrate significant variation in prognosis as measured by all-cause mortality and response to bucindolol as measured by reduction in mortality and increased likelihood of LVEF response (Figure 4). Several of the LCM A subtypes resemble previously described nonischemic HFREF phenotypes, while LCM B subtypes model HF progression and severity. The latent class models, particularly LCM A, remained significantly associated with certain outcomes after combining them with the SHFM, suggesting that the information in the latent class models is different from the information in the SHFM Score. Taken together, these results suggest that our approach to HFREF subtype identification may be useful for identifying patients with potentially ‘reversible’ HFREF as well as those more likely to benefit from bucindolol.

## Insight Into Mechanisms Of Disease And Treatment Response

...

## Identification Of Hfref Subtypes Using Latent Class Analysis

This analysis demonstrates the potential utility of combining high-dimensional clinical phenotyping and latent class analysis for identifying relevant subtypes of HFREF. It is impossible to determine multivariate odds ratios for all of the variables included in the latent class models presented here using a traditional regression model, as the number of possible interactions (26,542,080 and 432,000,000 for LCM A and LCM B, respectively) prevents calculation using realistic sample sizes. Latent class analysis provides a quantitative mechanism of reducing the number of comparisons by aggregating individuals with similar clinical profiles. Our approach produces data-driven definitions of HFREF subtypes that integrate a large number of clinical features but are not dependent on any one feature for classification. Consequently, a feature like age may not have the same implications among all individuals. For example, subtype A4 is associated with worse outcomes than subtypes A2 or A6 despite younger age and lower burden of comorbid diseases. Clinical features may therefore be associated with a conditional probability for different outcomes depending on their context, capturing relevant interactions between comorbid conditions without direct calculation of all possible interactions. The added value of LCM A and B membership to SHFM for predicting survival despite sharing several common variables suggests that LCM A and B subtype may provide additional prognostic information to the SHFM Score. Finally, the variability in clinical outcomes observed between subtypes suggests that this approach could be useful in identifying patients with higher likelihood of HFREF reversibility in the absence of an obvious reversible etiology or conversely for identifying high risk patients for accelerated advanced HFREF therapy.

## Implementation And Sharing

...

## Limitations

...

Another important limitation is the generalizability of these latent class definitions. Utilization of the coefficients derived in this analysis to determine LCM subtype for other patients assumes that the patient population is the same as the nonischemic patients enrolled in BEST. This assumption may be particularly problematic for LCM B, which includes LVEF in its definition. Like all clinical trials, the inclusion and exclusion criteria of the BEST study are a critical source of selection bias and limit the generalizability of any predictive models developed from BEST to patients that do not meet those entry criteria. [14] This is especially relevant for data-driven latent class models like those presented here, as subtype definitions are by definition dependent on the original study population, and patient subtypes not present in the derivation population might be misidentified. It must also be remembered that latent classes only represent patterns of the variables included in the models, and that those latent classes may not necessarily exist as recognizable patient types in an independent population, [6] due in part to other variables that may be important in a disease process. The utility of these models must therefore be validated further in other patient populations, and the definitions of subtypes will need to be revised over time as more diverse patient populations are incorporated.

...

## Conclusion

High-dimensional phenotyping combined with latent class analysis provide a method of identifying subtypes of nonischemic HFREF patients who may have shared pathophysiology with implications for prognosis and response to bucindolol therapy. Significant reduction in all-cause mortality and increase in likelihood of LVEF response was associated with bucindolol treatment in specific groups identified using these classification methods. Identification of patients’ HFREF subtype may provide a means of personalizing clinical prognosis and estimating likelihood of responding to medical treatment.

```
\end{lstlisting}
\end{minipage}
\end{figure*}

\begin{figure*}
\begin{minipage}{\textwidth}
  \centering
\begin{lstlisting}[language=SPARQL, style=sparqlstyle, caption={User prompt used for outline-guided paper-to-patent generation (Part 4)}]
Here is the outline of the desired patent application. Per bullet point, write roughly 1 paragraphs or 71 words.

First, here is the outline of what you have already written:

```md
# DESCRIPTION

## FIELD OF THE INVENTION

- define field of invention

## BACKGROUND OF THE INVENTION

- describe heart failure
- limitations of current therapy

## SUMMARY OF THE INVENTION

- motivate invention
- introduce HDCP and LCA
- describe subtypes and stages of non-ischemic HF
- correlate subtypes and stages with clinical course and β-blocker response
- describe retrospective analysis of BEST data
- compare results with SHFM predictions
- assess utility of SHFM in predicting response to β-blocker
- list exemplary β-blockers
- describe calculated SHFM Score
- evaluate β-blocker treatment, HF subtype, HF stage, and SHFM
- identify 6 HF subtypes and 5 HF stages
- associate HF subtype, HF stage, and SHFM with mortality and EF improvement
- perform multivariate analysis
- improve predictive performance with HF subtype and HF stage information
- describe method for predicting response to β-blocker therapy
- determine HF subtype and HF stage
- use HDCP to identify HF subtypes and HF stages
- describe method for treating non-ischemic HF patient
- determine treatment procedure for non-ischemic HF patient
- describe apparatus for determining HF subtype, HF stage, or combination thereof
```

Now, continue drafting and add the following points:

```md
# DESCRIPTION

## DETAILED DESCRIPTION OF THE INVENTION

- introduce non-ischemic HF patient prediction methods
- motivate classification of patients into subtypes
- describe clinical features influencing HF subtype and stage
- list clinical conditions used to determine HF subtype and stage
- outline steps to determine HF subtype
- explain calculation of HF subtype probability
- describe determination of HF stage
- outline apparatuses for determining HF subtype and stage
- describe input device for entering clinical condition information
- explain database for storing coefficients
- outline output device for displaying HF subtype and stage
- discuss updating coefficients with additional data
- describe high-throughput phenotyping method
- outline CPU and microprocessor for calculating HF subtype and stage
```
Limit your response to the sections mentioned in the summary: "DETAILED DESCRIPTION OF THE INVENTION". Remember what you have already written and do not repeat yourself.
\end{lstlisting}
\end{minipage}
\end{figure*}

\renewcommand{\sparqlbasicstyle}{\small\tt}

\FloatBarrier

\begin{figure*}[!ht]
  \section{Dataset Statistics}\label{sec:dataset_stats_appendix}
  \begin{minipage}{\textwidth}
  \centering
  \input{img/dataset_fields.pgf}
  \caption{Distribution of domains across dataset splits. Domains are extracted from OpenAlex.}
\end{minipage}
\end{figure*}

\FloatBarrier

\begin{figure*}[!ht]
\begin{minipage}{\textwidth}
  \centering
  \input{img/domains_over_time.pgf}
  \caption{Distribution of domains over time. Domains are extracted from OpenAlex.}
\end{minipage}
\end{figure*}

\begin{figure*}[!ht]
\begin{minipage}{\textwidth}
  \centering
  \input{img/licenses.pgf}
  \caption{Fraction of permissive licenses over time.}
\end{minipage}
\end{figure*}

\begin{figure*}[!ht]
\begin{minipage}{\textwidth}
  \centering
  \input{img/dates_compared.pgf}
  \caption{Date offsets between patents and papers. Negative offset means paper was published first.}
\end{minipage}
\end{figure*}

\begin{table}[t]
  \centering
  \small
  \resizebox{\linewidth}{!}{
    \begin{tabular}{lccccccccccccccc}
      \toprule
      & Style & Length & Coverage & Factuality \\
      \midrule
      \multicolumn{5}{l}{\textbf{Biology (n=152)}} \\
      Llama-3 8B 
                  & 41.7
                  & 12.2k / 23.0k
                  & 39.3 $\pm$ 1.4
                  & 61.0 $\pm$ 0.5
      \\
      Llama-3 70B 
                  & 48.9
                  & \phantom{1}7.7k / 23.0k
                  & 42.0 $\pm$ 0.9
                  & 66.2 $\pm$ 1.2
      \\
      Mixtral-8x7B 
                  & 47.7
                  & \phantom{1}7.4k / 23.0k
                  & 41.3 $\pm$ 1.1
                  & 63.7 $\pm$ 0.6
      \\
      Qwen2-72B 
                  & 48.3
                  & 10.0k / 23.0k
                  & 42.9 $\pm$ 1.1
                  & 63.2 $\pm$ 1.0
      \\
      \cmidrule{1-5}
      \multicolumn{5}{l}{\textbf{Computer Science (n=130)}} \\
      Llama-3 8B 
                  & 37.5
                  & \phantom{1}7.0k / 13.1k
                  & 42.5 $\pm$ 1.2
                  & 61.0 $\pm$ 0.6
      \\
      Llama-3 70B 
                  & 44.5
                  & \phantom{1}4.6k / 13.1k
                  & 44.2 $\pm$ 0.8
                  & 63.4 $\pm$ 0.7
      \\
      Mixtral-8x7B 
                  & 44.7
                  & \phantom{1}4.0k / 13.1k
                  & 43.5 $\pm$ 0.9
                  & 61.2 $\pm$ 0.5
      \\
      Qwen2-72B 
                  & 45.4
                  & \phantom{1}5.3k / 13.1k
                  & 46.7 $\pm$ 0.8
                  & 61.6 $\pm$ 1.0
      \\
      \bottomrule
    \end{tabular}
  }
  \caption{Domain comparison. We report the style score, the length of the generated and reference patents (Length, generated/reference), coverage \textit{Gen \textrightarrow{} Ref}, and factuality \textit{Ref \textrightarrow{} Gen}.}
  \label{tab:domains}
\end{table}

\noindent\textbf{Differences between Domains. }
We analyze the performance across domains (see Appendix \ref{sec:dataset_stats_appendix} for domain distributions) and show the results in \Cref{tab:domains}.
We include the two most represented domains in the dataset: computer science (CS) and biology (Bio).
Reference patents from the biology domain are much longer than computer science patents.
We find that generated Bio patents achieve better factuality but lower coverage across models, despite relative lengths being very similar.
Stylistic similarity is also higher for Bio patents.
Furthermore, model ranking differ between the domains: while Llama-3 70B performs best on Bio patents, Qwen2-72B is highly competitive on CS patents.
However, further analysis is needed due to limited sample sizes.

\FloatBarrier

%% file: sections/091_dataset_appendix.tex
\section{\DatasetName{} Dataset}\label{sec:dataset_appendix}

\begin{table*}[t]
  \small
  \centering
  \setlength{\tabcolsep}{7pt}
  \begin{tabular}{lllll}
    \toprule

    \multicolumn{2}{c}{\textbf{Field}} & \textbf{Patent}                        & \textbf{Paper Candidate 1 (\textcolor{darkgreen}{\checkmark})} & \textbf{Paper Candidate 2 (\textcolor{red}{\ding{55}})}                           \\
    \midrule
    \multirow{2}{*}[-1em]{Authors}
                                       & Content
                                       & Ge Wang, Wenxiang Cong
                                       & \makecell[tl]{Wenxiang Cong, Yan Xi,                                                                                                                                                        \\Bruno De Man, Ge Wang}
                                       & \makecell[tl]{Wenxiang Cong, Yan Xi,                                                                                                                                                        \\Peter Fitzgerald,\\Bruno De Man, Ge Wang}
    \\
                                       & $\text{sim}_\text{author}$             &                                                                & 1.0                                                     & 1.0                     \\
    \cmidrule{1-5}
    \multirow{2}{*}[-2em]{Title}
                                       & Content
                                       & \makecell[tl]{Monochromatic CT Image                                                                                                                                                        \\Reconstruction from Current-\\Integrating Data via Machine Learning}
                                       & \makecell[tl]{Monochromatic image                                                                                                                                                           \\reconstruction via\\machine learning}
                                       & \makecell[tl]{Virtual Monoenergetic                                                                                                                                                         \\CT Imaging via\\Deep Learning}
    \\
                                       & $\text{sim}_\text{term}$               &                                                                & 1.0 / 0.63                                              & 0.39 / 0.32             \\
    \cmidrule{1-5}
    \multirow{2}{*}[-1.5em]{Abstract}
                                       & Content
                                       & \makecell[tl]{A machine-learning-based                                                                                                                                                      \\monochromatic CT image\\reconstruction method is described ...}
                                       & \makecell[tl]{X-ray computed                                                                                                                                                                \\tomography (CT) is a\\ nondestructive imaging ...}
                                       & \makecell[tl]{The physical process                                                                                                                                                          \\of X-ray CT imaging\\is described ...}
    \\
                                       & $\text{sim}_\text{term}$               &                                                                & 0.71 / 0.19                                             & 0.50 / 0.23             \\
    \cmidrule{1-5}
    Date                               &                                        & 2021-08-30                                                     & 2020-11-01 (\checkmark)                                 & 2021-04-14 (\checkmark) \\
    \bottomrule
  \end{tabular}
  \caption{Example for the matching of a PPP. The scores correspond to the respective metrics, the term metrics are shown as $\min$-normalized / $\max$-normalized. Both papers have author lists that contain all the inventors, were published inside the date range, and have titles and abstracts with sufficiently high absolute term similarity to the patent. Since the term similarity scores of paper 1 are higher than those of paper 2 by the specified margin (see Distinctiveness filtering step), paper 1 is correctly matched to the patent.}
  \label{tab:match_example}
\end{table*}

We present the \DatasetName dataset containing 1.8k PPPs from a variety of domains, each annotated with multiple outlines.
It serves two main purposes.
First, it is a challenging benchmark for LLMs that requires long-text generation capabilities and deep understanding of the technical domain as well as patent law. Second, it facilitates the development of AI-powered tools for patent drafting, where the patent attorney only performs post-editing rather than writing from scratch, potentially incurring massive cost savings. In the following, we describe the steps taken for constructing the \DatasetName{} dataset: scraping and filtering PPPs, parsing the full-text documents and generating the patent outlines.

\subsection{Scraping Patent-Paper Pairs}\label{sec:matching}

The patent and paper in a PPP typically do not cite each other, so we cannot rely on front-page or in-text citations \cite{marxRelianceScienceWorldwide2020, marxRelianceScienceInventors2022} to find PPPs. 
The matching must rely on document metadata and content alone. 
We start out with the USPTO dataset\footnote{\url{https://bulkdata.uspto.gov}} containing 6.7M patent applications from 2005 to April 2024.
For each patent, we query SemOpenAlex \cite{farberSemOpenAlexScientificLandscape2023a} using SPARQL and retrieve papers with overlapping authors lists and publication dates.
Next, we filter the results based on their titles, abstracts, other candidate matches for the same patent, and %
paper licenses, as elaborated below. Table \ref{tab:dataset_filters} shows the remaining number of candidates after each filtering step.
As can be seen, the major limiting factor for our benchmark size is the issue of scientific articles being published under restrictive licenses.
An example match is shown in Table \ref{tab:match_example}.
We perform a systematic manual post-hoc evaluation of the heuristics to validate their precision (see Section \ref{sec:validation}).

\noindent\textbf{Author Overlap. } The patent and the paper of a PPP are by definition authored by overlapping sets of individuals. The overlap of author lists have therefore been identified as an effective (yet not sufficient) criteria for matching PPPs \cite{magermanExploringFeasibilityAccuracy2010}.
The requirements for paper authorship are typically much lower than those for patent inventorship \cite{konskiInventorshipAuthorship2015}.
In many cases, only the main author(s) and the senior author(s) are listed as inventors.
We accordingly employ an asymmetric score that only measures the fraction of inventors $i \in I$ that are also authors $a \in A$, not vice versa:
\begin{align*}
  \text{sim}_\text{author} & = \frac{|I \cap A|}{|I|} \geq 0.8
\end{align*}
This score's effectiveness increases with the number of inventors, so we only consider patents with at least two inventors.
Implementing $\text{sim}_\text{author}$ requires some form of author name disambiguation to avoid false negatives (different spellings, e.g., with, without, or with abbreviated middle name) and false positives (e.g., very common names like John Smith).
There are no author identifiers shared between the patent and paper datasets, so our disambiguation uses the surface names only.
To account for false negatives, we use the aliases stored in SemOpenAlex and consider an inventor to be an author if their name matches exactly with one of the aliases.
False positives are marginalized by author combinations and subsequent filters: it is highly unlikely that there exist two groups of people with the same set of names working on the same topic at the same time.

\noindent\textbf{Date Range. } We require the paper's publication date to be within one year before and two years after the patent application date. The former corresponds to the USPTO's grace period,\footnote{\url{https://www.uspto.gov/web/offices/pac/mpep/s2153.html}} which allows inventors to file patent applications up to one year after they disclosed the invention to the public.
The two-year period after the application date was selected because qualitative analyses identified it as the point of diminishing returns, beyond which the incidence of true positives notably decreases, while the rate of false positives significantly increases.

\noindent\textbf{Term Overlap.} We compare titles and abstracts of patents and papers using term overlap metrics.
We first obtain a set of terms $T$ using removal of stopwords and punctuation\footnote{based on spaCy's \texttt{en\_core\_web\_sm}} and stemming.\footnote{based on NLTK's PorterStemmer}
The score is then computed as the number of shared terms, normalized by the minimum or maximum number of terms, following \citet{magermanDoesInvolvementPatenting2015}. We additionally weight each term using its IDF value across all titles and abstracts of patents and papers.
\par\noindent
{
  \small
  \setlength{\abovedisplayskip}{0pt}
  \setlength{\abovedisplayshortskip}{0pt}
  \begin{align*}
    \text{sim}_\text{term}(pat, pap) & = \frac{
      \displaystyle\sum_{t \in T(pat) \cap T(pap)} \text{idf}(t)
    }{
      \displaystyle\text{agg}\Big(\sum_{t \in T(pat)} \text{idf}(t), \sum_{t \in T(pap)} \text{idf}(t)\Big)
    }
  \end{align*}
  \par
}
\noindent where $pat$ and $pap$ are either the titles or the abstracts of the documents, and $\text{agg}\in\{\min, \max\}$.
Thus, we have 4 scores in total, for which we set the thresholds to be 0.15 with $\min$ normalization and 0.1 with $\max$ normalization.
We choose the values based on interactive experimentation, but perform a post-hoc validation of the matching precision.

\noindent\textbf{Distinctiveness. }
In the remaining candidate pairs, there are still many cases where one patent is matched to multiple papers or vice versa. We disambiguate these cases by comparing the term overlap metrics among these ambiguous candidate groups. We only keep a pair if 3 out of 4 term metrics are higher than those of any other candidate in the group by a margin of 0.15 and 0.1, for $\min$ and $\max$, respectively.

\noindent\textbf{License. } We filter the matches for licenses that allow redistribution and commercial use, i.e., 
CC-BY,%
CC0,%
and public domain.
We use the license information provided by SemOpenAlex and by the ArXiv API.

\subsection{Manual Validation}\label{sec:validation}

To verify the precision of our matching pipeline, we perform a manual validation, conducted by the first author of this paper.
We randomly sample 60 PPPs, read both abstracts, skim the documents, compare the figures and get an overview of the authors' related work. We spend roughly 5 minutes per pair on average.
We find that in 55/60 ($91.7\%$) pairs, the paper indeed describes the invention as one of the core contributions.
In three pairs, the best match for the patent would have been a prior paper by the same authors.
In two pairs, the paper would have been best matched to a related but different patent by the same inventors.
This result validates the precision of our matching approach.
In the five imperfect matching cases, the papers still provide meaningful training and evaluation signals, as they are still closely related to the invention and contextualized by the outline.

\subsection{Document Parsing}

We parse all patents and papers into a nested JSON schema where each section has a title, paragraphs and subsections field. We make considerable efforts to obtain clean data: we perform LLM-based section hierarchy reconstruction for patents, font-based section hierarchy reconstruction for paper PDFs and formula conversion for patents and papers. We provide more details in Appendix \ref{sec:parsing_appendix}.

\subsection{Dataset Characteristics}

We split our dataset randomly into \textit{train} (n=1000), \textit{test} (n=500) and \textit{validation} (n=242). We additionally create a non-contaminated test set (\textit{nc-test}) that contains all pairs with a patent published in 2024 (n=71), i.e., after the pretraining cut-off date of all evaluated open-weight LLMs. 
Thus, we address the concern that LLMs might have seen test data during pretraining \cite{ravaut2024much}.
Table \ref{tab:dataset_stats} shows dataset statistics across the splits. Appendix \ref{sec:dataset_stats_appendix} shows further statistics and plots, including the distribution over domains and the number of pairs over time.

In general, both patents and papers contain information not present in the other.
The paper typically includes more experimental details and insights drawn from the experiments.
The patent usually contains more information on the applications and practical benefits of the invention.
We analyze the lexical overlaps between the documents and find that only 2.1\% of the 4-grams are shared.
This underlines the complexity of the task: the two documents describe the same invention from a different perspective using different language.
Nevertheless, we find that it is common for attorneys to copy content from the paper to the patent (or vice versa). 
For instance, many patents and papers share a portion of the figures, as well as verbatim copied text.

%% file: sections/main_table_nc-test.tex
\begin{table*}[t]
    \centering
    \small
    \resizebox{\textwidth}{!}{
      \setlength{\tabcolsep}{3pt}
      \begin{tabular}{lccccccccccc}
        \toprule
                              & \multirow{4}{*}{Tokens}
                              & \multicolumn{2}{c}{\multirow{3}{*}{Text Sim $\uparrow$}}
                              & \multicolumn{3}{c}{Content-Level Metrics (SCALE) $\uparrow$}
                              & \multicolumn{2}{c}{\multirow{3}{*}{Language $\uparrow$}}
                              & \multicolumn{2}{c}{\multirow{3}{*}{Repetitions}}
        \\
        \cmidrule(lr){5-7}
                              &
                              &
                              &
                              & Coverage
                              & \multicolumn{2}{c}{Factuality}
                              &
                              &
                              &
                              &
                              &
        \\
        \cmidrule(lr){3-4}\cmidrule(lr){6-7}\cmidrule(lr){8-9}\cmidrule(lr){10-11}
                              &
                              & BS
                              & R-L
                              & \textit{Gen} \textrightarrow{} \textit{Ref}
                              & \textit{Ref} \textrightarrow{} \textit{Gen}
                              & \textit{Ref} + \textit{Pap} \textrightarrow{} \textit{Gen}
                              & Style
                              & DiscoScore
                              & RR
                              & RR>80
        \\
        \midrule
    
        \multicolumn{8}{l}{\textbf{Heuristic Baselines / Skylines}}\vspace{.4em}                                                                                                      \\
        Reference Patent
                              & \textit{18.4k (100.0\%)}
                              & \textit{100.0}
                              & \textit{100.0}
                              & \textit{88.6 $\pm$ .57}
                              & \textit{88.5 $\pm$ .23}
                              & \textit{88.7 $\pm$ .21}
                              & \textit{99.8}
                              & \textit{100.0}
                              & 15.2
                              & 0.1
        \\
        Similar Patent
                              & 26.1k (141.8\%)
                              & 66.0
                              & 35.3
                              & 31.4 $\pm$ .93
                              & 28.0 $\pm$ .64
                              & 28.3 $\pm$ .68
                              & 53.9
                              & 98.1
                              & 12.5
                              & 0.1
        \\
        Outline
                              & 1.4k (7.6\%)
                              & 56.7
                              & 10.2
                              & 42.1 $\pm$ .94
                              & 63.4 $\pm$ .88
                              & 63.8 $\pm$ .86
                              & 25.9
                              & 85.1
                              & 19.4
                              & 0.0
        \\
        Paper
                              & 9.5k (51.7\%)
                              & 69.6
                              & 42.2
                              & 46.7 $\pm$ 1.01
                              & 46.5 $\pm$ .87
                              & \textit{88.8 $\pm$ .41}
                              & 37.2
                              & 98.0
                              & 8.2
                              & 0.0
        \\
        \cmidrule{1-11}
        \multicolumn{8}{l}{\textbf{Single LLM-call} ($\mathcal{T}$\textit{$\{$inst=2k; pap=$\infty$; pat=$\infty\}$})}\vspace{.4em}                                                                                                          \\
        Mixtral-8x7B
                              & 3.0k (16.3\%)
                              & \textbf{\itshape 66.5}
                              & 21.7
                              & 40.7 $\pm$ 1.16
                              & \textbf{67.2 }$\pm$ 1.19
                              & 73.9 $\pm$ 1.16
                              & \textbf{\itshape 36.5}
                              & 96.8
                              & 15.9
                              & 0.1
        \\
        Qwen2-72B
                              & 2.9k (15.5\%)
                              & 66.4
                              & 21.2
                              & \textbf{\itshape 42.0 }$\pm$ 1.21
                              & 65.0 $\pm$ .76
                              & 71.9 $\pm$ .23
                              & 34.5
                              & \textbf{\itshape 96.9}
                              & \textbf{8.9}
                              & \textbf{0.0}
        \\
        $\quad$w/o Paper
                              & 3.4k (18.2\%)
                              & 65.9
                              & \textbf{\itshape 21.8}
                              & 41.4 $\pm$ 1.25
                              & 66.3 $\pm$ 1.04
                              & 67.0 $\pm$ .99
                              & 34.1
                              & 96.1
                              & 9.0
                              & 0.2
        \\
        $\quad$w/o Outline
                              & 2.0k (11.0\%)
                              & 63.7
                              & 15.8
                              & 36.7 $\pm$ 1.27
                              & 56.5 $\pm$ .67
                              & \textbf{75.7 }$\pm$ 1.13
                              & 30.3
                              & 96.8
                              & \textbf{\itshape 8.2}
                              & \textbf{0.0}
        \\
    
        \cmidrule{1-11}
        \multicolumn{8}{l}{\textbf{\MethodName} ($\mathcal{T}$\textit{$\{$inst=2k; pap=3k; pat=2k$\}$})}\vspace{.4em}                                                                                \\
        Llama-3 8B
                              & 9.6k (52.1\%)
                              & 68.8
                              & 41.4
                              & 42.3 $\pm$ 1.11
                              & 60.3 $\pm$ 1.60
                              & 65.6 $\pm$ 1.23
                              & 36.8
                              & 96.6
                              & 25.4
                              & 2.7
        \\
        Llama-3 8B SFT
                              & 27.0k (146.8\%)
                              & 71.2
                              & 44.0
                              & 44.0 $\pm$ 1.47
                              & 49.4 $\pm$ 1.42
                              & 52.1 $\pm$ 1.36
                              & 45.0
                              & 97.8
                              & 53.7
                              & 27.6
        \\
        $\quad$w/ rep. removal
                              & 18.2k (99.1\%)
                              & \textbf{\itshape 72.1}
                              & \textbf{51.1}
                              & 44.7 $\pm$ 1.03
                              & 52.0 $\pm$ 1.13
                              & 55.4 $\pm$ 1.15
                              & \textbf{50.7}
                              & \textbf{98.2}
                              & 39.6
                              & 8.3
        \\
        Mixtral-8x7B
                              & 6.3k (34.4\%)
                              & 68.8
                              & 34.4
                              & 45.1 $\pm$ 2.00
                              & 64.4 $\pm$ 1.40
                              & \textbf{\itshape 70.6 }$\pm$ 1.24
                              & 39.5
                              & 96.9
                              & 14.4
                              & 0.1
        \\
        Llama-3 70B
                              & 6.1k (33.4\%)
                              & 70.5
                              & 39.5
                              & 45.9 $\pm$ 1.15
                              & 64.7 $\pm$ 1.06
                              & 68.7 $\pm$ .87
                              & 44.6
                              & 97.0
                              & 17.2
                              & 0.1
        \\
        Qwen2-72B
                              & 8.2k (44.7\%)
                              & 70.4
                              & 40.1
                              & \textbf{\itshape 46.9 }$\pm$ 1.26
                              & \textbf{\itshape 65.3 }$\pm$ .83
                              & 69.8 $\pm$ .42
                              & 44.3
                              & 97.0
                              & \textbf{\itshape 10.9}
                              & \textbf{0.0}
        \\
        \cmidrule{1-11}
        \multicolumn{8}{l}{\textbf{\MethodName} ($\mathcal{T}$\textit{$\{$inst=2k; pap=3k; pat=400$\}$})}\vspace{.4em}                                                                                    \\
        Qwen2-72B
                              & 17.2k (93.2\%)
                              & \textbf{\itshape 71.5}
                              & \textbf{\itshape 50.5}
                              & \textbf{49.9 }$\pm$ 1.70
                              & 59.2 $\pm$ 1.00
                              & 64.9 $\pm$ .43
                              & 41.9
                              & 96.6
                              & 8.8
                              & 0.2
        \\
        \bottomrule
    \end{tabular}
    
    }
    \caption{
      Experimental Results on the non-contaminated test set. 
      \textit{Italic values} represent upper bounds that used test data in the prediction. 
      The best value per column is \textbf{bold}, the best per section \textbf{\itshape bold italics}.
      Tokens are reported as absolute and relative to the reference.
      BS = BERTScore. R-L = ROUGE-L. Text Sim = Standard Text Similarity Metrics.
    }
    \label{tab:results-nc-test}
\end{table*}

%% file: img/dates_compared.pgf
\begingroup%
\makeatletter%
\begin{pgfpicture}%
\pgfpathrectangle{\pgfpointorigin}{\pgfqpoint{5.511868in}{1.762907in}}%
\pgfusepath{use as bounding box, clip}%
\begin{pgfscope}%
\pgfsetbuttcap%
\pgfsetmiterjoin%
\definecolor{currentfill}{rgb}{1.000000,1.000000,1.000000}%
\pgfsetfillcolor{currentfill}%
\pgfsetlinewidth{0.000000pt}%
\definecolor{currentstroke}{rgb}{0.500000,0.500000,0.500000}%
\pgfsetstrokecolor{currentstroke}%
\pgfsetdash{}{0pt}%
\pgfpathmoveto{\pgfqpoint{0.000000in}{0.000000in}}%
\pgfpathlineto{\pgfqpoint{5.511868in}{0.000000in}}%
\pgfpathlineto{\pgfqpoint{5.511868in}{1.762907in}}%
\pgfpathlineto{\pgfqpoint{0.000000in}{1.762907in}}%
\pgfpathlineto{\pgfqpoint{0.000000in}{0.000000in}}%
\pgfpathclose%
\pgfusepath{fill}%
\end{pgfscope}%
\begin{pgfscope}%
\pgfsetbuttcap%
\pgfsetmiterjoin%
\definecolor{currentfill}{rgb}{0.898039,0.898039,0.898039}%
\pgfsetfillcolor{currentfill}%
\pgfsetlinewidth{0.000000pt}%
\definecolor{currentstroke}{rgb}{0.000000,0.000000,0.000000}%
\pgfsetstrokecolor{currentstroke}%
\pgfsetstrokeopacity{0.000000}%
\pgfsetdash{}{0pt}%
\pgfpathmoveto{\pgfqpoint{0.529978in}{0.450308in}}%
\pgfpathlineto{\pgfqpoint{5.411868in}{0.450308in}}%
\pgfpathlineto{\pgfqpoint{5.411868in}{1.662907in}}%
\pgfpathlineto{\pgfqpoint{0.529978in}{1.662907in}}%
\pgfpathlineto{\pgfqpoint{0.529978in}{0.450308in}}%
\pgfpathclose%
\pgfusepath{fill}%
\end{pgfscope}%
\begin{pgfscope}%
\pgfpathrectangle{\pgfqpoint{0.529978in}{0.450308in}}{\pgfqpoint{4.881890in}{1.212598in}}%
\pgfusepath{clip}%
\pgfsetrectcap%
\pgfsetroundjoin%
\pgfsetlinewidth{0.803000pt}%
\definecolor{currentstroke}{rgb}{1.000000,1.000000,1.000000}%
\pgfsetstrokecolor{currentstroke}%
\pgfsetdash{}{0pt}%
\pgfpathmoveto{\pgfqpoint{0.594855in}{0.450308in}}%
\pgfpathlineto{\pgfqpoint{0.594855in}{1.662907in}}%
\pgfusepath{stroke}%
\end{pgfscope}%
\begin{pgfscope}%
\pgfsetbuttcap%
\pgfsetroundjoin%
\definecolor{currentfill}{rgb}{0.333333,0.333333,0.333333}%
\pgfsetfillcolor{currentfill}%
\pgfsetlinewidth{0.803000pt}%
\definecolor{currentstroke}{rgb}{0.333333,0.333333,0.333333}%
\pgfsetstrokecolor{currentstroke}%
\pgfsetdash{}{0pt}%
\pgfsys@defobject{currentmarker}{\pgfqpoint{0.000000in}{-0.048611in}}{\pgfqpoint{0.000000in}{0.000000in}}{%
\pgfpathmoveto{\pgfqpoint{0.000000in}{0.000000in}}%
\pgfpathlineto{\pgfqpoint{0.000000in}{-0.048611in}}%
\pgfusepath{stroke,fill}%
}%
\begin{pgfscope}%
\pgfsys@transformshift{0.594855in}{0.450308in}%
\pgfsys@useobject{currentmarker}{}%
\end{pgfscope}%
\end{pgfscope}%
\begin{pgfscope}%
\definecolor{textcolor}{rgb}{0.333333,0.333333,0.333333}%
\pgfsetstrokecolor{textcolor}%
\pgfsetfillcolor{textcolor}%
\pgftext[x=0.594855in,y=0.353086in,,top]{\color{textcolor}{\rmfamily\fontsize{7.000000}{8.400000}\selectfont\catcode`\^=\active\def^{\ifmmode\sp\else\^{}\fi}\catcode`\%=\active\def
\end{pgfscope}%
\begin{pgfscope}%
\pgfpathrectangle{\pgfqpoint{0.529978in}{0.450308in}}{\pgfqpoint{4.881890in}{1.212598in}}%
\pgfusepath{clip}%
\pgfsetrectcap%
\pgfsetroundjoin%
\pgfsetlinewidth{0.803000pt}%
\definecolor{currentstroke}{rgb}{1.000000,1.000000,1.000000}%
\pgfsetstrokecolor{currentstroke}%
\pgfsetdash{}{0pt}%
\pgfpathmoveto{\pgfqpoint{1.421313in}{0.450308in}}%
\pgfpathlineto{\pgfqpoint{1.421313in}{1.662907in}}%
\pgfusepath{stroke}%
\end{pgfscope}%
\begin{pgfscope}%
\pgfsetbuttcap%
\pgfsetroundjoin%
\definecolor{currentfill}{rgb}{0.333333,0.333333,0.333333}%
\pgfsetfillcolor{currentfill}%
\pgfsetlinewidth{0.803000pt}%
\definecolor{currentstroke}{rgb}{0.333333,0.333333,0.333333}%
\pgfsetstrokecolor{currentstroke}%
\pgfsetdash{}{0pt}%
\pgfsys@defobject{currentmarker}{\pgfqpoint{0.000000in}{-0.048611in}}{\pgfqpoint{0.000000in}{0.000000in}}{%
\pgfpathmoveto{\pgfqpoint{0.000000in}{0.000000in}}%
\pgfpathlineto{\pgfqpoint{0.000000in}{-0.048611in}}%
\pgfusepath{stroke,fill}%
}%
\begin{pgfscope}%
\pgfsys@transformshift{1.421313in}{0.450308in}%
\pgfsys@useobject{currentmarker}{}%
\end{pgfscope}%
\end{pgfscope}%
\begin{pgfscope}%
\definecolor{textcolor}{rgb}{0.333333,0.333333,0.333333}%
\pgfsetstrokecolor{textcolor}%
\pgfsetfillcolor{textcolor}%
\pgftext[x=1.421313in,y=0.353086in,,top]{\color{textcolor}{\rmfamily\fontsize{7.000000}{8.400000}\selectfont\catcode`\^=\active\def^{\ifmmode\sp\else\^{}\fi}\catcode`\%=\active\def
\end{pgfscope}%
\begin{pgfscope}%
\pgfpathrectangle{\pgfqpoint{0.529978in}{0.450308in}}{\pgfqpoint{4.881890in}{1.212598in}}%
\pgfusepath{clip}%
\pgfsetrectcap%
\pgfsetroundjoin%
\pgfsetlinewidth{0.803000pt}%
\definecolor{currentstroke}{rgb}{1.000000,1.000000,1.000000}%
\pgfsetstrokecolor{currentstroke}%
\pgfsetdash{}{0pt}%
\pgfpathmoveto{\pgfqpoint{2.247771in}{0.450308in}}%
\pgfpathlineto{\pgfqpoint{2.247771in}{1.662907in}}%
\pgfusepath{stroke}%
\end{pgfscope}%
\begin{pgfscope}%
\pgfsetbuttcap%
\pgfsetroundjoin%
\definecolor{currentfill}{rgb}{0.333333,0.333333,0.333333}%
\pgfsetfillcolor{currentfill}%
\pgfsetlinewidth{0.803000pt}%
\definecolor{currentstroke}{rgb}{0.333333,0.333333,0.333333}%
\pgfsetstrokecolor{currentstroke}%
\pgfsetdash{}{0pt}%
\pgfsys@defobject{currentmarker}{\pgfqpoint{0.000000in}{-0.048611in}}{\pgfqpoint{0.000000in}{0.000000in}}{%
\pgfpathmoveto{\pgfqpoint{0.000000in}{0.000000in}}%
\pgfpathlineto{\pgfqpoint{0.000000in}{-0.048611in}}%
\pgfusepath{stroke,fill}%
}%
\begin{pgfscope}%
\pgfsys@transformshift{2.247771in}{0.450308in}%
\pgfsys@useobject{currentmarker}{}%
\end{pgfscope}%
\end{pgfscope}%
\begin{pgfscope}%
\definecolor{textcolor}{rgb}{0.333333,0.333333,0.333333}%
\pgfsetstrokecolor{textcolor}%
\pgfsetfillcolor{textcolor}%
\pgftext[x=2.247771in,y=0.353086in,,top]{\color{textcolor}{\rmfamily\fontsize{7.000000}{8.400000}\selectfont\catcode`\^=\active\def^{\ifmmode\sp\else\^{}\fi}\catcode`\%=\active\def
\end{pgfscope}%
\begin{pgfscope}%
\pgfpathrectangle{\pgfqpoint{0.529978in}{0.450308in}}{\pgfqpoint{4.881890in}{1.212598in}}%
\pgfusepath{clip}%
\pgfsetrectcap%
\pgfsetroundjoin%
\pgfsetlinewidth{0.803000pt}%
\definecolor{currentstroke}{rgb}{1.000000,1.000000,1.000000}%
\pgfsetstrokecolor{currentstroke}%
\pgfsetdash{}{0pt}%
\pgfpathmoveto{\pgfqpoint{3.074230in}{0.450308in}}%
\pgfpathlineto{\pgfqpoint{3.074230in}{1.662907in}}%
\pgfusepath{stroke}%
\end{pgfscope}%
\begin{pgfscope}%
\pgfsetbuttcap%
\pgfsetroundjoin%
\definecolor{currentfill}{rgb}{0.333333,0.333333,0.333333}%
\pgfsetfillcolor{currentfill}%
\pgfsetlinewidth{0.803000pt}%
\definecolor{currentstroke}{rgb}{0.333333,0.333333,0.333333}%
\pgfsetstrokecolor{currentstroke}%
\pgfsetdash{}{0pt}%
\pgfsys@defobject{currentmarker}{\pgfqpoint{0.000000in}{-0.048611in}}{\pgfqpoint{0.000000in}{0.000000in}}{%
\pgfpathmoveto{\pgfqpoint{0.000000in}{0.000000in}}%
\pgfpathlineto{\pgfqpoint{0.000000in}{-0.048611in}}%
\pgfusepath{stroke,fill}%
}%
\begin{pgfscope}%
\pgfsys@transformshift{3.074230in}{0.450308in}%
\pgfsys@useobject{currentmarker}{}%
\end{pgfscope}%
\end{pgfscope}%
\begin{pgfscope}%
\definecolor{textcolor}{rgb}{0.333333,0.333333,0.333333}%
\pgfsetstrokecolor{textcolor}%
\pgfsetfillcolor{textcolor}%
\pgftext[x=3.074230in,y=0.353086in,,top]{\color{textcolor}{\rmfamily\fontsize{7.000000}{8.400000}\selectfont\catcode`\^=\active\def^{\ifmmode\sp\else\^{}\fi}\catcode`\%=\active\def
\end{pgfscope}%
\begin{pgfscope}%
\pgfpathrectangle{\pgfqpoint{0.529978in}{0.450308in}}{\pgfqpoint{4.881890in}{1.212598in}}%
\pgfusepath{clip}%
\pgfsetrectcap%
\pgfsetroundjoin%
\pgfsetlinewidth{0.803000pt}%
\definecolor{currentstroke}{rgb}{1.000000,1.000000,1.000000}%
\pgfsetstrokecolor{currentstroke}%
\pgfsetdash{}{0pt}%
\pgfpathmoveto{\pgfqpoint{3.900688in}{0.450308in}}%
\pgfpathlineto{\pgfqpoint{3.900688in}{1.662907in}}%
\pgfusepath{stroke}%
\end{pgfscope}%
\begin{pgfscope}%
\pgfsetbuttcap%
\pgfsetroundjoin%
\definecolor{currentfill}{rgb}{0.333333,0.333333,0.333333}%
\pgfsetfillcolor{currentfill}%
\pgfsetlinewidth{0.803000pt}%
\definecolor{currentstroke}{rgb}{0.333333,0.333333,0.333333}%
\pgfsetstrokecolor{currentstroke}%
\pgfsetdash{}{0pt}%
\pgfsys@defobject{currentmarker}{\pgfqpoint{0.000000in}{-0.048611in}}{\pgfqpoint{0.000000in}{0.000000in}}{%
\pgfpathmoveto{\pgfqpoint{0.000000in}{0.000000in}}%
\pgfpathlineto{\pgfqpoint{0.000000in}{-0.048611in}}%
\pgfusepath{stroke,fill}%
}%
\begin{pgfscope}%
\pgfsys@transformshift{3.900688in}{0.450308in}%
\pgfsys@useobject{currentmarker}{}%
\end{pgfscope}%
\end{pgfscope}%
\begin{pgfscope}%
\definecolor{textcolor}{rgb}{0.333333,0.333333,0.333333}%
\pgfsetstrokecolor{textcolor}%
\pgfsetfillcolor{textcolor}%
\pgftext[x=3.900688in,y=0.353086in,,top]{\color{textcolor}{\rmfamily\fontsize{7.000000}{8.400000}\selectfont\catcode`\^=\active\def^{\ifmmode\sp\else\^{}\fi}\catcode`\%=\active\def
\end{pgfscope}%
\begin{pgfscope}%
\pgfpathrectangle{\pgfqpoint{0.529978in}{0.450308in}}{\pgfqpoint{4.881890in}{1.212598in}}%
\pgfusepath{clip}%
\pgfsetrectcap%
\pgfsetroundjoin%
\pgfsetlinewidth{0.803000pt}%
\definecolor{currentstroke}{rgb}{1.000000,1.000000,1.000000}%
\pgfsetstrokecolor{currentstroke}%
\pgfsetdash{}{0pt}%
\pgfpathmoveto{\pgfqpoint{4.727147in}{0.450308in}}%
\pgfpathlineto{\pgfqpoint{4.727147in}{1.662907in}}%
\pgfusepath{stroke}%
\end{pgfscope}%
\begin{pgfscope}%
\pgfsetbuttcap%
\pgfsetroundjoin%
\definecolor{currentfill}{rgb}{0.333333,0.333333,0.333333}%
\pgfsetfillcolor{currentfill}%
\pgfsetlinewidth{0.803000pt}%
\definecolor{currentstroke}{rgb}{0.333333,0.333333,0.333333}%
\pgfsetstrokecolor{currentstroke}%
\pgfsetdash{}{0pt}%
\pgfsys@defobject{currentmarker}{\pgfqpoint{0.000000in}{-0.048611in}}{\pgfqpoint{0.000000in}{0.000000in}}{%
\pgfpathmoveto{\pgfqpoint{0.000000in}{0.000000in}}%
\pgfpathlineto{\pgfqpoint{0.000000in}{-0.048611in}}%
\pgfusepath{stroke,fill}%
}%
\begin{pgfscope}%
\pgfsys@transformshift{4.727147in}{0.450308in}%
\pgfsys@useobject{currentmarker}{}%
\end{pgfscope}%
\end{pgfscope}%
\begin{pgfscope}%
\definecolor{textcolor}{rgb}{0.333333,0.333333,0.333333}%
\pgfsetstrokecolor{textcolor}%
\pgfsetfillcolor{textcolor}%
\pgftext[x=4.727147in,y=0.353086in,,top]{\color{textcolor}{\rmfamily\fontsize{7.000000}{8.400000}\selectfont\catcode`\^=\active\def^{\ifmmode\sp\else\^{}\fi}\catcode`\%=\active\def
\end{pgfscope}%
\begin{pgfscope}%
\definecolor{textcolor}{rgb}{0.333333,0.333333,0.333333}%
\pgfsetstrokecolor{textcolor}%
\pgfsetfillcolor{textcolor}%
\pgftext[x=2.970923in,y=0.211111in,,top]{\color{textcolor}{\rmfamily\fontsize{9.000000}{10.800000}\selectfont\catcode`\^=\active\def^{\ifmmode\sp\else\^{}\fi}\catcode`\%=\active\def
\end{pgfscope}%
\begin{pgfscope}%
\pgfpathrectangle{\pgfqpoint{0.529978in}{0.450308in}}{\pgfqpoint{4.881890in}{1.212598in}}%
\pgfusepath{clip}%
\pgfsetrectcap%
\pgfsetroundjoin%
\pgfsetlinewidth{0.803000pt}%
\definecolor{currentstroke}{rgb}{1.000000,1.000000,1.000000}%
\pgfsetstrokecolor{currentstroke}%
\pgfsetdash{}{0pt}%
\pgfpathmoveto{\pgfqpoint{0.529978in}{0.450308in}}%
\pgfpathlineto{\pgfqpoint{5.411868in}{0.450308in}}%
\pgfusepath{stroke}%
\end{pgfscope}%
\begin{pgfscope}%
\pgfsetbuttcap%
\pgfsetroundjoin%
\definecolor{currentfill}{rgb}{0.333333,0.333333,0.333333}%
\pgfsetfillcolor{currentfill}%
\pgfsetlinewidth{0.803000pt}%
\definecolor{currentstroke}{rgb}{0.333333,0.333333,0.333333}%
\pgfsetstrokecolor{currentstroke}%
\pgfsetdash{}{0pt}%
\pgfsys@defobject{currentmarker}{\pgfqpoint{-0.048611in}{0.000000in}}{\pgfqpoint{-0.000000in}{0.000000in}}{%
\pgfpathmoveto{\pgfqpoint{-0.000000in}{0.000000in}}%
\pgfpathlineto{\pgfqpoint{-0.048611in}{0.000000in}}%
\pgfusepath{stroke,fill}%
}%
\begin{pgfscope}%
\pgfsys@transformshift{0.529978in}{0.450308in}%
\pgfsys@useobject{currentmarker}{}%
\end{pgfscope}%
\end{pgfscope}%
\begin{pgfscope}%
\definecolor{textcolor}{rgb}{0.333333,0.333333,0.333333}%
\pgfsetstrokecolor{textcolor}%
\pgfsetfillcolor{textcolor}%
\pgftext[x=0.377393in, y=0.416551in, left, base]{\color{textcolor}{\rmfamily\fontsize{7.000000}{8.400000}\selectfont\catcode`\^=\active\def^{\ifmmode\sp\else\^{}\fi}\catcode`\%=\active\def
\end{pgfscope}%
\begin{pgfscope}%
\pgfpathrectangle{\pgfqpoint{0.529978in}{0.450308in}}{\pgfqpoint{4.881890in}{1.212598in}}%
\pgfusepath{clip}%
\pgfsetrectcap%
\pgfsetroundjoin%
\pgfsetlinewidth{0.803000pt}%
\definecolor{currentstroke}{rgb}{1.000000,1.000000,1.000000}%
\pgfsetstrokecolor{currentstroke}%
\pgfsetdash{}{0pt}%
\pgfpathmoveto{\pgfqpoint{0.529978in}{0.672396in}}%
\pgfpathlineto{\pgfqpoint{5.411868in}{0.672396in}}%
\pgfusepath{stroke}%
\end{pgfscope}%
\begin{pgfscope}%
\pgfsetbuttcap%
\pgfsetroundjoin%
\definecolor{currentfill}{rgb}{0.333333,0.333333,0.333333}%
\pgfsetfillcolor{currentfill}%
\pgfsetlinewidth{0.803000pt}%
\definecolor{currentstroke}{rgb}{0.333333,0.333333,0.333333}%
\pgfsetstrokecolor{currentstroke}%
\pgfsetdash{}{0pt}%
\pgfsys@defobject{currentmarker}{\pgfqpoint{-0.048611in}{0.000000in}}{\pgfqpoint{-0.000000in}{0.000000in}}{%
\pgfpathmoveto{\pgfqpoint{-0.000000in}{0.000000in}}%
\pgfpathlineto{\pgfqpoint{-0.048611in}{0.000000in}}%
\pgfusepath{stroke,fill}%
}%
\begin{pgfscope}%
\pgfsys@transformshift{0.529978in}{0.672396in}%
\pgfsys@useobject{currentmarker}{}%
\end{pgfscope}%
\end{pgfscope}%
\begin{pgfscope}%
\definecolor{textcolor}{rgb}{0.333333,0.333333,0.333333}%
\pgfsetstrokecolor{textcolor}%
\pgfsetfillcolor{textcolor}%
\pgftext[x=0.322030in, y=0.638638in, left, base]{\color{textcolor}{\rmfamily\fontsize{7.000000}{8.400000}\selectfont\catcode`\^=\active\def^{\ifmmode\sp\else\^{}\fi}\catcode`\%=\active\def
\end{pgfscope}%
\begin{pgfscope}%
\pgfpathrectangle{\pgfqpoint{0.529978in}{0.450308in}}{\pgfqpoint{4.881890in}{1.212598in}}%
\pgfusepath{clip}%
\pgfsetrectcap%
\pgfsetroundjoin%
\pgfsetlinewidth{0.803000pt}%
\definecolor{currentstroke}{rgb}{1.000000,1.000000,1.000000}%
\pgfsetstrokecolor{currentstroke}%
\pgfsetdash{}{0pt}%
\pgfpathmoveto{\pgfqpoint{0.529978in}{0.894484in}}%
\pgfpathlineto{\pgfqpoint{5.411868in}{0.894484in}}%
\pgfusepath{stroke}%
\end{pgfscope}%
\begin{pgfscope}%
\pgfsetbuttcap%
\pgfsetroundjoin%
\definecolor{currentfill}{rgb}{0.333333,0.333333,0.333333}%
\pgfsetfillcolor{currentfill}%
\pgfsetlinewidth{0.803000pt}%
\definecolor{currentstroke}{rgb}{0.333333,0.333333,0.333333}%
\pgfsetstrokecolor{currentstroke}%
\pgfsetdash{}{0pt}%
\pgfsys@defobject{currentmarker}{\pgfqpoint{-0.048611in}{0.000000in}}{\pgfqpoint{-0.000000in}{0.000000in}}{%
\pgfpathmoveto{\pgfqpoint{-0.000000in}{0.000000in}}%
\pgfpathlineto{\pgfqpoint{-0.048611in}{0.000000in}}%
\pgfusepath{stroke,fill}%
}%
\begin{pgfscope}%
\pgfsys@transformshift{0.529978in}{0.894484in}%
\pgfsys@useobject{currentmarker}{}%
\end{pgfscope}%
\end{pgfscope}%
\begin{pgfscope}%
\definecolor{textcolor}{rgb}{0.333333,0.333333,0.333333}%
\pgfsetstrokecolor{textcolor}%
\pgfsetfillcolor{textcolor}%
\pgftext[x=0.322030in, y=0.860726in, left, base]{\color{textcolor}{\rmfamily\fontsize{7.000000}{8.400000}\selectfont\catcode`\^=\active\def^{\ifmmode\sp\else\^{}\fi}\catcode`\%=\active\def
\end{pgfscope}%
\begin{pgfscope}%
\pgfpathrectangle{\pgfqpoint{0.529978in}{0.450308in}}{\pgfqpoint{4.881890in}{1.212598in}}%
\pgfusepath{clip}%
\pgfsetrectcap%
\pgfsetroundjoin%
\pgfsetlinewidth{0.803000pt}%
\definecolor{currentstroke}{rgb}{1.000000,1.000000,1.000000}%
\pgfsetstrokecolor{currentstroke}%
\pgfsetdash{}{0pt}%
\pgfpathmoveto{\pgfqpoint{0.529978in}{1.116571in}}%
\pgfpathlineto{\pgfqpoint{5.411868in}{1.116571in}}%
\pgfusepath{stroke}%
\end{pgfscope}%
\begin{pgfscope}%
\pgfsetbuttcap%
\pgfsetroundjoin%
\definecolor{currentfill}{rgb}{0.333333,0.333333,0.333333}%
\pgfsetfillcolor{currentfill}%
\pgfsetlinewidth{0.803000pt}%
\definecolor{currentstroke}{rgb}{0.333333,0.333333,0.333333}%
\pgfsetstrokecolor{currentstroke}%
\pgfsetdash{}{0pt}%
\pgfsys@defobject{currentmarker}{\pgfqpoint{-0.048611in}{0.000000in}}{\pgfqpoint{-0.000000in}{0.000000in}}{%
\pgfpathmoveto{\pgfqpoint{-0.000000in}{0.000000in}}%
\pgfpathlineto{\pgfqpoint{-0.048611in}{0.000000in}}%
\pgfusepath{stroke,fill}%
}%
\begin{pgfscope}%
\pgfsys@transformshift{0.529978in}{1.116571in}%
\pgfsys@useobject{currentmarker}{}%
\end{pgfscope}%
\end{pgfscope}%
\begin{pgfscope}%
\definecolor{textcolor}{rgb}{0.333333,0.333333,0.333333}%
\pgfsetstrokecolor{textcolor}%
\pgfsetfillcolor{textcolor}%
\pgftext[x=0.322030in, y=1.082814in, left, base]{\color{textcolor}{\rmfamily\fontsize{7.000000}{8.400000}\selectfont\catcode`\^=\active\def^{\ifmmode\sp\else\^{}\fi}\catcode`\%=\active\def
\end{pgfscope}%
\begin{pgfscope}%
\pgfpathrectangle{\pgfqpoint{0.529978in}{0.450308in}}{\pgfqpoint{4.881890in}{1.212598in}}%
\pgfusepath{clip}%
\pgfsetrectcap%
\pgfsetroundjoin%
\pgfsetlinewidth{0.803000pt}%
\definecolor{currentstroke}{rgb}{1.000000,1.000000,1.000000}%
\pgfsetstrokecolor{currentstroke}%
\pgfsetdash{}{0pt}%
\pgfpathmoveto{\pgfqpoint{0.529978in}{1.338659in}}%
\pgfpathlineto{\pgfqpoint{5.411868in}{1.338659in}}%
\pgfusepath{stroke}%
\end{pgfscope}%
\begin{pgfscope}%
\pgfsetbuttcap%
\pgfsetroundjoin%
\definecolor{currentfill}{rgb}{0.333333,0.333333,0.333333}%
\pgfsetfillcolor{currentfill}%
\pgfsetlinewidth{0.803000pt}%
\definecolor{currentstroke}{rgb}{0.333333,0.333333,0.333333}%
\pgfsetstrokecolor{currentstroke}%
\pgfsetdash{}{0pt}%
\pgfsys@defobject{currentmarker}{\pgfqpoint{-0.048611in}{0.000000in}}{\pgfqpoint{-0.000000in}{0.000000in}}{%
\pgfpathmoveto{\pgfqpoint{-0.000000in}{0.000000in}}%
\pgfpathlineto{\pgfqpoint{-0.048611in}{0.000000in}}%
\pgfusepath{stroke,fill}%
}%
\begin{pgfscope}%
\pgfsys@transformshift{0.529978in}{1.338659in}%
\pgfsys@useobject{currentmarker}{}%
\end{pgfscope}%
\end{pgfscope}%
\begin{pgfscope}%
\definecolor{textcolor}{rgb}{0.333333,0.333333,0.333333}%
\pgfsetstrokecolor{textcolor}%
\pgfsetfillcolor{textcolor}%
\pgftext[x=0.266667in, y=1.304901in, left, base]{\color{textcolor}{\rmfamily\fontsize{7.000000}{8.400000}\selectfont\catcode`\^=\active\def^{\ifmmode\sp\else\^{}\fi}\catcode`\%=\active\def
\end{pgfscope}%
\begin{pgfscope}%
\pgfpathrectangle{\pgfqpoint{0.529978in}{0.450308in}}{\pgfqpoint{4.881890in}{1.212598in}}%
\pgfusepath{clip}%
\pgfsetrectcap%
\pgfsetroundjoin%
\pgfsetlinewidth{0.803000pt}%
\definecolor{currentstroke}{rgb}{1.000000,1.000000,1.000000}%
\pgfsetstrokecolor{currentstroke}%
\pgfsetdash{}{0pt}%
\pgfpathmoveto{\pgfqpoint{0.529978in}{1.560746in}}%
\pgfpathlineto{\pgfqpoint{5.411868in}{1.560746in}}%
\pgfusepath{stroke}%
\end{pgfscope}%
\begin{pgfscope}%
\pgfsetbuttcap%
\pgfsetroundjoin%
\definecolor{currentfill}{rgb}{0.333333,0.333333,0.333333}%
\pgfsetfillcolor{currentfill}%
\pgfsetlinewidth{0.803000pt}%
\definecolor{currentstroke}{rgb}{0.333333,0.333333,0.333333}%
\pgfsetstrokecolor{currentstroke}%
\pgfsetdash{}{0pt}%
\pgfsys@defobject{currentmarker}{\pgfqpoint{-0.048611in}{0.000000in}}{\pgfqpoint{-0.000000in}{0.000000in}}{%
\pgfpathmoveto{\pgfqpoint{-0.000000in}{0.000000in}}%
\pgfpathlineto{\pgfqpoint{-0.048611in}{0.000000in}}%
\pgfusepath{stroke,fill}%
}%
\begin{pgfscope}%
\pgfsys@transformshift{0.529978in}{1.560746in}%
\pgfsys@useobject{currentmarker}{}%
\end{pgfscope}%
\end{pgfscope}%
\begin{pgfscope}%
\definecolor{textcolor}{rgb}{0.333333,0.333333,0.333333}%
\pgfsetstrokecolor{textcolor}%
\pgfsetfillcolor{textcolor}%
\pgftext[x=0.266667in, y=1.526989in, left, base]{\color{textcolor}{\rmfamily\fontsize{7.000000}{8.400000}\selectfont\catcode`\^=\active\def^{\ifmmode\sp\else\^{}\fi}\catcode`\%=\active\def
\end{pgfscope}%
\begin{pgfscope}%
\definecolor{textcolor}{rgb}{0.333333,0.333333,0.333333}%
\pgfsetstrokecolor{textcolor}%
\pgfsetfillcolor{textcolor}%
\pgftext[x=0.211111in,y=1.056608in,,bottom,rotate=90.000000]{\color{textcolor}{\rmfamily\fontsize{9.000000}{10.800000}\selectfont\catcode`\^=\active\def^{\ifmmode\sp\else\^{}\fi}\catcode`\%=\active\def
\end{pgfscope}%
\begin{pgfscope}%
\pgfpathrectangle{\pgfqpoint{0.529978in}{0.450308in}}{\pgfqpoint{4.881890in}{1.212598in}}%
\pgfusepath{clip}%
\pgfsetbuttcap%
\pgfsetmiterjoin%
\definecolor{currentfill}{rgb}{0.886275,0.290196,0.200000}%
\pgfsetfillcolor{currentfill}%
\pgfsetlinewidth{0.000000pt}%
\definecolor{currentstroke}{rgb}{0.000000,0.000000,0.000000}%
\pgfsetstrokecolor{currentstroke}%
\pgfsetstrokeopacity{0.000000}%
\pgfsetdash{}{0pt}%
\pgfpathmoveto{\pgfqpoint{0.751882in}{0.450308in}}%
\pgfpathlineto{\pgfqpoint{0.851057in}{0.450308in}}%
\pgfpathlineto{\pgfqpoint{0.851057in}{1.605164in}}%
\pgfpathlineto{\pgfqpoint{0.751882in}{1.605164in}}%
\pgfpathlineto{\pgfqpoint{0.751882in}{0.450308in}}%
\pgfpathclose%
\pgfusepath{fill}%
\end{pgfscope}%
\begin{pgfscope}%
\pgfpathrectangle{\pgfqpoint{0.529978in}{0.450308in}}{\pgfqpoint{4.881890in}{1.212598in}}%
\pgfusepath{clip}%
\pgfsetbuttcap%
\pgfsetmiterjoin%
\definecolor{currentfill}{rgb}{0.886275,0.290196,0.200000}%
\pgfsetfillcolor{currentfill}%
\pgfsetlinewidth{0.000000pt}%
\definecolor{currentstroke}{rgb}{0.000000,0.000000,0.000000}%
\pgfsetstrokecolor{currentstroke}%
\pgfsetstrokeopacity{0.000000}%
\pgfsetdash{}{0pt}%
\pgfpathmoveto{\pgfqpoint{0.875850in}{0.450308in}}%
\pgfpathlineto{\pgfqpoint{0.975026in}{0.450308in}}%
\pgfpathlineto{\pgfqpoint{0.975026in}{1.205406in}}%
\pgfpathlineto{\pgfqpoint{0.875850in}{1.205406in}}%
\pgfpathlineto{\pgfqpoint{0.875850in}{0.450308in}}%
\pgfpathclose%
\pgfusepath{fill}%
\end{pgfscope}%
\begin{pgfscope}%
\pgfpathrectangle{\pgfqpoint{0.529978in}{0.450308in}}{\pgfqpoint{4.881890in}{1.212598in}}%
\pgfusepath{clip}%
\pgfsetbuttcap%
\pgfsetmiterjoin%
\definecolor{currentfill}{rgb}{0.886275,0.290196,0.200000}%
\pgfsetfillcolor{currentfill}%
\pgfsetlinewidth{0.000000pt}%
\definecolor{currentstroke}{rgb}{0.000000,0.000000,0.000000}%
\pgfsetstrokecolor{currentstroke}%
\pgfsetstrokeopacity{0.000000}%
\pgfsetdash{}{0pt}%
\pgfpathmoveto{\pgfqpoint{0.999819in}{0.450308in}}%
\pgfpathlineto{\pgfqpoint{1.098994in}{0.450308in}}%
\pgfpathlineto{\pgfqpoint{1.098994in}{1.160989in}}%
\pgfpathlineto{\pgfqpoint{0.999819in}{1.160989in}}%
\pgfpathlineto{\pgfqpoint{0.999819in}{0.450308in}}%
\pgfpathclose%
\pgfusepath{fill}%
\end{pgfscope}%
\begin{pgfscope}%
\pgfpathrectangle{\pgfqpoint{0.529978in}{0.450308in}}{\pgfqpoint{4.881890in}{1.212598in}}%
\pgfusepath{clip}%
\pgfsetbuttcap%
\pgfsetmiterjoin%
\definecolor{currentfill}{rgb}{0.886275,0.290196,0.200000}%
\pgfsetfillcolor{currentfill}%
\pgfsetlinewidth{0.000000pt}%
\definecolor{currentstroke}{rgb}{0.000000,0.000000,0.000000}%
\pgfsetstrokecolor{currentstroke}%
\pgfsetstrokeopacity{0.000000}%
\pgfsetdash{}{0pt}%
\pgfpathmoveto{\pgfqpoint{1.123788in}{0.450308in}}%
\pgfpathlineto{\pgfqpoint{1.222963in}{0.450308in}}%
\pgfpathlineto{\pgfqpoint{1.222963in}{1.027736in}}%
\pgfpathlineto{\pgfqpoint{1.123788in}{1.027736in}}%
\pgfpathlineto{\pgfqpoint{1.123788in}{0.450308in}}%
\pgfpathclose%
\pgfusepath{fill}%
\end{pgfscope}%
\begin{pgfscope}%
\pgfpathrectangle{\pgfqpoint{0.529978in}{0.450308in}}{\pgfqpoint{4.881890in}{1.212598in}}%
\pgfusepath{clip}%
\pgfsetbuttcap%
\pgfsetmiterjoin%
\definecolor{currentfill}{rgb}{0.886275,0.290196,0.200000}%
\pgfsetfillcolor{currentfill}%
\pgfsetlinewidth{0.000000pt}%
\definecolor{currentstroke}{rgb}{0.000000,0.000000,0.000000}%
\pgfsetstrokecolor{currentstroke}%
\pgfsetstrokeopacity{0.000000}%
\pgfsetdash{}{0pt}%
\pgfpathmoveto{\pgfqpoint{1.247757in}{0.450308in}}%
\pgfpathlineto{\pgfqpoint{1.346932in}{0.450308in}}%
\pgfpathlineto{\pgfqpoint{1.346932in}{1.196523in}}%
\pgfpathlineto{\pgfqpoint{1.247757in}{1.196523in}}%
\pgfpathlineto{\pgfqpoint{1.247757in}{0.450308in}}%
\pgfpathclose%
\pgfusepath{fill}%
\end{pgfscope}%
\begin{pgfscope}%
\pgfpathrectangle{\pgfqpoint{0.529978in}{0.450308in}}{\pgfqpoint{4.881890in}{1.212598in}}%
\pgfusepath{clip}%
\pgfsetbuttcap%
\pgfsetmiterjoin%
\definecolor{currentfill}{rgb}{0.886275,0.290196,0.200000}%
\pgfsetfillcolor{currentfill}%
\pgfsetlinewidth{0.000000pt}%
\definecolor{currentstroke}{rgb}{0.000000,0.000000,0.000000}%
\pgfsetstrokecolor{currentstroke}%
\pgfsetstrokeopacity{0.000000}%
\pgfsetdash{}{0pt}%
\pgfpathmoveto{\pgfqpoint{1.371726in}{0.450308in}}%
\pgfpathlineto{\pgfqpoint{1.470901in}{0.450308in}}%
\pgfpathlineto{\pgfqpoint{1.470901in}{1.196523in}}%
\pgfpathlineto{\pgfqpoint{1.371726in}{1.196523in}}%
\pgfpathlineto{\pgfqpoint{1.371726in}{0.450308in}}%
\pgfpathclose%
\pgfusepath{fill}%
\end{pgfscope}%
\begin{pgfscope}%
\pgfpathrectangle{\pgfqpoint{0.529978in}{0.450308in}}{\pgfqpoint{4.881890in}{1.212598in}}%
\pgfusepath{clip}%
\pgfsetbuttcap%
\pgfsetmiterjoin%
\definecolor{currentfill}{rgb}{0.886275,0.290196,0.200000}%
\pgfsetfillcolor{currentfill}%
\pgfsetlinewidth{0.000000pt}%
\definecolor{currentstroke}{rgb}{0.000000,0.000000,0.000000}%
\pgfsetstrokecolor{currentstroke}%
\pgfsetstrokeopacity{0.000000}%
\pgfsetdash{}{0pt}%
\pgfpathmoveto{\pgfqpoint{1.495694in}{0.450308in}}%
\pgfpathlineto{\pgfqpoint{1.594869in}{0.450308in}}%
\pgfpathlineto{\pgfqpoint{1.594869in}{1.285358in}}%
\pgfpathlineto{\pgfqpoint{1.495694in}{1.285358in}}%
\pgfpathlineto{\pgfqpoint{1.495694in}{0.450308in}}%
\pgfpathclose%
\pgfusepath{fill}%
\end{pgfscope}%
\begin{pgfscope}%
\pgfpathrectangle{\pgfqpoint{0.529978in}{0.450308in}}{\pgfqpoint{4.881890in}{1.212598in}}%
\pgfusepath{clip}%
\pgfsetbuttcap%
\pgfsetmiterjoin%
\definecolor{currentfill}{rgb}{0.886275,0.290196,0.200000}%
\pgfsetfillcolor{currentfill}%
\pgfsetlinewidth{0.000000pt}%
\definecolor{currentstroke}{rgb}{0.000000,0.000000,0.000000}%
\pgfsetstrokecolor{currentstroke}%
\pgfsetstrokeopacity{0.000000}%
\pgfsetdash{}{0pt}%
\pgfpathmoveto{\pgfqpoint{1.619663in}{0.450308in}}%
\pgfpathlineto{\pgfqpoint{1.718838in}{0.450308in}}%
\pgfpathlineto{\pgfqpoint{1.718838in}{1.178756in}}%
\pgfpathlineto{\pgfqpoint{1.619663in}{1.178756in}}%
\pgfpathlineto{\pgfqpoint{1.619663in}{0.450308in}}%
\pgfpathclose%
\pgfusepath{fill}%
\end{pgfscope}%
\begin{pgfscope}%
\pgfpathrectangle{\pgfqpoint{0.529978in}{0.450308in}}{\pgfqpoint{4.881890in}{1.212598in}}%
\pgfusepath{clip}%
\pgfsetbuttcap%
\pgfsetmiterjoin%
\definecolor{currentfill}{rgb}{0.886275,0.290196,0.200000}%
\pgfsetfillcolor{currentfill}%
\pgfsetlinewidth{0.000000pt}%
\definecolor{currentstroke}{rgb}{0.000000,0.000000,0.000000}%
\pgfsetstrokecolor{currentstroke}%
\pgfsetstrokeopacity{0.000000}%
\pgfsetdash{}{0pt}%
\pgfpathmoveto{\pgfqpoint{1.743632in}{0.450308in}}%
\pgfpathlineto{\pgfqpoint{1.842807in}{0.450308in}}%
\pgfpathlineto{\pgfqpoint{1.842807in}{1.107688in}}%
\pgfpathlineto{\pgfqpoint{1.743632in}{1.107688in}}%
\pgfpathlineto{\pgfqpoint{1.743632in}{0.450308in}}%
\pgfpathclose%
\pgfusepath{fill}%
\end{pgfscope}%
\begin{pgfscope}%
\pgfpathrectangle{\pgfqpoint{0.529978in}{0.450308in}}{\pgfqpoint{4.881890in}{1.212598in}}%
\pgfusepath{clip}%
\pgfsetbuttcap%
\pgfsetmiterjoin%
\definecolor{currentfill}{rgb}{0.886275,0.290196,0.200000}%
\pgfsetfillcolor{currentfill}%
\pgfsetlinewidth{0.000000pt}%
\definecolor{currentstroke}{rgb}{0.000000,0.000000,0.000000}%
\pgfsetstrokecolor{currentstroke}%
\pgfsetstrokeopacity{0.000000}%
\pgfsetdash{}{0pt}%
\pgfpathmoveto{\pgfqpoint{1.867601in}{0.450308in}}%
\pgfpathlineto{\pgfqpoint{1.966776in}{0.450308in}}%
\pgfpathlineto{\pgfqpoint{1.966776in}{1.143222in}}%
\pgfpathlineto{\pgfqpoint{1.867601in}{1.143222in}}%
\pgfpathlineto{\pgfqpoint{1.867601in}{0.450308in}}%
\pgfpathclose%
\pgfusepath{fill}%
\end{pgfscope}%
\begin{pgfscope}%
\pgfpathrectangle{\pgfqpoint{0.529978in}{0.450308in}}{\pgfqpoint{4.881890in}{1.212598in}}%
\pgfusepath{clip}%
\pgfsetbuttcap%
\pgfsetmiterjoin%
\definecolor{currentfill}{rgb}{0.886275,0.290196,0.200000}%
\pgfsetfillcolor{currentfill}%
\pgfsetlinewidth{0.000000pt}%
\definecolor{currentstroke}{rgb}{0.000000,0.000000,0.000000}%
\pgfsetstrokecolor{currentstroke}%
\pgfsetstrokeopacity{0.000000}%
\pgfsetdash{}{0pt}%
\pgfpathmoveto{\pgfqpoint{1.991569in}{0.450308in}}%
\pgfpathlineto{\pgfqpoint{2.090744in}{0.450308in}}%
\pgfpathlineto{\pgfqpoint{2.090744in}{1.054387in}}%
\pgfpathlineto{\pgfqpoint{1.991569in}{1.054387in}}%
\pgfpathlineto{\pgfqpoint{1.991569in}{0.450308in}}%
\pgfpathclose%
\pgfusepath{fill}%
\end{pgfscope}%
\begin{pgfscope}%
\pgfpathrectangle{\pgfqpoint{0.529978in}{0.450308in}}{\pgfqpoint{4.881890in}{1.212598in}}%
\pgfusepath{clip}%
\pgfsetbuttcap%
\pgfsetmiterjoin%
\definecolor{currentfill}{rgb}{0.886275,0.290196,0.200000}%
\pgfsetfillcolor{currentfill}%
\pgfsetlinewidth{0.000000pt}%
\definecolor{currentstroke}{rgb}{0.000000,0.000000,0.000000}%
\pgfsetstrokecolor{currentstroke}%
\pgfsetstrokeopacity{0.000000}%
\pgfsetdash{}{0pt}%
\pgfpathmoveto{\pgfqpoint{2.115538in}{0.450308in}}%
\pgfpathlineto{\pgfqpoint{2.214713in}{0.450308in}}%
\pgfpathlineto{\pgfqpoint{2.214713in}{0.992202in}}%
\pgfpathlineto{\pgfqpoint{2.115538in}{0.992202in}}%
\pgfpathlineto{\pgfqpoint{2.115538in}{0.450308in}}%
\pgfpathclose%
\pgfusepath{fill}%
\end{pgfscope}%
\begin{pgfscope}%
\pgfpathrectangle{\pgfqpoint{0.529978in}{0.450308in}}{\pgfqpoint{4.881890in}{1.212598in}}%
\pgfusepath{clip}%
\pgfsetbuttcap%
\pgfsetmiterjoin%
\definecolor{currentfill}{rgb}{0.886275,0.290196,0.200000}%
\pgfsetfillcolor{currentfill}%
\pgfsetlinewidth{0.000000pt}%
\definecolor{currentstroke}{rgb}{0.000000,0.000000,0.000000}%
\pgfsetstrokecolor{currentstroke}%
\pgfsetstrokeopacity{0.000000}%
\pgfsetdash{}{0pt}%
\pgfpathmoveto{\pgfqpoint{2.239507in}{0.450308in}}%
\pgfpathlineto{\pgfqpoint{2.338682in}{0.450308in}}%
\pgfpathlineto{\pgfqpoint{2.338682in}{1.036620in}}%
\pgfpathlineto{\pgfqpoint{2.239507in}{1.036620in}}%
\pgfpathlineto{\pgfqpoint{2.239507in}{0.450308in}}%
\pgfpathclose%
\pgfusepath{fill}%
\end{pgfscope}%
\begin{pgfscope}%
\pgfpathrectangle{\pgfqpoint{0.529978in}{0.450308in}}{\pgfqpoint{4.881890in}{1.212598in}}%
\pgfusepath{clip}%
\pgfsetbuttcap%
\pgfsetmiterjoin%
\definecolor{currentfill}{rgb}{0.886275,0.290196,0.200000}%
\pgfsetfillcolor{currentfill}%
\pgfsetlinewidth{0.000000pt}%
\definecolor{currentstroke}{rgb}{0.000000,0.000000,0.000000}%
\pgfsetstrokecolor{currentstroke}%
\pgfsetstrokeopacity{0.000000}%
\pgfsetdash{}{0pt}%
\pgfpathmoveto{\pgfqpoint{2.363476in}{0.450308in}}%
\pgfpathlineto{\pgfqpoint{2.462651in}{0.450308in}}%
\pgfpathlineto{\pgfqpoint{2.462651in}{0.992202in}}%
\pgfpathlineto{\pgfqpoint{2.363476in}{0.992202in}}%
\pgfpathlineto{\pgfqpoint{2.363476in}{0.450308in}}%
\pgfpathclose%
\pgfusepath{fill}%
\end{pgfscope}%
\begin{pgfscope}%
\pgfpathrectangle{\pgfqpoint{0.529978in}{0.450308in}}{\pgfqpoint{4.881890in}{1.212598in}}%
\pgfusepath{clip}%
\pgfsetbuttcap%
\pgfsetmiterjoin%
\definecolor{currentfill}{rgb}{0.886275,0.290196,0.200000}%
\pgfsetfillcolor{currentfill}%
\pgfsetlinewidth{0.000000pt}%
\definecolor{currentstroke}{rgb}{0.000000,0.000000,0.000000}%
\pgfsetstrokecolor{currentstroke}%
\pgfsetstrokeopacity{0.000000}%
\pgfsetdash{}{0pt}%
\pgfpathmoveto{\pgfqpoint{2.487444in}{0.450308in}}%
\pgfpathlineto{\pgfqpoint{2.586619in}{0.450308in}}%
\pgfpathlineto{\pgfqpoint{2.586619in}{0.938901in}}%
\pgfpathlineto{\pgfqpoint{2.487444in}{0.938901in}}%
\pgfpathlineto{\pgfqpoint{2.487444in}{0.450308in}}%
\pgfpathclose%
\pgfusepath{fill}%
\end{pgfscope}%
\begin{pgfscope}%
\pgfpathrectangle{\pgfqpoint{0.529978in}{0.450308in}}{\pgfqpoint{4.881890in}{1.212598in}}%
\pgfusepath{clip}%
\pgfsetbuttcap%
\pgfsetmiterjoin%
\definecolor{currentfill}{rgb}{0.886275,0.290196,0.200000}%
\pgfsetfillcolor{currentfill}%
\pgfsetlinewidth{0.000000pt}%
\definecolor{currentstroke}{rgb}{0.000000,0.000000,0.000000}%
\pgfsetstrokecolor{currentstroke}%
\pgfsetstrokeopacity{0.000000}%
\pgfsetdash{}{0pt}%
\pgfpathmoveto{\pgfqpoint{2.611413in}{0.450308in}}%
\pgfpathlineto{\pgfqpoint{2.710588in}{0.450308in}}%
\pgfpathlineto{\pgfqpoint{2.710588in}{0.894484in}}%
\pgfpathlineto{\pgfqpoint{2.611413in}{0.894484in}}%
\pgfpathlineto{\pgfqpoint{2.611413in}{0.450308in}}%
\pgfpathclose%
\pgfusepath{fill}%
\end{pgfscope}%
\begin{pgfscope}%
\pgfpathrectangle{\pgfqpoint{0.529978in}{0.450308in}}{\pgfqpoint{4.881890in}{1.212598in}}%
\pgfusepath{clip}%
\pgfsetbuttcap%
\pgfsetmiterjoin%
\definecolor{currentfill}{rgb}{0.886275,0.290196,0.200000}%
\pgfsetfillcolor{currentfill}%
\pgfsetlinewidth{0.000000pt}%
\definecolor{currentstroke}{rgb}{0.000000,0.000000,0.000000}%
\pgfsetstrokecolor{currentstroke}%
\pgfsetstrokeopacity{0.000000}%
\pgfsetdash{}{0pt}%
\pgfpathmoveto{\pgfqpoint{2.735382in}{0.450308in}}%
\pgfpathlineto{\pgfqpoint{2.834557in}{0.450308in}}%
\pgfpathlineto{\pgfqpoint{2.834557in}{0.876717in}}%
\pgfpathlineto{\pgfqpoint{2.735382in}{0.876717in}}%
\pgfpathlineto{\pgfqpoint{2.735382in}{0.450308in}}%
\pgfpathclose%
\pgfusepath{fill}%
\end{pgfscope}%
\begin{pgfscope}%
\pgfpathrectangle{\pgfqpoint{0.529978in}{0.450308in}}{\pgfqpoint{4.881890in}{1.212598in}}%
\pgfusepath{clip}%
\pgfsetbuttcap%
\pgfsetmiterjoin%
\definecolor{currentfill}{rgb}{0.886275,0.290196,0.200000}%
\pgfsetfillcolor{currentfill}%
\pgfsetlinewidth{0.000000pt}%
\definecolor{currentstroke}{rgb}{0.000000,0.000000,0.000000}%
\pgfsetstrokecolor{currentstroke}%
\pgfsetstrokeopacity{0.000000}%
\pgfsetdash{}{0pt}%
\pgfpathmoveto{\pgfqpoint{2.859351in}{0.450308in}}%
\pgfpathlineto{\pgfqpoint{2.958526in}{0.450308in}}%
\pgfpathlineto{\pgfqpoint{2.958526in}{0.885600in}}%
\pgfpathlineto{\pgfqpoint{2.859351in}{0.885600in}}%
\pgfpathlineto{\pgfqpoint{2.859351in}{0.450308in}}%
\pgfpathclose%
\pgfusepath{fill}%
\end{pgfscope}%
\begin{pgfscope}%
\pgfpathrectangle{\pgfqpoint{0.529978in}{0.450308in}}{\pgfqpoint{4.881890in}{1.212598in}}%
\pgfusepath{clip}%
\pgfsetbuttcap%
\pgfsetmiterjoin%
\definecolor{currentfill}{rgb}{0.886275,0.290196,0.200000}%
\pgfsetfillcolor{currentfill}%
\pgfsetlinewidth{0.000000pt}%
\definecolor{currentstroke}{rgb}{0.000000,0.000000,0.000000}%
\pgfsetstrokecolor{currentstroke}%
\pgfsetstrokeopacity{0.000000}%
\pgfsetdash{}{0pt}%
\pgfpathmoveto{\pgfqpoint{2.983319in}{0.450308in}}%
\pgfpathlineto{\pgfqpoint{3.082494in}{0.450308in}}%
\pgfpathlineto{\pgfqpoint{3.082494in}{0.921134in}}%
\pgfpathlineto{\pgfqpoint{2.983319in}{0.921134in}}%
\pgfpathlineto{\pgfqpoint{2.983319in}{0.450308in}}%
\pgfpathclose%
\pgfusepath{fill}%
\end{pgfscope}%
\begin{pgfscope}%
\pgfpathrectangle{\pgfqpoint{0.529978in}{0.450308in}}{\pgfqpoint{4.881890in}{1.212598in}}%
\pgfusepath{clip}%
\pgfsetbuttcap%
\pgfsetmiterjoin%
\definecolor{currentfill}{rgb}{0.886275,0.290196,0.200000}%
\pgfsetfillcolor{currentfill}%
\pgfsetlinewidth{0.000000pt}%
\definecolor{currentstroke}{rgb}{0.000000,0.000000,0.000000}%
\pgfsetstrokecolor{currentstroke}%
\pgfsetstrokeopacity{0.000000}%
\pgfsetdash{}{0pt}%
\pgfpathmoveto{\pgfqpoint{3.107288in}{0.450308in}}%
\pgfpathlineto{\pgfqpoint{3.206463in}{0.450308in}}%
\pgfpathlineto{\pgfqpoint{3.206463in}{0.894484in}}%
\pgfpathlineto{\pgfqpoint{3.107288in}{0.894484in}}%
\pgfpathlineto{\pgfqpoint{3.107288in}{0.450308in}}%
\pgfpathclose%
\pgfusepath{fill}%
\end{pgfscope}%
\begin{pgfscope}%
\pgfpathrectangle{\pgfqpoint{0.529978in}{0.450308in}}{\pgfqpoint{4.881890in}{1.212598in}}%
\pgfusepath{clip}%
\pgfsetbuttcap%
\pgfsetmiterjoin%
\definecolor{currentfill}{rgb}{0.886275,0.290196,0.200000}%
\pgfsetfillcolor{currentfill}%
\pgfsetlinewidth{0.000000pt}%
\definecolor{currentstroke}{rgb}{0.000000,0.000000,0.000000}%
\pgfsetstrokecolor{currentstroke}%
\pgfsetstrokeopacity{0.000000}%
\pgfsetdash{}{0pt}%
\pgfpathmoveto{\pgfqpoint{3.231257in}{0.450308in}}%
\pgfpathlineto{\pgfqpoint{3.330432in}{0.450308in}}%
\pgfpathlineto{\pgfqpoint{3.330432in}{0.867833in}}%
\pgfpathlineto{\pgfqpoint{3.231257in}{0.867833in}}%
\pgfpathlineto{\pgfqpoint{3.231257in}{0.450308in}}%
\pgfpathclose%
\pgfusepath{fill}%
\end{pgfscope}%
\begin{pgfscope}%
\pgfpathrectangle{\pgfqpoint{0.529978in}{0.450308in}}{\pgfqpoint{4.881890in}{1.212598in}}%
\pgfusepath{clip}%
\pgfsetbuttcap%
\pgfsetmiterjoin%
\definecolor{currentfill}{rgb}{0.886275,0.290196,0.200000}%
\pgfsetfillcolor{currentfill}%
\pgfsetlinewidth{0.000000pt}%
\definecolor{currentstroke}{rgb}{0.000000,0.000000,0.000000}%
\pgfsetstrokecolor{currentstroke}%
\pgfsetstrokeopacity{0.000000}%
\pgfsetdash{}{0pt}%
\pgfpathmoveto{\pgfqpoint{3.355226in}{0.450308in}}%
\pgfpathlineto{\pgfqpoint{3.454401in}{0.450308in}}%
\pgfpathlineto{\pgfqpoint{3.454401in}{0.805649in}}%
\pgfpathlineto{\pgfqpoint{3.355226in}{0.805649in}}%
\pgfpathlineto{\pgfqpoint{3.355226in}{0.450308in}}%
\pgfpathclose%
\pgfusepath{fill}%
\end{pgfscope}%
\begin{pgfscope}%
\pgfpathrectangle{\pgfqpoint{0.529978in}{0.450308in}}{\pgfqpoint{4.881890in}{1.212598in}}%
\pgfusepath{clip}%
\pgfsetbuttcap%
\pgfsetmiterjoin%
\definecolor{currentfill}{rgb}{0.886275,0.290196,0.200000}%
\pgfsetfillcolor{currentfill}%
\pgfsetlinewidth{0.000000pt}%
\definecolor{currentstroke}{rgb}{0.000000,0.000000,0.000000}%
\pgfsetstrokecolor{currentstroke}%
\pgfsetstrokeopacity{0.000000}%
\pgfsetdash{}{0pt}%
\pgfpathmoveto{\pgfqpoint{3.479195in}{0.450308in}}%
\pgfpathlineto{\pgfqpoint{3.578370in}{0.450308in}}%
\pgfpathlineto{\pgfqpoint{3.578370in}{0.743464in}}%
\pgfpathlineto{\pgfqpoint{3.479195in}{0.743464in}}%
\pgfpathlineto{\pgfqpoint{3.479195in}{0.450308in}}%
\pgfpathclose%
\pgfusepath{fill}%
\end{pgfscope}%
\begin{pgfscope}%
\pgfpathrectangle{\pgfqpoint{0.529978in}{0.450308in}}{\pgfqpoint{4.881890in}{1.212598in}}%
\pgfusepath{clip}%
\pgfsetbuttcap%
\pgfsetmiterjoin%
\definecolor{currentfill}{rgb}{0.886275,0.290196,0.200000}%
\pgfsetfillcolor{currentfill}%
\pgfsetlinewidth{0.000000pt}%
\definecolor{currentstroke}{rgb}{0.000000,0.000000,0.000000}%
\pgfsetstrokecolor{currentstroke}%
\pgfsetstrokeopacity{0.000000}%
\pgfsetdash{}{0pt}%
\pgfpathmoveto{\pgfqpoint{3.603163in}{0.450308in}}%
\pgfpathlineto{\pgfqpoint{3.702338in}{0.450308in}}%
\pgfpathlineto{\pgfqpoint{3.702338in}{0.787882in}}%
\pgfpathlineto{\pgfqpoint{3.603163in}{0.787882in}}%
\pgfpathlineto{\pgfqpoint{3.603163in}{0.450308in}}%
\pgfpathclose%
\pgfusepath{fill}%
\end{pgfscope}%
\begin{pgfscope}%
\pgfpathrectangle{\pgfqpoint{0.529978in}{0.450308in}}{\pgfqpoint{4.881890in}{1.212598in}}%
\pgfusepath{clip}%
\pgfsetbuttcap%
\pgfsetmiterjoin%
\definecolor{currentfill}{rgb}{0.886275,0.290196,0.200000}%
\pgfsetfillcolor{currentfill}%
\pgfsetlinewidth{0.000000pt}%
\definecolor{currentstroke}{rgb}{0.000000,0.000000,0.000000}%
\pgfsetstrokecolor{currentstroke}%
\pgfsetstrokeopacity{0.000000}%
\pgfsetdash{}{0pt}%
\pgfpathmoveto{\pgfqpoint{3.727132in}{0.450308in}}%
\pgfpathlineto{\pgfqpoint{3.826307in}{0.450308in}}%
\pgfpathlineto{\pgfqpoint{3.826307in}{0.716813in}}%
\pgfpathlineto{\pgfqpoint{3.727132in}{0.716813in}}%
\pgfpathlineto{\pgfqpoint{3.727132in}{0.450308in}}%
\pgfpathclose%
\pgfusepath{fill}%
\end{pgfscope}%
\begin{pgfscope}%
\pgfpathrectangle{\pgfqpoint{0.529978in}{0.450308in}}{\pgfqpoint{4.881890in}{1.212598in}}%
\pgfusepath{clip}%
\pgfsetbuttcap%
\pgfsetmiterjoin%
\definecolor{currentfill}{rgb}{0.886275,0.290196,0.200000}%
\pgfsetfillcolor{currentfill}%
\pgfsetlinewidth{0.000000pt}%
\definecolor{currentstroke}{rgb}{0.000000,0.000000,0.000000}%
\pgfsetstrokecolor{currentstroke}%
\pgfsetstrokeopacity{0.000000}%
\pgfsetdash{}{0pt}%
\pgfpathmoveto{\pgfqpoint{3.851101in}{0.450308in}}%
\pgfpathlineto{\pgfqpoint{3.950276in}{0.450308in}}%
\pgfpathlineto{\pgfqpoint{3.950276in}{0.743464in}}%
\pgfpathlineto{\pgfqpoint{3.851101in}{0.743464in}}%
\pgfpathlineto{\pgfqpoint{3.851101in}{0.450308in}}%
\pgfpathclose%
\pgfusepath{fill}%
\end{pgfscope}%
\begin{pgfscope}%
\pgfpathrectangle{\pgfqpoint{0.529978in}{0.450308in}}{\pgfqpoint{4.881890in}{1.212598in}}%
\pgfusepath{clip}%
\pgfsetbuttcap%
\pgfsetmiterjoin%
\definecolor{currentfill}{rgb}{0.886275,0.290196,0.200000}%
\pgfsetfillcolor{currentfill}%
\pgfsetlinewidth{0.000000pt}%
\definecolor{currentstroke}{rgb}{0.000000,0.000000,0.000000}%
\pgfsetstrokecolor{currentstroke}%
\pgfsetstrokeopacity{0.000000}%
\pgfsetdash{}{0pt}%
\pgfpathmoveto{\pgfqpoint{3.975070in}{0.450308in}}%
\pgfpathlineto{\pgfqpoint{4.074245in}{0.450308in}}%
\pgfpathlineto{\pgfqpoint{4.074245in}{0.716813in}}%
\pgfpathlineto{\pgfqpoint{3.975070in}{0.716813in}}%
\pgfpathlineto{\pgfqpoint{3.975070in}{0.450308in}}%
\pgfpathclose%
\pgfusepath{fill}%
\end{pgfscope}%
\begin{pgfscope}%
\pgfpathrectangle{\pgfqpoint{0.529978in}{0.450308in}}{\pgfqpoint{4.881890in}{1.212598in}}%
\pgfusepath{clip}%
\pgfsetbuttcap%
\pgfsetmiterjoin%
\definecolor{currentfill}{rgb}{0.886275,0.290196,0.200000}%
\pgfsetfillcolor{currentfill}%
\pgfsetlinewidth{0.000000pt}%
\definecolor{currentstroke}{rgb}{0.000000,0.000000,0.000000}%
\pgfsetstrokecolor{currentstroke}%
\pgfsetstrokeopacity{0.000000}%
\pgfsetdash{}{0pt}%
\pgfpathmoveto{\pgfqpoint{4.099038in}{0.450308in}}%
\pgfpathlineto{\pgfqpoint{4.198213in}{0.450308in}}%
\pgfpathlineto{\pgfqpoint{4.198213in}{0.716813in}}%
\pgfpathlineto{\pgfqpoint{4.099038in}{0.716813in}}%
\pgfpathlineto{\pgfqpoint{4.099038in}{0.450308in}}%
\pgfpathclose%
\pgfusepath{fill}%
\end{pgfscope}%
\begin{pgfscope}%
\pgfpathrectangle{\pgfqpoint{0.529978in}{0.450308in}}{\pgfqpoint{4.881890in}{1.212598in}}%
\pgfusepath{clip}%
\pgfsetbuttcap%
\pgfsetmiterjoin%
\definecolor{currentfill}{rgb}{0.886275,0.290196,0.200000}%
\pgfsetfillcolor{currentfill}%
\pgfsetlinewidth{0.000000pt}%
\definecolor{currentstroke}{rgb}{0.000000,0.000000,0.000000}%
\pgfsetstrokecolor{currentstroke}%
\pgfsetstrokeopacity{0.000000}%
\pgfsetdash{}{0pt}%
\pgfpathmoveto{\pgfqpoint{4.223007in}{0.450308in}}%
\pgfpathlineto{\pgfqpoint{4.322182in}{0.450308in}}%
\pgfpathlineto{\pgfqpoint{4.322182in}{0.707930in}}%
\pgfpathlineto{\pgfqpoint{4.223007in}{0.707930in}}%
\pgfpathlineto{\pgfqpoint{4.223007in}{0.450308in}}%
\pgfpathclose%
\pgfusepath{fill}%
\end{pgfscope}%
\begin{pgfscope}%
\pgfpathrectangle{\pgfqpoint{0.529978in}{0.450308in}}{\pgfqpoint{4.881890in}{1.212598in}}%
\pgfusepath{clip}%
\pgfsetbuttcap%
\pgfsetmiterjoin%
\definecolor{currentfill}{rgb}{0.886275,0.290196,0.200000}%
\pgfsetfillcolor{currentfill}%
\pgfsetlinewidth{0.000000pt}%
\definecolor{currentstroke}{rgb}{0.000000,0.000000,0.000000}%
\pgfsetstrokecolor{currentstroke}%
\pgfsetstrokeopacity{0.000000}%
\pgfsetdash{}{0pt}%
\pgfpathmoveto{\pgfqpoint{4.346976in}{0.450308in}}%
\pgfpathlineto{\pgfqpoint{4.446151in}{0.450308in}}%
\pgfpathlineto{\pgfqpoint{4.446151in}{0.752348in}}%
\pgfpathlineto{\pgfqpoint{4.346976in}{0.752348in}}%
\pgfpathlineto{\pgfqpoint{4.346976in}{0.450308in}}%
\pgfpathclose%
\pgfusepath{fill}%
\end{pgfscope}%
\begin{pgfscope}%
\pgfpathrectangle{\pgfqpoint{0.529978in}{0.450308in}}{\pgfqpoint{4.881890in}{1.212598in}}%
\pgfusepath{clip}%
\pgfsetbuttcap%
\pgfsetmiterjoin%
\definecolor{currentfill}{rgb}{0.886275,0.290196,0.200000}%
\pgfsetfillcolor{currentfill}%
\pgfsetlinewidth{0.000000pt}%
\definecolor{currentstroke}{rgb}{0.000000,0.000000,0.000000}%
\pgfsetstrokecolor{currentstroke}%
\pgfsetstrokeopacity{0.000000}%
\pgfsetdash{}{0pt}%
\pgfpathmoveto{\pgfqpoint{4.470945in}{0.450308in}}%
\pgfpathlineto{\pgfqpoint{4.570120in}{0.450308in}}%
\pgfpathlineto{\pgfqpoint{4.570120in}{0.716813in}}%
\pgfpathlineto{\pgfqpoint{4.470945in}{0.716813in}}%
\pgfpathlineto{\pgfqpoint{4.470945in}{0.450308in}}%
\pgfpathclose%
\pgfusepath{fill}%
\end{pgfscope}%
\begin{pgfscope}%
\pgfpathrectangle{\pgfqpoint{0.529978in}{0.450308in}}{\pgfqpoint{4.881890in}{1.212598in}}%
\pgfusepath{clip}%
\pgfsetbuttcap%
\pgfsetmiterjoin%
\definecolor{currentfill}{rgb}{0.886275,0.290196,0.200000}%
\pgfsetfillcolor{currentfill}%
\pgfsetlinewidth{0.000000pt}%
\definecolor{currentstroke}{rgb}{0.000000,0.000000,0.000000}%
\pgfsetstrokecolor{currentstroke}%
\pgfsetstrokeopacity{0.000000}%
\pgfsetdash{}{0pt}%
\pgfpathmoveto{\pgfqpoint{4.594913in}{0.450308in}}%
\pgfpathlineto{\pgfqpoint{4.694088in}{0.450308in}}%
\pgfpathlineto{\pgfqpoint{4.694088in}{0.778998in}}%
\pgfpathlineto{\pgfqpoint{4.594913in}{0.778998in}}%
\pgfpathlineto{\pgfqpoint{4.594913in}{0.450308in}}%
\pgfpathclose%
\pgfusepath{fill}%
\end{pgfscope}%
\begin{pgfscope}%
\pgfpathrectangle{\pgfqpoint{0.529978in}{0.450308in}}{\pgfqpoint{4.881890in}{1.212598in}}%
\pgfusepath{clip}%
\pgfsetbuttcap%
\pgfsetmiterjoin%
\definecolor{currentfill}{rgb}{0.886275,0.290196,0.200000}%
\pgfsetfillcolor{currentfill}%
\pgfsetlinewidth{0.000000pt}%
\definecolor{currentstroke}{rgb}{0.000000,0.000000,0.000000}%
\pgfsetstrokecolor{currentstroke}%
\pgfsetstrokeopacity{0.000000}%
\pgfsetdash{}{0pt}%
\pgfpathmoveto{\pgfqpoint{4.718882in}{0.450308in}}%
\pgfpathlineto{\pgfqpoint{4.818057in}{0.450308in}}%
\pgfpathlineto{\pgfqpoint{4.818057in}{0.690163in}}%
\pgfpathlineto{\pgfqpoint{4.718882in}{0.690163in}}%
\pgfpathlineto{\pgfqpoint{4.718882in}{0.450308in}}%
\pgfpathclose%
\pgfusepath{fill}%
\end{pgfscope}%
\begin{pgfscope}%
\pgfpathrectangle{\pgfqpoint{0.529978in}{0.450308in}}{\pgfqpoint{4.881890in}{1.212598in}}%
\pgfusepath{clip}%
\pgfsetbuttcap%
\pgfsetmiterjoin%
\definecolor{currentfill}{rgb}{0.886275,0.290196,0.200000}%
\pgfsetfillcolor{currentfill}%
\pgfsetlinewidth{0.000000pt}%
\definecolor{currentstroke}{rgb}{0.000000,0.000000,0.000000}%
\pgfsetstrokecolor{currentstroke}%
\pgfsetstrokeopacity{0.000000}%
\pgfsetdash{}{0pt}%
\pgfpathmoveto{\pgfqpoint{4.842851in}{0.450308in}}%
\pgfpathlineto{\pgfqpoint{4.942026in}{0.450308in}}%
\pgfpathlineto{\pgfqpoint{4.942026in}{0.645745in}}%
\pgfpathlineto{\pgfqpoint{4.842851in}{0.645745in}}%
\pgfpathlineto{\pgfqpoint{4.842851in}{0.450308in}}%
\pgfpathclose%
\pgfusepath{fill}%
\end{pgfscope}%
\begin{pgfscope}%
\pgfpathrectangle{\pgfqpoint{0.529978in}{0.450308in}}{\pgfqpoint{4.881890in}{1.212598in}}%
\pgfusepath{clip}%
\pgfsetbuttcap%
\pgfsetmiterjoin%
\definecolor{currentfill}{rgb}{0.886275,0.290196,0.200000}%
\pgfsetfillcolor{currentfill}%
\pgfsetlinewidth{0.000000pt}%
\definecolor{currentstroke}{rgb}{0.000000,0.000000,0.000000}%
\pgfsetstrokecolor{currentstroke}%
\pgfsetstrokeopacity{0.000000}%
\pgfsetdash{}{0pt}%
\pgfpathmoveto{\pgfqpoint{4.966820in}{0.450308in}}%
\pgfpathlineto{\pgfqpoint{5.065995in}{0.450308in}}%
\pgfpathlineto{\pgfqpoint{5.065995in}{0.716813in}}%
\pgfpathlineto{\pgfqpoint{4.966820in}{0.716813in}}%
\pgfpathlineto{\pgfqpoint{4.966820in}{0.450308in}}%
\pgfpathclose%
\pgfusepath{fill}%
\end{pgfscope}%
\begin{pgfscope}%
\pgfpathrectangle{\pgfqpoint{0.529978in}{0.450308in}}{\pgfqpoint{4.881890in}{1.212598in}}%
\pgfusepath{clip}%
\pgfsetbuttcap%
\pgfsetmiterjoin%
\definecolor{currentfill}{rgb}{0.886275,0.290196,0.200000}%
\pgfsetfillcolor{currentfill}%
\pgfsetlinewidth{0.000000pt}%
\definecolor{currentstroke}{rgb}{0.000000,0.000000,0.000000}%
\pgfsetstrokecolor{currentstroke}%
\pgfsetstrokeopacity{0.000000}%
\pgfsetdash{}{0pt}%
\pgfpathmoveto{\pgfqpoint{5.090788in}{0.450308in}}%
\pgfpathlineto{\pgfqpoint{5.189963in}{0.450308in}}%
\pgfpathlineto{\pgfqpoint{5.189963in}{0.672396in}}%
\pgfpathlineto{\pgfqpoint{5.090788in}{0.672396in}}%
\pgfpathlineto{\pgfqpoint{5.090788in}{0.450308in}}%
\pgfpathclose%
\pgfusepath{fill}%
\end{pgfscope}%
\begin{pgfscope}%
\pgfsetrectcap%
\pgfsetmiterjoin%
\pgfsetlinewidth{1.003750pt}%
\definecolor{currentstroke}{rgb}{1.000000,1.000000,1.000000}%
\pgfsetstrokecolor{currentstroke}%
\pgfsetdash{}{0pt}%
\pgfpathmoveto{\pgfqpoint{0.529978in}{0.450308in}}%
\pgfpathlineto{\pgfqpoint{0.529978in}{1.662907in}}%
\pgfusepath{stroke}%
\end{pgfscope}%
\begin{pgfscope}%
\pgfsetrectcap%
\pgfsetmiterjoin%
\pgfsetlinewidth{1.003750pt}%
\definecolor{currentstroke}{rgb}{1.000000,1.000000,1.000000}%
\pgfsetstrokecolor{currentstroke}%
\pgfsetdash{}{0pt}%
\pgfpathmoveto{\pgfqpoint{5.411868in}{0.450308in}}%
\pgfpathlineto{\pgfqpoint{5.411868in}{1.662907in}}%
\pgfusepath{stroke}%
\end{pgfscope}%
\begin{pgfscope}%
\pgfsetrectcap%
\pgfsetmiterjoin%
\pgfsetlinewidth{1.003750pt}%
\definecolor{currentstroke}{rgb}{1.000000,1.000000,1.000000}%
\pgfsetstrokecolor{currentstroke}%
\pgfsetdash{}{0pt}%
\pgfpathmoveto{\pgfqpoint{0.529978in}{0.450308in}}%
\pgfpathlineto{\pgfqpoint{5.411868in}{0.450308in}}%
\pgfusepath{stroke}%
\end{pgfscope}%
\begin{pgfscope}%
\pgfsetrectcap%
\pgfsetmiterjoin%
\pgfsetlinewidth{1.003750pt}%
\definecolor{currentstroke}{rgb}{1.000000,1.000000,1.000000}%
\pgfsetstrokecolor{currentstroke}%
\pgfsetdash{}{0pt}%
\pgfpathmoveto{\pgfqpoint{0.529978in}{1.662907in}}%
\pgfpathlineto{\pgfqpoint{5.411868in}{1.662907in}}%
\pgfusepath{stroke}%
\end{pgfscope}%
\end{pgfpicture}%
\makeatother%
\endgroup%